\documentclass{article} 
\usepackage{iclr2027_conference,times}

\usepackage{amsmath,amsfonts,bm}

\def\eqref#1{equation~\ref{#1}}

\def\1{\bm{1}}

\DeclareMathAlphabet{\mathsfit}{\encodingdefault}{\sfdefault}{m}{sl}
\SetMathAlphabet{\mathsfit}{bold}{\encodingdefault}{\sfdefault}{bx}{n}

\usepackage{hyperref}
\usepackage{enumitem}
\usepackage{url}

\usepackage{graphicx}
\usepackage{booktabs}
\usepackage{xcolor}
\usepackage{colortbl}
\usepackage{tabularx}
\newcolumntype{Y}{>{\raggedright\arraybackslash}X}

\title{Mandela-Bench: Multimodal Models \\ Remember Canonical Images \\ Instead of Seeing Them}

\author{
\normalfont
Yicheng Bao\textsuperscript{1,*},
Zhenkun Gao\textsuperscript{1,*},
Xiahui Guo\textsuperscript{1,*},
Mingqian Yang\textsuperscript{1},\\
Xueheng Li\textsuperscript{2},
Bangwei Liu\textsuperscript{1},
Mingang Chen\textsuperscript{3},\\
Lijun Li\textsuperscript{4},
Xuhong Wang\textsuperscript{4},
Xin Tan\textsuperscript{1}\\[4pt]
\textsuperscript{1}East China Normal University\\
\textsuperscript{2}University of Science and Technology of China\\
\textsuperscript{3}Shanghai Development Center of Computer Software Technology\\
\textsuperscript{4}Shanghai Artificial Intelligence Laboratory\\[2pt]
\textsuperscript{*}Equal contribution.
}

\iclrfinalcopy 
\begin{document}

\maketitle

\begin{abstract}
Historical photographs and other canonical images can now be edited seamlessly with a single instruction, often leaving no reliable pixel-level trace. In such cases, the only evidence of manipulation may be a fact about what the image depicts. Existing benchmarks instead rely on generator artefacts, image–caption inconsistencies, visual implausibilities, or external references, and therefore do not test whether a model can use its own world knowledge to verify a recognized image. We introduce Mandela-Bench, containing 1,507 edits of canonical images: 1,359 knowledge-only forgeries, each contradicting one verifiable fact, and 148 anchor-free controls that preserve the editing process without introducing a factual contradiction, together with 474 untouched originals. We score not only whether a model detects a forgery, but whether its explanation identifies the inserted entity or the fact being violated. Across 36 multimodal models, from 0.8B parameters to frontier scale, we find a consistent failure mode. When a public figure is removed from a familiar photograph, models still name that person in up to 72.7\% of responses. Some models can distinguish the replacement face from the original when shown in isolation, yet still judge the full edited photograph as authentic. Providing the true event and date does not improve knowledge-grounded detection, whereas providing the same information after cropping away the recognizable composition does. Even under explicit verification prompts, only one of the 36 models meets the KGR criterion on at least half of the forged images. These results suggest that the failures cannot be explained by missing knowledge or inadequate perception alone. Instead, they are consistent with recognition biasing verification toward the remembered canonical image rather than the observed edit.
\end{abstract}

\section{Introduction}
\label{sec:intro}

Historical photographs are how a society remembers what happened: who stood on the balcony, who signed the treaty, what the banner said. Photographic manipulation has a long history, but modern instruction-based editors \citep{brooks2023instructpix2pix,zhao2024ultraedit} can now produce seamless edits whose only inconsistency is factual. A person may appear in an event that occurred after their death, an object may appear before it existed, or a banner may contain text that was never there.
When paired with its correct caption, such an image can evade pixel-level detectors \citep{xu2025aigisurvey} while altering the record it appears to document. The only evidence against such an image is knowledge of what it depicts. The same problem arises for other images treated as records, including championship photographs, film stills, and famous paintings. We refer to these as \emph{canonical images}. Multimodal large language models (MLLMs) can both inspect these images and possess relevant factual knowledge. We ask whether they use that knowledge to verify a recognized image when no caption, reference database, or external tool is available. Existing approaches instead rely on generator traces, cross-modal inconsistencies, visible semantic anomalies, or external references. Section~\ref{sec:related} places our setting in the context of this prior work.

\begin{figure}[t]
\centering
\includegraphics[width=\linewidth]{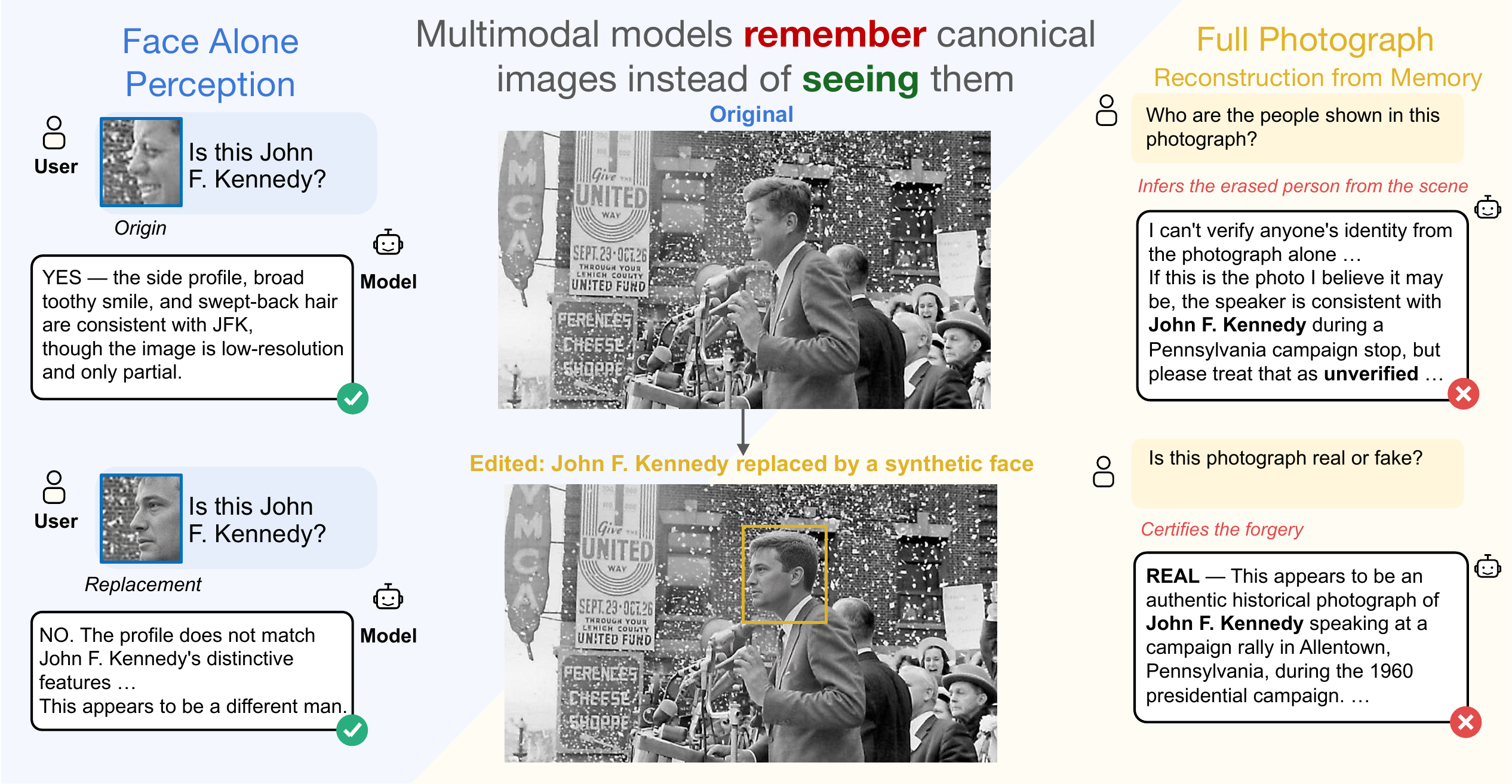}
\caption{A canonical photograph of John F. Kennedy with his face replaced by a synthetic one. The model rejects the replacement face in isolation, yet infers Kennedy from the full scene and judges the edited photograph authentic.}
\label{fig:teaser}
\end{figure}

Three shortcuts complicate a knowledge benchmark: editor traces can reveal an edit, familiarity can bias authenticity judgments, and a correct ``fake'' verdict can rest on the wrong evidence. This motivates two questions: when a model recognizes an image, does recognition trigger verification or replace it? And when a forgery passes, does the model lack the relevant fact, fail to see the edit, or fail to use what it knows?

We introduce \textbf{Mandela-Bench}, a single-image forgery benchmark in which the only evidence of manipulation is a fact about the depicted referent (Figure~\ref{fig:teaser}). The benchmark contains 1,507 seamlessly edited canonical images: 1,359 \emph{knowledge-only} forgeries, each violating exactly one recorded and verifiable fact, and 148 \emph{anchor-free} controls that undergo comparable edits without introducing any factual contradiction. Two pixel detectors based on different detection principles perform near chance on these edited images, suggesting that low-level forensic traces provide little reliable evidence for detection. These items cover six knowledge families—people, objects, places, visible text, event attributes, and works of art—and six image domains ranging from historical photographs to film stills. Pairing controls for editing artefacts: anchor-free controls are generated from the same source images using the same editors and operations, so they share the same editing traces while introducing no factual contradiction. Each source image also contributes an untouched authentic version (474 in total), so familiarity with the canonical image itself cannot serve as a reliable cue for authenticity. 

To address the final shortcut, we use an \emph{evidence ladder} to grade the reason behind each verdict. Our primary metric, the \emph{knowledge-grounded rate} (KGR), credits only verdicts that identify the manipulated entity or the fact it contradicts. We then use a set of diagnostic probes to determine where failures arise: a text-only probe tests whether the model possesses the relevant fact, recognition and identity/face probes test whether it recognizes the scene and perceives the manipulated identity, and the verification probe tests whether that knowledge is actually used in the final judgment. Finally, to test whether recognition itself affects verification, we introduce \emph{de-recognition conditions} that alter the image context without changing the manipulated region itself: cropping removes the familiar global composition while preserving the edited content, and a correct caption restores the event and date without restoring the visual recognition cue.

Across 36 MLLMs from 0.8B parameters to frontier scale, we find a consistent pattern: \emph{When models recognize a canonical image, their verification often follows the remembered version more than the visual evidence in the edited image.} Models may name erased people and accept a full forgery even when they reject its replacement face in isolation. Providing the true event and date does not improve knowledge-grounded detection, whereas providing the same information after cropping away the recognizable composition does. We refer to this behavioral pattern as reconstruction from memory, and call it the \emph{Mandela effect} in multimodal models. Three findings support reconstruction from memory rather than a purely perceptual explanation.
When a public figure is erased from a canonical photograph and the model is asked who appears in it, Kimi-K3 still names the erased person in 62.1\% of responses. Even when it rejects the replacement face in isolation, it judges the full edited photograph authentic 63.6\% of the time on those same items. Providing the true event and date does not improve knowledge-grounded detection. Yet when the same information is paired with a crop that removes the recognizable composition, all twenty-one models produce more knowledge-grounded verdicts. Overall, Knowledge-grounded detection remains limited. Under prompted verification, the best model reaches 55.1 KGR and the next-best 36.0 (Table~\ref{tab:main}). Models ground object and text contradictions far more often than a person who could not have been present, while anchor-free edits almost never receive knowledge-grounded verdicts.

\section{Related Work}
\label{sec:related}

Image-forgery detection typically draws evidence from the input itself or from information supplied alongside it. Forensic detectors exploit generator-specific traces \citep{xu2025aigisurvey,ojha2023ufd,tan2024npr}, while cross-modal methods detect inconsistencies between images and captions \citep{shao2023dgm4,huang2024miragenews}. Semantic-anomaly, reasoning-based, and manipulation-localization methods instead identify visible implausibilities or forensic disruptions around edited regions \citep{tan2025anomreason,tan2026sml,xu2025manipshield,nguyen2026editsleuth,li2026impostor,bao2026spared}. Reference-grounded systems retrieve authentic images or records and compare them with the query \citep{cui2024lookupforensics,zhou2026reveal}. Benchmarks such as LOKI and FakeBench similarly evaluate MLLMs on synthetic or otherwise cue-bearing content \citep{ye2024loki,li2024fakebench}. Mandela-Bench targets a different setting: a single seamless image with no caption or external reference, for which the decisive evidence is a fact the model may already know about the recognized referent.

Our question is also related to work on conflicts between visual evidence and parametric priors. HallusionBench shows cases in which stored knowledge overrides altered visual input \citep{guan2024hallusionbench}, while Pixels Versus Priors studies controlled visual counterfacts \citep{golovanevsky-etal-2025-pixels}, and MLLMs Know Where to Look examines failures on small visual details despite attention to the relevant region \citep{zhang2025mllmsknow}. We instead study real canonical images, manipulate their recognizability while preserving the edited region, and separately probe perception, fact possession, and final verification. A fuller comparison, including detector robustness, human memory, and canonical-image recognition, is provided in Appendix~\ref{app:related}.

\section{Mandela-Bench: Forgeries That Only Knowledge Can Catch}
\label{sec:benchmark}

Mandela-Bench tests whether multimodal models use the facts they possess to verify images they recognize. To isolate the role of recognition, each item must satisfy two requirements: the forgery must be detectable only through a fact about the depicted referent (\S\ref{sec:forgeries}), and recognizability must be manipulable without altering the edited region (\S\ref{sec:conditions}). 

\subsection{Knowledge-only forgeries and their controls}
\label{sec:forgeries}

A \emph{knowledge-only forgery} is an edit of a canonical image that (i)~contradicts exactly one recorded, externally verifiable fact about the depicted referent and (ii)~is otherwise indistinguishable from an authentic image of the same scene.
Operationally, the \emph{seed} (the canonical image to be edited) is anchored to a knowledge-base entry, so that contradictions are checkable by machine as well as by the reader.
The \emph{single contradiction} is stored with the item (``George Orwell died in 1950, before 1986 World Cup final: Maradona lifts the trophy (1986-06-29)'').
The edit is \emph{seamless} in era, medium, and style, and \emph{intrinsically plausible} (absent the fact, the scene is unremarkable).
\emph{Pixel cleanliness} is measured after construction (\S\ref{sec:qc}).
What the forgery gets wrong defines its \emph{knowledge family}: the people present (K1), the objects (K2), the place (K3), the visible text (K4), an event attribute (K5), or the content of a work of art or film (K6).
The K1/K2 contrast between knowing an individual and knowing a class of things is taken up in Section~\ref{sec:analysis}.

\begin{table}[t]
\centering
\caption{Mandela-Bench item taxonomy. Full visual examples and the
six-domain distribution are shown in Appendix Figure~\ref{fig:overview-full}.}
\label{tab:taxonomy}

\small
\setlength{\tabcolsep}{8pt}
\renewcommand{\arraystretch}{1.0}

\begin{tabularx}{\linewidth}{@{}XX@{}}
\toprule

\textbf{K1 People}
&
\textbf{K5 Event attribute}
\\[-1pt]
Karl May (died in 1912) inserted
&
Zidane's shirt recolored blue (France wore white)
\\[-1pt]
\textit{Scene:} \emph{Casablanca} (1942)
&
\textit{Scene:} 2006 World Cup final
\\[4pt]

\textbf{K2 Objects}
&
\textbf{K6 Artwork}
\\[-1pt]
Jet airliner (first flew in 1949) inserted
&
Extra figure added (not in the canonical composition)
\\[-1pt]
\textit{Scene:} McKinley campaign print (1900)
&
\textit{Scene:} Leonardo's \emph{The Last Supper}
\\[4pt]

\textbf{K3 Place}
&
\textbf{K0 Control}
\\[-1pt]
Vietnam Memorial (D.C.) added
&
Unnamed person swapped (no factual contradiction)
\\[-1pt]
\textit{Scene:} Salt March, India (1930)
&
\textit{Scene:} 2008 Brazilian Grand Prix
\\[4pt]

\textbf{K4 Visible text}
&
\textbf{Authentic}
\\[-1pt]
NAGOYA replaces NAGANO (wrong host city)
&
Untouched original (no edit)
\\[-1pt]
\textit{Scene:} Nagano Winter Olympics emblem (1998)
&
\textit{Scene:} 1986 World Cup final
\\

\bottomrule
\end{tabularx}
\end{table}

\label{sec:controls}Every family is paired with an \emph{anchor-free control} (K0): the same seeds, editors, and operations, but the inserted element (an unnamed bystander, an ordinary object) carries no fact.
With traces and composition matched, the detection difference between K0 and a family is the net contribution of knowledge.
Every seed also contributes its untouched original as an \emph{authentic item} (1{,}507 forged, 474 authentic).
Forged and authentic images thus share provenance, fame, medium, and resolution; calling every famous image fake gains nothing on balanced accuracy. Table~\ref{tab:taxonomy} summarizes the item types; full visual examples and the six-domain distribution are shown in Appendix Figure~\ref{fig:overview-full}. Obscure but genuinely present people are left untouched, so flagging every unrecognized face gains nothing either.
For the people family we add the reverse operation.
\emph{Absent-celebrity} items replace the public figure who \emph{is} in the canonical image with a synthetic face under a local elliptical mask, so the only evidence that anything is wrong is the memory of who should be there.
Each is paired on the same seed with a celebrity-swap item and a K0 item.

\subsection{Generation and quality control}
\label{sec:qc}

\begin{figure}[t]
\centering
\includegraphics[width=\linewidth]{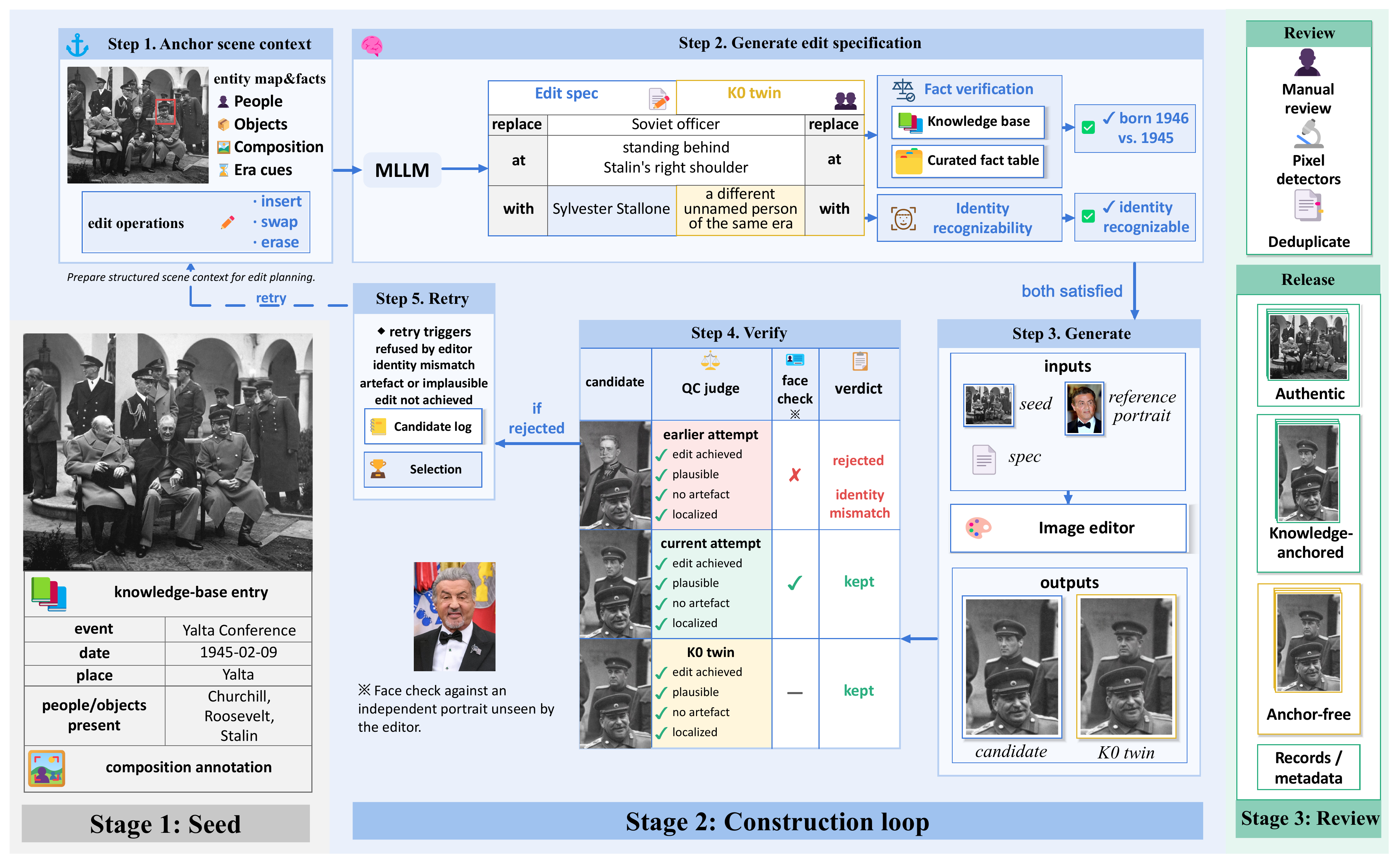}
\caption{
Construction of Mandela-Bench for one 1945 Yalta Conference seed.
Stage~1 anchors the seed to a knowledge-base entry.
Stage~2 generates a knowledge-anchored edit and matched K0 control, automatically checks candidates, and retries failures.
Stage~3 applies manual review, pixel-detector auditing, and deduplication before release. Here the first proposal failed the face check.}
\label{fig:pipeline}
\end{figure}

Figure~\ref{fig:pipeline} traces one seed through the construction. Seeds are canonical images from Wikimedia Commons and open-access film, museum, and meme archives, with each image linked to a knowledge-base entry and annotated for composition. A multimodal model proposes edit specifications, which are verified against Wikidata or curated fact tables before rendering. Named people are used only when probe model (Kimi-K3) can identify their portraits without scene context, thereby ensuring that the identities the benchmark relies on are recognizable from the face alone. One image editor (GPT Image 2) then renders both knowledge-anchored edits and their anchor-free controls, ensuring that both inherit the same editor-specific traces. Quality control combines automated identity, object, and artefact checks with manual review and a final pixel audit. Across 1,981 content-deduplicated images, NPR~\citep{tan2024npr} and UniversalFakeDetect~\citep{ojha2023ufd} remain near chance (AUC 0.577 and 0.538), with true-positive rates of 0.00\% and 7.28\% at 5\% false positives.

\section{Evaluation Protocol: Probes, Conditions, and Scoring}
\label{sec:protocol}

\begin{figure}[t]
\centering
\includegraphics[width=\linewidth]{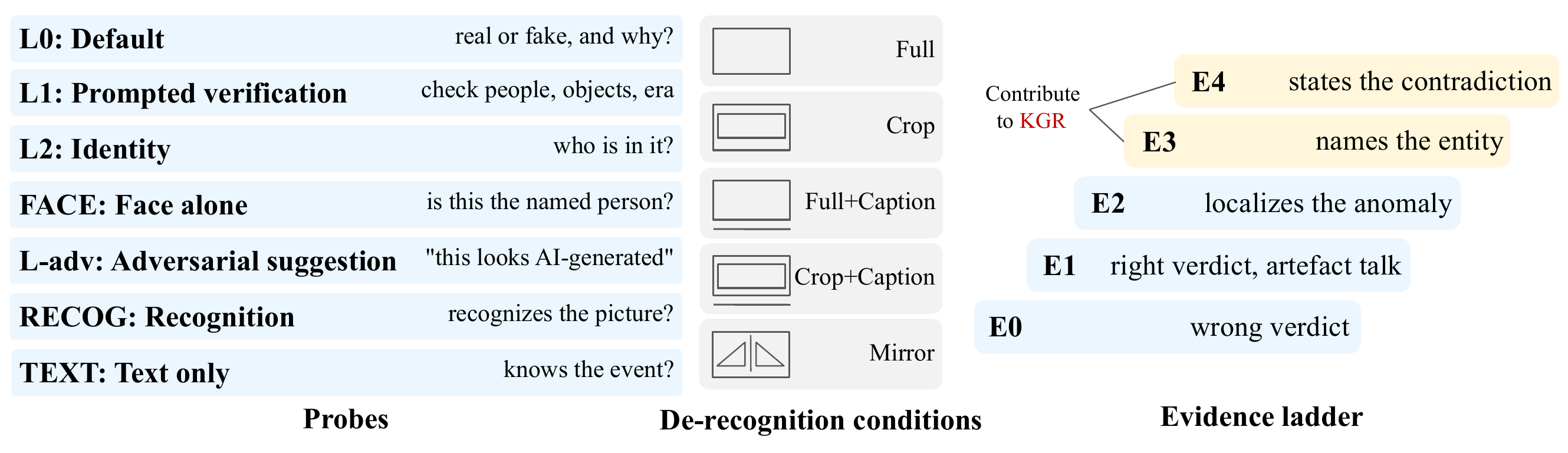}
\caption{Evaluation overview. Left: 7 probes (\S\ref{sec:probes}). Middle: 5 de-recognition conditions under L1 (\S\ref{sec:conditions}). Right: E0--E4 evidence ladder, with E3+ defining KGR (\S\ref{sec:scoring}).}
\label{fig:protocol}
\end{figure}

\subsection{Probes}
\label{sec:probes}

A single verdict cannot tell knowledge from luck, so every image is shown under seven probes (Figure~\ref{fig:protocol}) and conditions that vary the cue of recognition (\S\ref{sec:conditions}), with the reason behind each verdict graded (\S\ref{sec:scoring}).
\textbf{Default} (L0) asks whether the image is authentic or generated or manipulated, and why; it is the primary condition.
\textbf{Prompted verification} (L1) asks the model to check whether the people, objects, and era shown are consistent with what it knows.
\textbf{Identity} (L2) asks who is in the picture and nothing about authenticity; on absent-celebrity items, naming the erased person provides a direct behavioral signature of reconstruction from memory.
\textbf{Face} (FACE), on absent-celebrity items only, shows the face region alone, from the original and the edit, and asks whether it is the named person.
\textbf{Adversarial suggestion} (L-adv), on authentic images only, asserts the image looks AI-generated.
\textbf{Recognition} (RECOG), the self-reported mediator, asks whether the model recognizes the picture.
\textbf{Text-only} (TEXT) asks, without an image, whether the named person could have been present.

\subsection{De-recognition conditions}
\label{sec:conditions}

We apply de-recognition conditions to \emph{peripheral} forgeries, where edits lie outside the main figure. This lets us crop away the recognizable composition without changing the edited region.
\textbf{Full} shows the complete image, while \textbf{Full+Caption} adds a sentence naming the true event and date.
\textbf{Crop} keeps the edited region and nearby figures but removes the familiar composition and some event context. This isolation control can bias models toward ``fake''.
\textbf{Crop+Caption} adds the same caption to the crop, restoring event context but not the composition. Its gain in balanced accuracy over Full is the recognition penalty.
\textbf{Mirror} horizontally flips the full image as a perturbation control, changing its appearance while largely preserving scene recognition.
Authentic images undergo the same operations, and all conditions use prompted verification.

\subsection{Scoring}
\label{sec:scoring}

We first parse each response into a REAL or FAKE verdict. Responses that provide neither are counted as misses. For forged images, we report the detection rate, while for authentic images we report accuracy. Their average gives balanced accuracy, which prevents a model from appearing strong simply because it has a general tendency to answer FAKE.

We then evaluate the evidence supporting each verdict using the \emph{evidence ladder} in Figure~\ref{fig:protocol}. E0 denotes an incorrect verdict. E1 denotes a correct FAKE verdict supported only by generic or unverifiable artefact descriptions. E2 is assigned when the model correctly localizes the manipulated region but does not identify the anomalous entity. E3 requires the model to correctly identify the inserted, removed, or replaced entity, and E4 additionally requires it to state the factual contradiction that makes the image inconsistent with the recorded event. An LLM judge (Qwen3.8-27B) assigns these levels by comparing each explanation with the item’s recorded contradiction.

Our primary metric is the \emph{Knowledge-Grounded Rate} (KGR), defined as the fraction of the 1,507 forged items, including K0 controls, that receive a score of E3 or above. Because E3 requires correct identification of the manipulated entity but not the event-specific contradiction, KGR should be interpreted as entity-or-fact grounding rather than explicit factual reasoning. E4 is the stricter fact-grounded criterion, requiring the recorded contradiction itself. Simply labeling many images as FAKE, or explaining them using generic artefact cues, does not increase KGR. We also report two derived quantities. The \emph{elicitation gap} is the difference in KGR between prompted verification (L1) and the default probe (L0), measuring how much explicit verification prompting improves knowledge use. The \emph{recognition penalty} is the difference in balanced accuracy between Crop+Caption and Full, measuring whether performance improves when visual recognition of the canonical image is weakened while event context is restored. We treat an effect as established only when detection, balanced accuracy, and KGR move in the same direction. If detection rises only because authentic images are increasingly mislabeled as fake, we interpret this as a shift in response bias rather than improved knowledge-grounded verification. Construction details, verbatim prompts, the judge rubric and calibration, and per-model settings and references appear in Appendices~\ref{app:construction}--\ref{app:models}.

\section{Results}
\label{sec:results}
We evaluate 36 MLLMs, including two proprietary (GPT-5.6 and Gemini-3.8-Flash) and 34 open-weight models ranging from 0.8B to 2.8T parameters (Appendix~\ref{app:models}).
Table~\ref{tab:main} reports the 15 newest models on the release set. The de-recognition analysis uses 145 peripheral people-family forgeries and their authentic counterparts, while the absent-celebrity analysis uses 132 verified items with same-seed swap and K0 controls (Table~\ref{tab:conditions}).
Full 36-model results are in Appendix Tables~\ref{tab:main-all} and~\ref{tab:people-all}.

\begin{table}[t]
\centering
\caption{Main results on the release set (1,507 forged, 474 authentic; strict scoring) for the 15 main-text models; all 36 are in Appendix Table~\ref{tab:main-all}. KGR, Det, Auth, and Bal denote knowledge-grounded rate, forgery detection, authentic-image accuracy, and balanced accuracy. Left: L0, L1, and L-adv results. Right: L0 KGR by knowledge family. Colors show changes from the corresponding baseline (L1 and L-adv vs.\ L0; K1--K6 vs.\ K0): blue better, red worse; $\Delta$KGR = L1$-$L0 is defined as elicitation gap; $^{\dagger}$Judge model. Rows are sorted by L1 KGR.}
\label{tab:main}

\scriptsize\renewcommand{\arraystretch}{0.96}
\setlength{\tabcolsep}{1.6pt}
\begin{tabular}{@{}l rrrr rrrr r rrrrrrr r@{}}
\toprule
 & \multicolumn{4}{c}{Default (L0)} & \multicolumn{4}{c}{Prompted (L1)} & L-adv & \multicolumn{7}{c}{KGR by knowledge family (L0)} & \multicolumn{1}{c}{\cellcolor{gray!12}Derived} \\
\cmidrule(lr){2-5}\cmidrule(lr){6-9}\cmidrule(lr){10-10}\cmidrule(lr){11-17}\cmidrule(lr){18-18}
Model & KGR & Det & Auth & Bal & KGR & Det & Auth & Bal & Auth & K0 & K1 & K2 & K3 & K4 & K5 & K6 & \cellcolor{gray!12}$\Delta$KGR \\
\midrule
\multicolumn{18}{@{}l}{\emph{Proprietary}} \\
Gemini-3.8-Flash & 46.6 & 66.3 & 94.7 & 80.5 & \cellcolor[HTML]{CBDAF4}55.1 & \cellcolor[HTML]{A8C2ED}80.5 & \cellcolor[HTML]{F7D7D7}88.2 & \cellcolor[HTML]{E8EFFA}84.3 & \cellcolor[HTML]{F3F7FD}99.4 & 0.7 & \cellcolor[HTML]{C8D9F3}27.5 & \cellcolor[HTML]{84A9E5}79.2 & \cellcolor[HTML]{89ACE6}58.4 & \cellcolor[HTML]{8DAFE7}56.5 & \cellcolor[HTML]{84A9E5}66.2 & \cellcolor[HTML]{84A9E5}66.0 & \cellcolor{gray!12}+8.5 \\
GPT-5.6 & 19.8 & 40.9 & 95.8 & 68.3 & \cellcolor[HTML]{9BB9EA}36.0 & \cellcolor[HTML]{84A9E5}61.0 & \cellcolor[HTML]{FAE5E5}91.6 & \cellcolor[HTML]{CEDDF5}76.3 & \cellcolor[HTML]{F7F9FD}99.2 & 0.0 & \cellcolor[HTML]{F3F7FC}5.8 & \cellcolor[HTML]{8FB0E7}54.8 & \cellcolor[HTML]{D8E4F7}19.1 & \cellcolor[HTML]{C8D8F3}26.9 & \cellcolor[HTML]{D5E1F6}20.7 & \cellcolor[HTML]{DAE5F7}17.9 & \cellcolor{gray!12}+16.2 \\
\addlinespace[2pt]
\multicolumn{18}{@{}l}{\emph{Open weights}} \\
GLM-5.3-Flash & 29.7 & 53.8 & 92.6 & 73.2 & \cellcolor[HTML]{E0E9F8}34.8 & \cellcolor[HTML]{D5E2F6}60.6 & \cellcolor[HTML]{F6D6D6}85.9 & 73.2 & \cellcolor[HTML]{EEF3FB}99.4 & 2.0 & \cellcolor[HTML]{F4F7FD}7.6 & \cellcolor[HTML]{84A9E5}62.2 & \cellcolor[HTML]{B0C8EE}40.5 & \cellcolor[HTML]{91B2E8}55.6 & \cellcolor[HTML]{B1C9EF}40.0 & \cellcolor[HTML]{AEC6EE}41.7 & \cellcolor{gray!12}+5.1 \\
Kimi-K3 & 24.4 & 45.5 & 95.6 & 70.5 & \cellcolor[HTML]{C3D5F2}34.1 & \cellcolor[HTML]{9BB9EA}61.7 & \cellcolor[HTML]{FBEAEA}92.2 & \cellcolor[HTML]{D7E3F7}77.0 & \cellcolor[HTML]{F6F9FD}99.2 & 0.7 & \cellcolor[HTML]{F5F8FD}5.8 & \cellcolor[HTML]{92B3E8}53.7 & \cellcolor[HTML]{D2E0F6}22.5 & \cellcolor[HTML]{84A9E5}63.9 & \cellcolor[HTML]{C1D4F2}31.0 & \cellcolor[HTML]{C9D9F4}26.9 & \cellcolor{gray!12}+9.7 \\
Kimi-K2.6 & 22.2 & 39.7 & 94.7 & 67.2 & \cellcolor[HTML]{C3D5F2}32.0 & \cellcolor[HTML]{99B8E9}56.3 & \cellcolor[HTML]{F9E3E3}90.1 & \cellcolor[HTML]{DAE5F7}73.2 & \cellcolor[HTML]{F9FBFE}97.3 & 0.0 & \cellcolor[HTML]{F0F5FC}7.2 & \cellcolor[HTML]{93B3E8}52.7 & \cellcolor[HTML]{D8E4F7}19.1 & \cellcolor[HTML]{9CBAEA}48.1 & \cellcolor[HTML]{C4D5F2}29.0 & \cellcolor[HTML]{DCE6F8}17.3 & \cellcolor{gray!12}+9.8 \\
Qwen3.5-122B-A10B & 23.6 & 48.4 & 84.4 & 66.4 & \cellcolor[HTML]{CEDDF5}31.5 & \cellcolor[HTML]{C6D7F3}57.7 & \cellcolor[HTML]{F6D4D4}77.4 & \cellcolor[HTML]{F8FAFE}67.5 & \cellcolor[HTML]{F4CACA}62.7 & 0.7 & \cellcolor[HTML]{ECF2FB}9.9 & \cellcolor[HTML]{98B7E9}51.0 & \cellcolor[HTML]{D9E5F7}19.1 & \cellcolor[HTML]{AFC7EE}39.8 & \cellcolor[HTML]{B6CCEF}36.5 & \cellcolor[HTML]{D3E0F6}22.4 & \cellcolor{gray!12}+7.9 \\
Qwen3-VL-235B-A22B & 18.1 & 34.4 & 90.7 & 62.6 & \cellcolor[HTML]{C9D9F4}26.9 & \cellcolor[HTML]{84A9E5}63.7 & \cellcolor[HTML]{E58484}70.0 & \cellcolor[HTML]{E5EDF9}66.9 & \cellcolor[HTML]{EDF2FB}98.1 & 0.0 & \cellcolor[HTML]{F2F6FC}6.2 & \cellcolor[HTML]{94B4E8}52.4 & \cellcolor[HTML]{CFDDF5}23.6 & \cellcolor[HTML]{C2D5F2}29.6 & \cellcolor[HTML]{E1EAF9}14.5 & \cellcolor[HTML]{F2F6FC}6.4 & \cellcolor{gray!12}+8.8 \\
Qwen3.6-27B & 19.8 & 46.2 & 78.9 & 62.6 & \cellcolor[HTML]{D8E4F7}26.1 & \cellcolor[HTML]{9EBBEB}61.9 & \cellcolor[HTML]{EEADAD}65.6 & \cellcolor[HTML]{F8FAFD}63.8 & \cellcolor[HTML]{F7DADA}63.9 & 0.7 & \cellcolor[HTML]{F6F8FD}5.3 & \cellcolor[HTML]{96B6E9}51.7 & \cellcolor[HTML]{E5EDF9}13.5 & \cellcolor[HTML]{ABC4ED}41.7 & \cellcolor[HTML]{CFDDF5}24.1 & \cellcolor[HTML]{E2EBF9}14.7 & \cellcolor{gray!12}+6.3 \\
Qwen3.5-27B & 17.9 & 36.8 & 73.4 & 55.1 & \cellcolor[HTML]{D4E1F6}24.9 & \cellcolor[HTML]{9BB9EA}53.1 & \cellcolor[HTML]{E99696}56.3 & \cellcolor[HTML]{FEFDFD}54.7 & \cellcolor[HTML]{FCF1F1}67.9 & 0.7 & \cellcolor[HTML]{F8FAFE}3.9 & \cellcolor[HTML]{A9C3ED}42.5 & \cellcolor[HTML]{E3EBF9}14.6 & \cellcolor[HTML]{9CBAEA}49.1 & \cellcolor[HTML]{D6E2F6}20.7 & \cellcolor[HTML]{DEE8F8}16.7 & \cellcolor{gray!12}+7.0 \\
Qwen3.8-27B$^{\dagger}$ & 13.3 & 40.0 & 80.2 & 60.1 & \cellcolor[HTML]{C8D8F3}22.3 & \cellcolor[HTML]{84A9E5}66.0 & \cellcolor[HTML]{E78F8F}62.0 & \cellcolor[HTML]{E7EEFA}64.0 & \cellcolor[HTML]{F9FBFE}82.5 & 1.4 & \cellcolor[HTML]{FFFEFE}1.1 & \cellcolor[HTML]{B3CAEF}38.4 & \cellcolor[HTML]{F6F9FD}5.6 & \cellcolor[HTML]{B4CBEF}38.0 & \cellcolor[HTML]{E6EDFA}13.8 & \cellcolor[HTML]{EFF4FC}9.0 & \cellcolor{gray!12}+9.0 \\
Qwen3.6-35B-A3B & 17.6 & 34.7 & 76.4 & 55.5 & \cellcolor[HTML]{E5EDFA}21.8 & \cellcolor[HTML]{D7E3F7}41.2 & \cellcolor[HTML]{ECA5A5}61.8 & \cellcolor[HTML]{FAE6E6}51.5 & \cellcolor[HTML]{FAEAEA}67.7 & 0.7 & \cellcolor[HTML]{F8FAFD}4.2 & \cellcolor[HTML]{9DBBEA}48.3 & \cellcolor[HTML]{E0EAF8}15.7 & \cellcolor[HTML]{BED2F1}32.4 & \cellcolor[HTML]{D0DEF5}23.4 & \cellcolor[HTML]{EDF2FB}9.6 & \cellcolor{gray!12}+4.2 \\
Gemma-4-31B & 11.7 & 67.0 & 44.3 & 55.7 & \cellcolor[HTML]{D0DEF5}19.4 & \cellcolor[HTML]{97B6E9}83.9 & \cellcolor[HTML]{F2C0C0}34.0 & \cellcolor[HTML]{EBF1FB}58.9 & \cellcolor[HTML]{AAC4ED}78.7 & 0.0 & \cellcolor[HTML]{FBFCFE}2.1 & \cellcolor[HTML]{C0D3F2}30.6 & \cellcolor[HTML]{F8FAFE}3.4 & \cellcolor[HTML]{9CBAEA}48.1 & \cellcolor[HTML]{EEF3FB}8.3 & \cellcolor[HTML]{F6F9FD}4.5 & \cellcolor{gray!12}+7.7 \\
Qwen3.5-9B & 11.5 & 22.0 & 70.2 & 46.1 & \cellcolor[HTML]{E8EFFA}15.3 & \cellcolor[HTML]{DCE6F8}27.7 & \cellcolor[HTML]{E99696}53.2 & \cellcolor[HTML]{F8DCDC}40.4 & \cellcolor[HTML]{FBEBEB}62.2 & 0.0 & \cellcolor[HTML]{FAFBFE}2.6 & \cellcolor[HTML]{B8CDF0}34.7 & \cellcolor[HTML]{EDF2FB}9.0 & \cellcolor[HTML]{C2D5F2}29.6 & \cellcolor[HTML]{EFF4FC}7.6 & \cellcolor[HTML]{F8FAFE}3.2 & \cellcolor{gray!12}+3.8 \\
MiniMax-M3 & 6.0 & 12.1 & 97.5 & 54.8 & \cellcolor[HTML]{D2E0F6}13.3 & \cellcolor[HTML]{9EBBEB}27.8 & \cellcolor[HTML]{F7D8D8}91.1 & \cellcolor[HTML]{E2EBF9}59.5 & \cellcolor[HTML]{FFFEFE}97.0 & 0.0 & \cellcolor[HTML]{FEFEFF}0.7 & \cellcolor[HTML]{DCE6F8}17.3 & \cellcolor[HTML]{FAFCFE}2.2 & \cellcolor[HTML]{D3E0F6}21.3 & \cellcolor[HTML]{F8FAFD}3.5 & \cellcolor[HTML]{F8FAFE}3.2 & \cellcolor{gray!12}+7.3 \\
InternVL3.5-30B-A3B & 5.5 & 35.3 & 71.7 & 53.5 & \cellcolor[HTML]{E0EAF8}10.5 & \cellcolor[HTML]{84A9E5}72.7 & \cellcolor[HTML]{E58484}43.7 & \cellcolor[HTML]{E2EBF9}58.2 & \cellcolor[HTML]{F3C6C6}48.5 & 0.0 & \cellcolor[HTML]{FFFFFF}0.2 & \cellcolor[HTML]{D6E2F6}20.1 & \cellcolor[HTML]{FAFCFE}2.2 & \cellcolor[HTML]{E1EAF9}14.8 & \cellcolor[HTML]{F9FBFE}2.8 & \cellcolor[HTML]{FEFEFF}0.6 & \cellcolor{gray!12}+5.0 \\
\bottomrule
\end{tabular}
\end{table}

\paragraph{Only one model reaches a KGR of 50\%, while the next highest reaches 36.0 (Table~\ref{tab:main}).}
\label{sec:results-main} Under the default probe, the highest KGR is 46.6 for Gemini-3.8-Flash, followed by GLM-5.3-Flash at 29.7. Prompted verification improves KGR across most models, but only one model exceeds 50. Gemini-3.8-Flash reaches 55.1, only six models reach 30, and 28 of the 36 models remain below 25. Under the stricter E4 criterion, which requires stating the factual contradiction itself, the corresponding rates are lower (37.2 for Gemini-3.8-Flash, 29.7 for Kimi-K3, and 26.3 for GPT-5.6; Appendix Table~\ref{tab:efour}). Balanced accuracy is much higher, ranging from 49 to 84 among the 18 models with KGR above 19, because a correct ``fake'' verdict counts even when it is not supported by the relevant fact.

\begin{figure}[t]
\centering
\includegraphics[width=\linewidth]{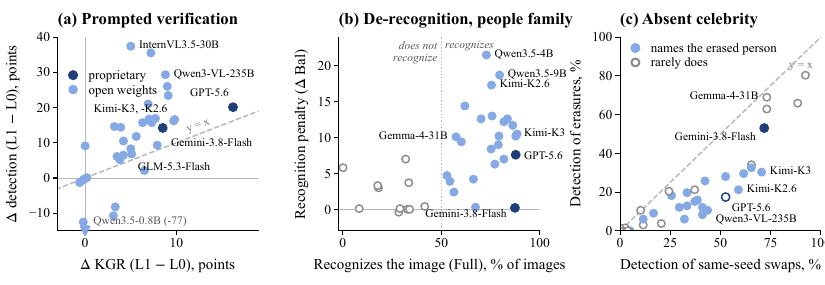}
\caption{Results across all 36 models. (a) Prompting increases detection more than KGR. (points above $y=x$) (b) Recognition penalty is positive for models recognizing at least half of the de-recognition images and near zero otherwise (median +0.4). (c) Models that often name the erased person show larger erase--swap gaps. Wedges mark off-axis points.}
\label{fig:results}
\end{figure}

\paragraph{Prompted verification increases detection more than knowledge grounding (Figure~\ref{fig:results}a).}
For 29 of the 36 models, explicit verification increases KGR by 3.1 to 16.2 points.
Among the 29 models above floor, 26 show a larger increase in forgery detection than in KGR, meaning that many additional ``fake'' verdicts do not meet the entity-or-fact grounding criterion.
Accuracy on authentic images falls for 27 models. As a result, balanced accuracy improves by at most 8.0 points and decreases for 13 models.
Under the adversarial suggestion (L-adv), 17 models still keep their verdict on at least 90\% of authentic images.

\paragraph{Knowledge-anchored edits are detected more often than matched anchor-free controls, especially for category-level facts (Table~\ref{tab:main}, right block).}
\label{sec:results-knowledge}

KGR on the anchor-free control (K0) is only 0.0--2.0 for every model. For 26 of the 36 models, K0 items are flagged less often than authentic images are falsely rejected, and no model's K0 detection rate exceeds its false-alarm rate by more than 7 points.
In contrast, object edits produced by the same editor but tied to a factual contradiction reach a KGR of 79.2.
Among the 19 models with default-probe KGR above 10, every model grounds the class-level families K2 and K4 more often than the people family K1. K1 remains below 10 for all models except Gemini-3.8-Flash at 27.5, even though the three strongest models answer 80.0\% to 83.9\% of the corresponding text-only questions correctly (Appendix Table~\ref{tab:identity}).
This suggests that performance on the people family is limited by whether the inserted face is recognized. The absent-celebrity design instead requires noticing that the canonical person is missing.

\begin{table}[t]
\centering
\caption{People-family analyses. Det, Bal, KGR as in Table~\ref{tab:main}. Left: 145 peripheral forgeries under L1 in the four de-recognition conditions. Right: 132 absent-celebrity items under the default probe L0. Named denotes answers naming the erased person; Real denotes naming that person while judging the image authentic. Colors show changes from Full (Erase from Swap); bold Det indicates McNemar $p<0.05$. Cap. = Det(Full+Caption)$-$Det(Full); Pen. = Bal(Crop+Caption)$-$Bal(Full). $^{\P}$Recognizes fewer than half of the images. All models: Appendix Table~\ref{tab:people-all}.}
\label{tab:conditions}

\scriptsize\renewcommand{\arraystretch}{0.96}
\setlength{\tabcolsep}{1.2pt}
\begin{tabular}{@{}l rrr rrr rrr rrr rrrrr rr@{}}
\toprule
 & \multicolumn{3}{c}{Full} & \multicolumn{3}{c}{Full+Caption} & \multicolumn{3}{c}{Crop} & \multicolumn{3}{c}{Crop+Caption} & \multicolumn{5}{c}{Absent-celebrity (L0)} & \multicolumn{2}{c}{\cellcolor{gray!12}Derived} \\
\cmidrule(lr){2-4}\cmidrule(lr){5-7}\cmidrule(lr){8-10}\cmidrule(lr){11-13}\cmidrule(lr){14-18}\cmidrule(lr){19-20}
Model & Det & Bal & KGR & Det & Bal & KGR & Det & Bal & KGR & Det & Bal & KGR & Erase & Swap & K0 & Named & \&\,Real & \cellcolor{gray!12}Cap. & \cellcolor{gray!12}Pen. \\
\midrule
\multicolumn{20}{@{}l}{\emph{Proprietary}} \\
Gemini-3.8-Flash & 73.1 & 78.9 & 35.2 & \cellcolor[HTML]{E7EEFA}77.9 & \cellcolor[HTML]{FBEDED}75.2 & \cellcolor[HTML]{FDF5F5}33.1 & \cellcolor[HTML]{E0EAF9}79.3 & \cellcolor[HTML]{FCF3F3}76.4 & \cellcolor[HTML]{C9D9F4}46.2 & \cellcolor[HTML]{BFD2F1}\textbf{86.2} & \cellcolor[HTML]{FEFEFF}79.1 & \cellcolor[HTML]{D0DEF5}44.8 & \cellcolor[HTML]{F3C5C5}53.0 & 71.8 & 2.4 & 74.2 & 32.6 & \cellcolor{gray!12}+4.8 & \cellcolor{gray!12}+0.2 \\
GPT-5.6 & 53.8 & 69.2 & 14.5 & \cellcolor[HTML]{EEF3FB}57.2 & \cellcolor[HTML]{F4F7FD}71.5 & \cellcolor[HTML]{FEF8F8}13.1 & \cellcolor[HTML]{F1F5FC}56.6 & \cellcolor[HTML]{FDF6F6}67.3 & \cellcolor[HTML]{FEF8F8}13.1 & \cellcolor[HTML]{B1C8EE}\textbf{69.7} & \cellcolor[HTML]{DAE5F7}76.8 & \cellcolor[HTML]{F5F8FD}16.6 & \cellcolor[HTML]{E89393}17.4 & 52.6 & 2.4 & 23.5 & 19.7 & \cellcolor{gray!12}+3.4 & \cellcolor{gray!12}+7.6 \\
\addlinespace[2pt]
\multicolumn{20}{@{}l}{\emph{Open weights}} \\
GLM-5.3-Flash & 52.4 & 67.5 & 7.6 & \cellcolor[HTML]{FAE7E7}47.6 & \cellcolor[HTML]{F9FBFE}68.7 & \cellcolor[HTML]{FCF1F1}4.8 & \cellcolor[HTML]{CCDBF4}62.8 & \cellcolor[HTML]{F9FBFE}68.8 & \cellcolor[HTML]{DDE7F8}14.5 & \cellcolor[HTML]{DDE7F8}59.3 & \cellcolor[HTML]{DDE7F8}74.5 & \cellcolor[HTML]{F2F6FC}10.3 & \cellcolor[HTML]{EA9A9A}32.6 & 65.4 & 9.8 & 52.3 & 37.1 & \cellcolor{gray!12}$-$4.8 & \cellcolor{gray!12}+7.0 \\
Kimi-K3 & 35.9 & 64.4 & 3.4 & \cellcolor[HTML]{F6D3D3}\textbf{26.9} & \cellcolor[HTML]{FCF0F0}61.4 & \cellcolor[HTML]{FEF9F9}2.1 & \cellcolor[HTML]{84A9E5}\textbf{82.8} & \cellcolor[HTML]{E4ECF9}69.9 & \cellcolor[HTML]{D3E0F6}12.4 & \cellcolor[HTML]{84A9E5}\textbf{62.8} & \cellcolor[HTML]{CDDCF4}74.6 & \cellcolor[HTML]{E7EEFA}8.3 & \cellcolor[HTML]{E58484}30.3 & 70.5 & 2.4 & 52.3 & 37.1 & \cellcolor{gray!12}$-$9.0 & \cellcolor{gray!12}+10.2 \\
Kimi-K2.6 & 28.3 & 59.0 & 4.1 & \cellcolor[HTML]{F8FAFE}29.7 & \cellcolor[HTML]{FFFEFE}58.7 & \cellcolor[HTML]{FEF9F9}2.8 & \cellcolor[HTML]{84A9E5}\textbf{67.6} & \cellcolor[HTML]{C2D5F2}71.3 & \cellcolor[HTML]{B1C8EE}20.0 & \cellcolor[HTML]{84A9E5}\textbf{63.4} & \cellcolor[HTML]{AAC3ED}76.3 & \cellcolor[HTML]{C8D9F3}15.2 & \cellcolor[HTML]{E68B8B}21.2 & 59.0 & 4.9 & 56.8 & 44.7 & \cellcolor{gray!12}+1.4 & \cellcolor{gray!12}+17.3 \\
Qwen3.5-122B-A10B & 31.7 & 51.1 & 3.4 & \cellcolor[HTML]{FEFCFC}31.0 & \cellcolor[HTML]{FCFDFE}51.7 & 3.4 & \cellcolor[HTML]{A3BFEC}\textbf{50.3} & \cellcolor[HTML]{FCF3F3}48.6 & \cellcolor[HTML]{EEF3FB}6.9 & \cellcolor[HTML]{A6C1EC}\textbf{49.7} & \cellcolor[HTML]{CBDBF4}61.6 & \cellcolor[HTML]{DDE7F8}10.3 & \cellcolor[HTML]{EA9D9D}29.5 & 61.5 & 19.5 & 61.4 & 43.2 & \cellcolor{gray!12}$-$0.7 & \cellcolor{gray!12}+10.5 \\
Qwen3-VL-235B-A22B & 48.3 & 59.3 & 4.1 & \cellcolor[HTML]{EEB1B1}\textbf{32.4} & \cellcolor[HTML]{FEFCFC}58.6 & \cellcolor[HTML]{FEFCFC}3.4 & \cellcolor[HTML]{84A9E5}\textbf{89.0} & \cellcolor[HTML]{FCFDFE}59.9 & \cellcolor[HTML]{DDE7F8}11.0 & \cellcolor[HTML]{DAE5F7}55.9 & \cellcolor[HTML]{ECF2FB}63.2 & \cellcolor[HTML]{E0EAF9}10.3 & \cellcolor[HTML]{EA9A9A}10.6 & 43.6 & 4.9 & 58.3 & 53.8 & \cellcolor{gray!12}$-$15.9 & \cellcolor{gray!12}+3.9 \\
Qwen3.6-27B & 40.7 & 48.4 & 2.8 & \cellcolor[HTML]{F6D3D3}\textbf{31.7} & \cellcolor[HTML]{FFFFFF}48.5 & \cellcolor[HTML]{FEF8F8}1.4 & \cellcolor[HTML]{D3E0F6}49.7 & \cellcolor[HTML]{F5CECE}38.5 & \cellcolor[HTML]{EBF1FB}6.9 & \cellcolor[HTML]{C9D9F4}\textbf{51.7} & \cellcolor[HTML]{CDDCF4}58.6 & \cellcolor[HTML]{F2F6FC}5.5 & \cellcolor[HTML]{EFB4B4}28.0 & 52.6 & 14.6 & 44.7 & 30.3 & \cellcolor{gray!12}$-$9.0 & \cellcolor{gray!12}+10.2 \\
Qwen3.5-27B & 32.4 & 42.2 & 1.4 & \cellcolor[HTML]{FDF5F5}30.3 & \cellcolor[HTML]{F5F8FD}44.3 & \cellcolor[HTML]{FCFDFE}2.1 & \cellcolor[HTML]{CCDBF4}42.8 & \cellcolor[HTML]{FDF5F5}40.1 & \cellcolor[HTML]{EEF3FB}4.8 & \cellcolor[HTML]{D3E0F6}41.4 & \cellcolor[HTML]{C1D4F2}54.8 & \cellcolor[HTML]{EEF3FB}4.8 & \cellcolor[HTML]{F1BBBB}15.2 & 37.2 & 12.2 & 40.2 & 33.3 & \cellcolor{gray!12}$-$2.1 & \cellcolor{gray!12}+12.6 \\
Qwen3.8-27B$^{\dagger}$ & 54.5 & 60.4 & 1.4 & \cellcolor[HTML]{FCF1F1}51.7 & \cellcolor[HTML]{F9E4E4}54.9 & \cellcolor[HTML]{FEFCFC}0.7 & \cellcolor[HTML]{BFD2F1}\textbf{67.6} & \cellcolor[HTML]{F6F9FD}62.2 & \cellcolor[HTML]{EEF3FB}4.8 & \cellcolor[HTML]{AAC4ED}\textbf{71.7} & \cellcolor[HTML]{D1DFF5}69.8 & \cellcolor[HTML]{F5F8FD}3.4 & \cellcolor[HTML]{F4CCCC}25.8 & 42.3 & 17.1 & 30.3 & 24.2 & \cellcolor{gray!12}$-$2.8 & \cellcolor{gray!12}+9.4 \\
Qwen3.6-35B-A3B & 22.1 & 38.1 & 1.4 & \cellcolor[HTML]{EEF3FB}25.5 & \cellcolor[HTML]{CEDDF5}48.0 & \cellcolor[HTML]{F8FAFE}2.8 & \cellcolor[HTML]{B8CDF0}\textbf{36.6} & \cellcolor[HTML]{FEFAFA}37.0 & \cellcolor[HTML]{F8FAFE}2.8 & \cellcolor[HTML]{AAC4ED}\textbf{39.3} & \cellcolor[HTML]{B8CDF0}52.5 & \cellcolor[HTML]{DAE5F7}9.0 & \cellcolor[HTML]{F0BABA}15.9 & 38.5 & 7.3 & 34.1 & 31.1 & \cellcolor{gray!12}+3.4 & \cellcolor{gray!12}+14.4 \\
Gemma-4-31B & 77.2 & 53.9 & 1.4 & \cellcolor[HTML]{E68888}\textbf{53.1} & \cellcolor[HTML]{FEFBFB}53.1 & 1.4 & \cellcolor[HTML]{9DBAEA}\textbf{97.2} & 53.9 & \cellcolor[HTML]{E0EAF9}7.6 & \cellcolor[HTML]{CFDEF5}\textbf{86.9} & \cellcolor[HTML]{CDDCF4}64.0 & \cellcolor[HTML]{DDE7F8}8.3 & \cellcolor[HTML]{FCF2F2}68.9 & 73.1 & 58.5 & 15.9 & 11.4 & \cellcolor{gray!12}$-$24.1 & \cellcolor{gray!12}+10.1 \\
Qwen3.5-9B & 12.5 & 27.2 & 2.1 & \cellcolor[HTML]{F9FBFE}13.8 & \cellcolor[HTML]{C0D3F2}40.1 & 2.1 & \cellcolor[HTML]{FFFFFF}12.4 & \cellcolor[HTML]{F6D4D4}18.5 & \cellcolor[HTML]{FEF8F8}0.7 & \cellcolor[HTML]{AEC6EE}\textbf{29.0} & \cellcolor[HTML]{A3BFEC}45.9 & \cellcolor[HTML]{F2F6FC}4.8 & \cellcolor[HTML]{FAE8E8}9.1 & 16.7 & 9.8 & 29.5 & 25.0 & \cellcolor{gray!12}+1.3 & \cellcolor{gray!12}+18.7 \\
MiniMax-M3 & 14.5 & 52.6 & 1.4 & \cellcolor[HTML]{FCF1F1}11.7 & \cellcolor[HTML]{FEFEFF}52.8 & \cellcolor[HTML]{FEFCFC}0.7 & \cellcolor[HTML]{DDE7F8}21.4 & \cellcolor[HTML]{FDF4F4}50.3 & \cellcolor[HTML]{F8FAFE}2.8 & \cellcolor[HTML]{B4CBEF}\textbf{29.7} & \cellcolor[HTML]{D6E2F6}61.0 & \cellcolor[HTML]{F8FAFE}2.8 & \cellcolor[HTML]{FBEEEE}6.1 & 11.5 & 2.4 & 36.4 & 34.1 & \cellcolor{gray!12}$-$2.8 & \cellcolor{gray!12}+8.4 \\
InternVL3.5-30B-A3B$^{\P}$ & 70.3 & 51.0 & 0.0 & \cellcolor[HTML]{E58484}\textbf{20.0} & \cellcolor[HTML]{FFFEFE}50.8 & 0.0 & \cellcolor[HTML]{88ACE6}\textbf{94.5} & \cellcolor[HTML]{FFFEFE}50.7 & 0.0 & \cellcolor[HTML]{F7DADA}62.8 & \cellcolor[HTML]{DDE7F8}58.0 & 0.0 & \cellcolor[HTML]{F5CECE}21.2 & 37.2 & 17.1 & 6.1 & 3.8 & \cellcolor{gray!12}$-$50.3 & \cellcolor{gray!12}+7.0 \\
\bottomrule
\end{tabular}
\end{table}

\paragraph{Knowledge-grounded detection improves when scene recognition is reduced and the event context is restored (Table~\ref{tab:conditions}, Figure~\ref{fig:results}b).}
\label{sec:results-recognition}

Crop substantially reduces self-reported recognition. For every model that recognizes at least half of these images under Full, recognition falls by at least 30.9 points. Mirror keeps both recognition and detection within 11.2 points of Full.
Among the 21 recognizing models that detect at least 10\% of the forgeries under Full, the recognition penalty is positive for every model. KGR increases under Crop+Caption for all 21, and detection increases significantly for 17.
Gemini-3.8-Flash has the smallest penalty at +0.2. Its gains in detection and KGR under Crop+Caption are offset by a similar drop in accuracy on authentic crops.
The 11 models that recognize fewer than half of these images show almost no recognition penalty, with a median of +0.4, and their KGR remains at or near zero after cropping even though they receive the same crop.
The same pattern also appears within the Full condition, without experimentally changing recognizability. Across recognizing models, the odds of detecting a people-family forgery in a recognized scene are 0.65 times those in an unrecognized scene (Appendix Table~\ref{tab:itemreg}).

\paragraph{On people-family peripheral forgeries, providing the true event context does not improve knowledge-grounded detection.}

The caption gives the event and date, which are the two pieces of context used in an E4 explanation.
Under Full+Caption, no model improves its KGR on more than 2 of the 145 forgeries. Detection decreases for 15 of the 21 models, significantly for six, and does not increase significantly for any model.
Crop+Caption uses the same sentence and preserves the same manipulated pixels. Yet all 21 models ground more detections when cropping reduces scene recognition, with KGR under Crop+Caption reaching 1.3 to 8.9 times its value under Full+Caption.
The wrong-caption control shows a different failure mode. When the sentence names another event from the same era, Gemini-3.8-Flash and GPT-5.6 label 93.8\% and 91.0\% of the forgeries as fake, but they also label 80.6\% and 83.9\% of authentic images as fake. KGR against the recorded contradiction does not improve (Appendix Table~\ref{tab:wrongcaption}). This pattern suggests that the caption is being compared with the remembered scene rather than used to verify the edited pixels.

\paragraph{Models often report an erased person, but detect the edit more reliably when another public figure is inserted (Table~\ref{tab:conditions}, right block, Figure~\ref{fig:results}c).}

When asked only whether the image is authentic, Gemini-3.8-Flash names the erased public figure in 74.2\% of its answers. Across all 36 models, 18 name the erased person and still call the image authentic in the same answer on at least a quarter of the items.
When asked only who is in the image, six models name the erased person on at least half of the items, with the highest rate reaching 72.7\%.
Because the erased person is no longer visible, the name must come from the model's learned representation of the canonical photograph (Figure~\ref{fig:teaser}).
On the same seeds, Kimi-K3 detects 70.5\% of celebrity swaps but only 30.3\% of erasures, while detection on anchor-free K0 edits is 2.4\%.
Both swaps and erasures leave a face that does not belong to the canonical scene, but only a swap provides a second recognizable identity that can be checked.
The gap is concentrated among models that identify the scene. The median erase-to-swap detection ratio is 0.41 for models that name the erased person in at least a quarter of their answers, compared with 0.80 for models that rarely do so.
This difference remains after controlling for edit size, image domain, and pixel-detector scores (Appendix~\ref{app:footprint}).

\paragraph{Models can distinguish the replacement face in isolation yet still accept the full forgery.}
When shown the face alone, Kimi-K3 distinguishes the replacement from the original on 55 of 119 items, and Qwen3.6-27B does so on 47.
On those same items, the models still judge the full photograph authentic 63.6\% and 69.8\% of the time, while naming the erased person in 30.9\% and 30.2\% of their answers. Gemini-3.8-Flash distinguishes the two faces on 81 items but still calls the full image authentic on 29.6\% of them. These misses therefore cannot be explained simply by missing factual knowledge or by a failure to perceive the face difference.

\section{Analysis}
\label{sec:analysis}

\paragraph{Identity recognition limits the use of person-specific knowledge.}

Models use category-level knowledge more reliably than knowledge about specific individuals.
Detecting a wrongly inserted person (K1) first requires recognizing that person, whereas category-level contradictions such as an anachronistic object (K2) can be identified from general knowledge.
Accordingly, Gemini-3.8-Flash reaches a K1 KGR of 27.5, while every other model remains below 10.
Yet the absent-celebrity results show that the relevant knowledge is often present but not used.
Kimi-K3 names the erased person in 52.3\% of its answers, while Qwen3.5-122B-A10B and Qwen3-VL-235B-A22B often name that person and still judge the image authentic in 43.2\% and 53.8\% of cases, respectively (Table~\ref{tab:conditions}).
For class-level facts, the same true caption does not reduce detection and improves it for two families (Appendix Table~\ref{tab:families}).

\paragraph{The recognition effect persists with scale and additional reasoning.}
Among the 21 models that recognize at least half of these images and detect at least 10\% of the forgeries under Full, the recognition penalty ranges from +0.2 to +18.7 points, and it is near zero when there is little recognition to remove (Figure~\ref{fig:results}b).
Absent-celebrity errors also occur on full images whose replacement faces can be distinguished in isolation, so the effect is not simply caused by cropping.
Gemini-3.8-Flash is the least affected: it names the erased person in 74.2\% of its answers, detects 53.0\% of the erasures, and has the smallest recognition penalty (+0.2). Larger models achieve higher KGR but do not eliminate the recognition penalty: Qwen3.5-122B reaches 31.5 KGR while retaining a +10.5 recognition penalty, and thinking variants likewise retain positive penalties.
Together with the elicitation gap in \S\ref{sec:results-main}, these results suggest that a key bottleneck is whether verification is invoked after recognition.
Models may possess the relevant fact and can perceive the visual difference, yet their final judgments don't consistently reflect that information once the scene is recognized.

\paragraph{Implications.}

The problem is most relevant for recognizable, correctly captioned people-family forgeries, which closely match how such images may circulate.
In this setting, no model grounds more than one third of the forgeries, and adding the true caption lowers detection more often than it raises it.
A model that accepts a forged canonical image while describing the remembered original can make the forgery appear more credible.
Simply suppressing the prior is not sufficient, however.
Anchor-free edits produced by the same editor are rarely detected, meaning that memory of the canonical image can both hide the forgery and provide the knowledge needed to expose it.
More broadly, reliably reconciling visual observations with prior knowledge is part of the challenge of building trustworthy AI systems whose perceptual capabilities remain aligned with their safety requirements \citep{tan2025trustworthyembodied}.

\section{Conclusion}
\label{sec:conclusion}
We introduced Mandela-Bench, a single-image forgery benchmark in which the only reliable evidence of manipulation is a factual contradiction about what the image depicts.
Across 36 multimodal models, our results show that recognizing a canonical image can shift verification toward reconstruction from memory.
Models often name people who have been erased, and some judge a photograph authentic despite distinguishing the replacement face in isolation. With the same true caption, removing the familiar composition yields more knowledge-grounded forgery verdicts than showing the full image.
Even under explicit verification prompts, only one of the 36 models produces explanations meeting the KGR criterion for at least half of the forged images.
These failures therefore cannot be explained simply by missing knowledge or poor visual perception.
Instead, the results are consistent with recognition biasing verification toward the remembered canonical image, suggesting that the main bottleneck may be whether verification is invoked after recognition.

\newpage

\subsection*{AI use statement}
The benchmark images are synthetic data, rendered by a commercial image editor (GPT Image 2) from structured edit specifications; candidate identities were nominated by a multimodal model (Kimi-K3) and verified against Wikidata; every rendered candidate passed an automated quality check by a multimodal model before the manual review of Appendix~\ref{app:construction}; the explanations of the evaluated models were graded on the evidence ladder by an LLM judge (Qwen3.8-27B), calibrated against 206 human-graded answers (Appendix~\ref{app:judge}).
Beyond the benchmark itself, AI assistants were used to search the literature, as coding assistants in implementing and running the evaluation pipeline, and to polish the writing.
The authors reviewed all AI-assisted material and take responsibility for the content of the paper.

\subsection*{Ethics statement}
The benchmark depicts public figures only.
Every inserted identity is a public figure verified against Wikidata; living heads of state or government and contemporary living singers and actors were excluded from insertion, and the absent-celebrity items replace the public figure with a synthetic face, so no private individual appears in any item.
Source images come from Wikimedia Commons and from open-access museum, film-still, and meme-template archives, and the release records the source of every item; politically charged meme templates were excluded, and the historical retouching cases of \S\ref{sec:intro} are described without judgement of the people involved.
The only human labelling was done by the authors (the manual review of items and the 206 judge-calibration answers), so no human-subject study was involved.
The forgeries are released as labelled items, each with its recorded contradiction, for detection research.
They were produced with a commercial editor from one-sentence instructions, so the release adds no editing capability that is not already public; what it adds is a measurement of the failure mode that currently lets such forgeries pass.

\subsection*{Reproducibility statement}
The release set (1,507 forged images, 148 of them anchor-free controls, and 474 authentic images, each with its knowledge family, source, edit specification, recorded contradiction, and footprint), the verbatim prompts of every probe and condition (Appendix~\ref{app:prompts}), the judge rubric with its calibration data (Appendix~\ref{app:judge}), and the evaluated models with their references (Appendix~\ref{app:models}; temperature 0 throughout) are described in the appendices and are released with the code.
Construction and quality control, including the pixel-detector audit and the metadata-leakage check, are given in Appendix~\ref{app:construction}; the formal definition of a knowledge-only forgery in Appendix~\ref{app:definition}.

\bibliography{references}
\bibliographystyle{iclr2027_conference}

\clearpage
\appendix
\begin{center}
{\Large\bfseries Appendix}\\[0.35em]
{\large Mandela-Bench: Multimodal Models Remember Canonical Images Instead of Seeing Them}
\end{center}
\vspace{0.5em}
\noindent
Appendix~\ref{app:related} places the benchmark among prior work.
Appendices~\ref{app:definition}--\ref{app:example} cover the benchmark itself: the formal definition of a knowledge-only forgery, construction and quality-control details, and a worked example.
Appendices~\ref{app:prompts}--\ref{app:models} give the evaluation instruments: the verbatim prompts, the judge rubric with its calibration, and the evaluated models with their settings.
Appendices~\ref{app:results}--\ref{app:face} report the remaining results: the full-roster tables and the recognition and identity probes, the item-level association between recognition and detection, the caption controls, validity checks, the footprint regression, the preregistered decision rule, and the face probe.
Appendix~\ref{app:analysis} collects analyses that support Section~\ref{sec:analysis}.
\vspace{0.5em}

\section{Extended Related Work}
\label{app:related}

\paragraph{Detecting the generator's imprint.}
Detectors of generated images learn the regularities a generator leaves in pixel statistics or the spectrum \citep{xu2025aigisurvey}.
Examples are nearest neighbours in a frozen CLIP space \citep{ojha2023ufd} and the regularity that up-sampling leaves \citep{tan2024npr}.
Chameleon collects generated images that human annotators passed as real; nine off-the-shelf detectors mostly label them real \citep{yan2024chameleon}.
Open-source image detectors lose 45\% of their AUC on deepfakes that circulated in 2024 \citep{chandra2025deepfakeeval}.
ITW-SM traces the collapse on social-media images to detector design choices \citep{konstantinidou2025itwsm}; NTIRE 2026 scores robustness to 36 transformations across 42 generators \citep{gushchin2026ntire}.
DailyBench adds object-level edits, on which detectors at 91--96\% balanced accuracy on GenImage fall to 54--66\% \citep{jiang2026dailybench}; UniAIDet adds instruction-guided editing and artistic images \citep{zhang2025uniaidet}.
OpenFake contributes three million captioned real images and a crowdsourced adversarial platform \citep{livernoche2025openfake}, while Celeb-DF++ covers 22 face-manipulation methods \citep{li2025celebdfpp}; paired adversarially edited data have also been used to train reasoning-based detectors \citep{bao2026spared}. Across these settings, the evidence still comes from generator or manipulation traces in the pixels.

\paragraph{Evidence in the image or in its caption.}
DGM4 swaps or edits faces and text in news image--text pairs and grounds the altered boxes and tokens, reading the broken cross-modal correlation \citep{shao2023dgm4}.
MiRAGeNews pairs Midjourney images with misleading GPT-4 captions and detects them with fused image and text branches \citep{huang2024miragenews}.
AnomReason annotates generated scenes with content-level anomalies such as physical violations and commonsense errors \citep{tan2025anomreason}.
Semantic Manipulation Localization targets meaning-altering edits to an object's attributes or relations \citep{tan2026sml}; Impostor localizes regions that are semantically plausible yet forensically disrupted \citep{li2026impostor}.
ManipShield adds boxes, judgment cues and explanations for 25 editors \citep{xu2025manipshield}; EditSleuth derives reasoning chains deterministically from the source image and edit mask \citep{nguyen2026editsleuth}.
These methods rely on a mismatch between supplied modalities, a visible violation, or forensic residue around the edit. Mandela-Bench preserves such editing residue in its anchor-free controls while removing the factual contradiction.

\paragraph{Reference-grounded verification.}
LookupForensics splits image-based fact verification into forgery identification and fact retrieval, returning the original image from a reference database with the verdict \citep{cui2024lookupforensics}.
REVEAL scores a query by its differences from authentic news image--text pairs retrieved from a 170K-item library covering 40K public figures \citep{zhou2026reveal}.
Tool-augmented multimodal agents can also gather external evidence through multimodal search \citep{gao2026videosearcher}.
Swapping the library adapts it without training.
Both hold the knowledge in an external memory the system is given; their metrics measure retrieval and comparison.
Mandela-Bench withholds the library and makes the detector's own knowledge of the depicted referent the only evidence.

\paragraph{Multimodal models as detectors and explainers.}
LOKI poses judgment, anomaly-selection and explanation questions over synthetic video, images, 3D assets, text and audio \citep{ye2024loki}.
FakeBench pairs images from ten generator types with human descriptions of forgery cues under a human-validated taxonomy \citep{li2024fakebench}.
DeepfakeJudge scores how faithfully a rationale follows the visual evidence, citing vision-language models' over-reliance on textual priors and world knowledge \citep{kuckreja2026deepfakejudge}.
These benchmarks ask whether a model can identify cues left by generation or editing. DeepfakeJudge additionally treats unsupported world knowledge as a source of ungrounded explanations. Mandela-Bench reverses this role: KGR rewards a verdict only when the model uses relevant knowledge to identify the contradiction.

\paragraph{Priors against perception inside multimodal models.}
HallusionBench defines language hallucination as a perception formed without relevant visual input \citep{guan2024hallusionbench}.
Its Visual Supplement questions concern charts, tables and maps, many edited by human experts; models holding the answer in parametric memory prioritize it over the edited figure.
Pixels Versus Priors builds attribute counterfacts; by the late layers the image outweighs the prior, and steering vectors shift the balance \citep{golovanevsky-etal-2025-pixels}.
MLLMs Know Where to Look shows accuracy falling with the size of the visual subject while attention lands on the right region even when the answer is wrong \citep{zhang2025mllmsknow}.
Mandela-Bench differs in three respects. It uses real canonical images with pixel-clean, knowledge-only edits. Its de-recognition conditions manipulate recognizability while keeping the manipulated region fixed, and the recognition penalty measures the resulting change in balanced accuracy. Perception and fact possession are also tested on the same items.

\paragraph{Human memory and canonical images.}
\citet{groh2022deepfake} find participants and the leading detector similarly accurate on unknown speakers.
On videos of Putin and Kim Jong-un the participants pull ahead, which the authors attribute to reasoning about what a well-known figure would say.
\citet{prasad2022vme} document the visual Mandela effect: shared, specific false memories of familiar images.
\citet{bates2025kymkb} build a knowledge base of more than 5,200 meme templates and match memes to them by distance-based lookup.
Templates are reused images the viewer must recognize, canonical images in our sense; they form one domain of Mandela-Bench.
The name borrows the human phenomenology and claims no shared mechanism.

\section{Formal definition of a knowledge-only forgery}
\label{app:definition}
Let $s$ be a canonical image with knowledge-base entry $k(s)$ (a Wikidata identifier with the event, its date and place, and the people and objects recorded as present), and let $\mathcal{F}(s)$ be the set of propositions about the depicted referent that $k(s)$ makes checkable.
An item is a triple $(s, e, \varphi)$: an edit $e$ confined to a region $R$ of $s$, producing $\hat{s} = e(s)$, and a proposition $\varphi \in \mathcal{F}(s)$.
The item is a \emph{knowledge-only forgery} when the five properties of \S\ref{sec:forgeries} hold.
\begin{enumerate}[leftmargin=*, itemsep=1pt]
\item \textbf{Anchored.} $k(s)$ exists, and every element inserted by $e$ has its own knowledge-base entry. This includes people with birth and death dates, objects with introduction dates, places, text, event attributes, and the content of a work. The relevant proposition $\varphi$ can therefore be verified by machine.
\item \textbf{Single contradiction.} $\varphi$ is true of $s$ and false of $\hat{s}$, while every other proposition in $\mathcal{F}(s)$ keeps the same truth value. We store $\varphi$ with the item as ``$\langle$inserted entity$\rangle$ $\langle$relation that fails$\rangle$ $\langle$event, date$\rangle$''. The type of $\varphi$ defines the knowledge family: people (K1), objects (K2), place (K3), visible text (K4), event attributes (K5), or the content of a work of art or film (K6).
\item \textbf{Seamless.} $e$ matches the era, medium, and style of $s$: the inserted element is rendered in the period dress, film stock, grain, colour, and lighting of the original, so that no cue of the edit is available from the appearance of $R$ relative to $s \setminus R$.
\item \textbf{Intrinsically plausible.} $\hat{s}$ violates no physical, perceptual, or commonsense regularity: a viewer who does not hold $\varphi$ finds the scene unremarkable. Candidates whose contradiction is visible without the fact (a scale error, an impossible shadow) are rejected at quality control.
\item \textbf{Pixel-clean.} Pixel cleanliness is measured after construction. Two pixel detectors based on different principles score $\hat{s}$ near chance against the authentic images (\S\ref{sec:qc}), and a classifier using file-level metadata performs no better (Appendix~\ref{app:construction}).
\end{enumerate}
The \emph{anchor-free control} (K0) is $(s,e',\bot)$: it uses the same seed and class of edit, but the inserted element carries no proposition in $\mathcal{F}(s)$, so no recorded fact changes truth value. The \emph{authentic item} is the unchanged seed $s$. On matched seeds, the difference between a knowledge family and K0 isolates the contribution of the factual contradiction, while the difference between an edited item and its authentic counterpart captures the contribution of the edit itself.

K1 contains three contradiction types: \emph{date-impossible}, \emph{cross-domain}, and \emph{absent-celebrity} (\S\ref{sec:controls}). Label semantics follow the medium: in photographic domains, ``this did not happen'' and ``this image was altered'' coincide, whereas paintings are labelled relative to the canonical original.

\begin{figure}[t]
\centering
\includegraphics[width=\linewidth]{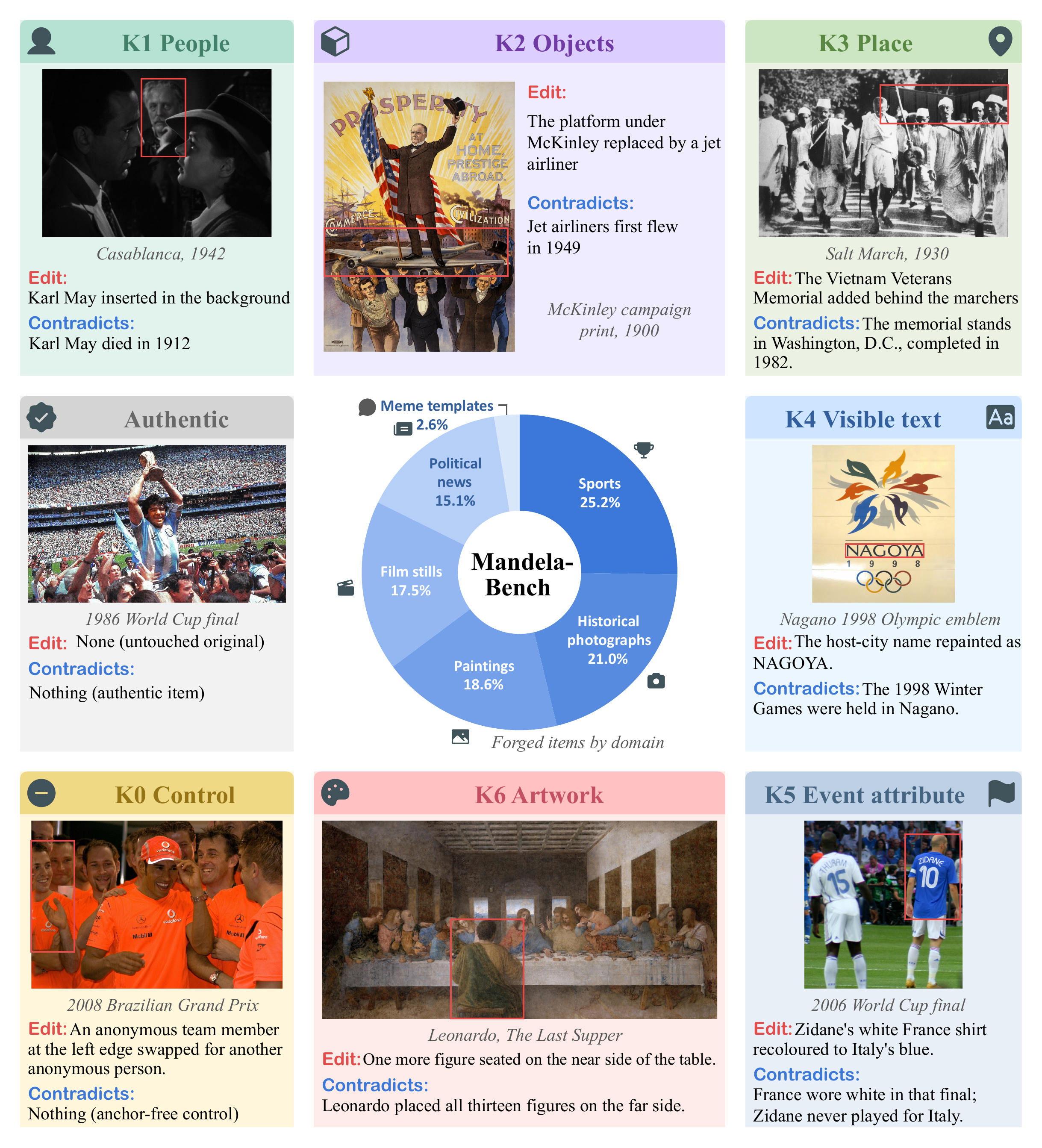}
\caption{Mandela-Bench examples for each knowledge family, an anchor-free control (K0), and an untouched authentic image. Center: distribution of 1,507 forged items across six image domains.}
\label{fig:overview-full}
\end{figure}

\section{Construction details}
\label{app:construction}

Seeds come from Wikimedia Commons (1,555 of the 1,981 release items), open-access film and museum collections (film-grab.com 320; Metropolitan Museum of Art 55), and meme archives. Each seed has a Wikidata identifier and an annotated composition (single figure, small group, or crowd), which determines where peripheral edits are possible. Before evaluation, images are re-encoded at JPEG quality 92 with the longest side set to 1280\,px.

Each edit specification is rendered once initially. A new identity or a fresh render is generated only when the editor refuses the request or the candidate fails quality control, with at most two to four candidates per specification. The best valid candidate is retained. Automated checks verify named insertions with InsightFace \texttt{buffalo\_l} (ArcFace embeddings with RetinaFace detection). Of the 435 named insertions, 398 (91.5\%) are compared with independently retrieved portraits and 37 with the portrait supplied to the editor. Candidates below cosine similarity 0.35 are rejected (0.5 in the earliest runs). A usable reference face is detected for 265 insertions (60.9\%); the remaining cases rely on the recognition probe in \S\ref{sec:qc}. For absent-celebrity items, the check is reversed: the synthetic replacement must have similarity below 0.30 to the erased person, and none of the 132 retained replacements carries an identity. Inserted objects are verified with a grounding check, and candidates with visible low-level artefacts are rejected.

Manual review then checks whether the edit reached its intended target and whether any visible artefact reveals the manipulation. It includes a blind spot-check of 99 items from the largest run (87 passed; the 12 failures and all implicated items were removed), a full review of 194 absent-celebrity items (35 rejected), and audits of 61 duplicate-candidate or small-edit cases. We also test for non-semantic leakage. A classifier using only file-level features reaches AUC 0.56 on forged versus authentic images; individual metadata features remain at AUC 0.50, with the residual signal coming from bytes per pixel (0.55) and width (0.54) after re-encoding. On photographic domains, NPR and UniversalFakeDetect reach AUC 0.543 and 0.536, compared with 0.577 and 0.538 on the full corpus. K6 is the only family above 0.61 (NPR 0.749). All absent-celebrity replacement faces are synthetic, so no real private individual appears in the benchmark.

\section{A worked example}
\label{app:example}

\begin{figure}[h]
\centering
\includegraphics[width=\linewidth]{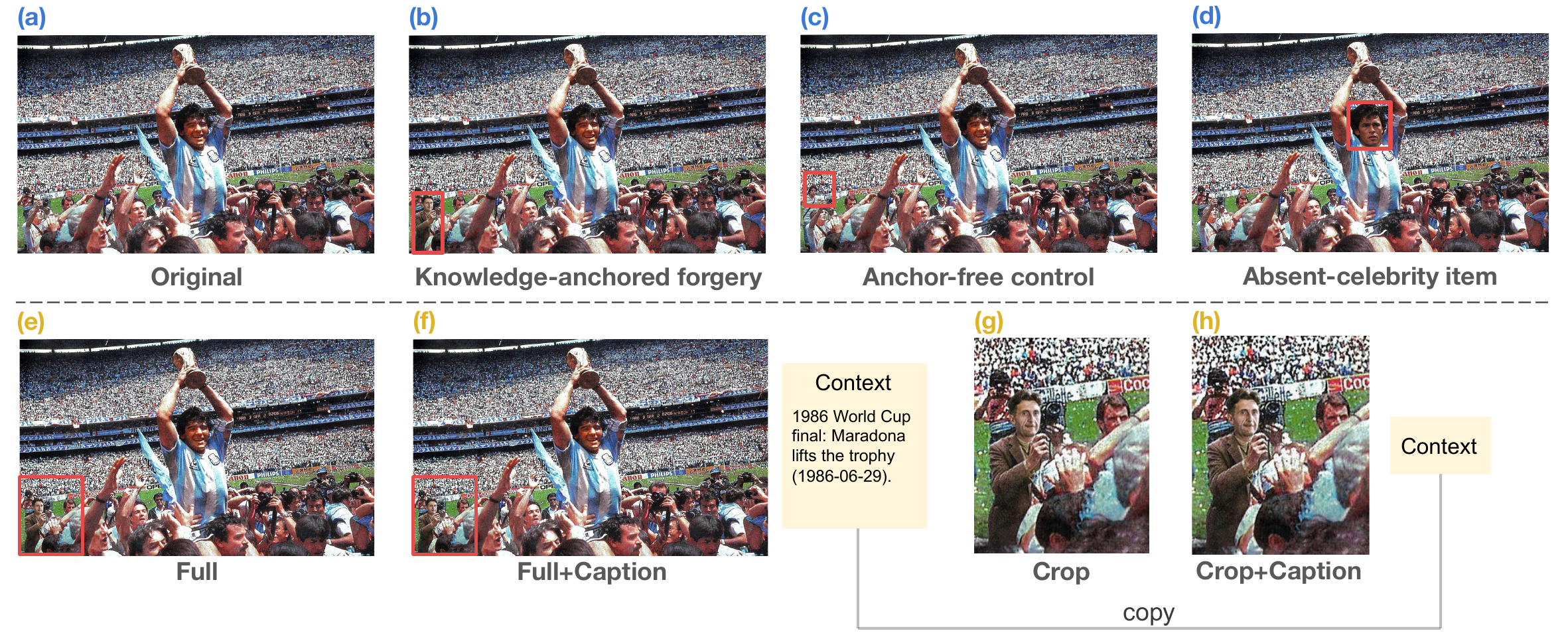}
\caption{One seed and its derived benchmark items (top), and the de-recognition conditions applied to the knowledge-anchored forgery (bottom). Top: (a) authentic original; (b) knowledge-anchored forgery with George Orwell inserted, for which the ideal verdict is fake, grounded in the recorded factual contradiction; (c) matched anchor-free control (K0) with an unnamed face inserted at the same position by the same editor; (d) absent-celebrity item with Maradona replaced by a synthetic face. Bottom: (e) Full; (f) Full+Caption; (g) Crop; (h) Crop+Caption. The crop preserves the edited region while removing much of the recognizable composition, and the box marks the crop region. (f) and (h) receive the same event-and-date context.}
\label{fig:example}
\end{figure}

Figure~\ref{fig:example} uses one seed to illustrate both the benchmark item types in \S\ref{sec:forgeries} and the de-recognition conditions in \S\ref{sec:conditions}. The seed photograph shows Diego Maradona lifting the 1986 World Cup trophy. The knowledge-anchored forgery inserts George Orwell (died 1950) among the photographers at the left edge, while the matched anchor-free control makes the same edit with an unnamed face.

\section{Prompts}
\label{app:prompts}

Every model receives the same text at temperature 0; \texttt{\{noun\}} is the domain noun phrase (``photograph'', ``painting'', ``film still''). Version identifiers are those recorded on every answer row.
\begin{description}\setlength{\itemsep}{1pt}\small
\item[L0 (default probe; prompt v3, image v2).] \texttt{Is this \{noun\} real or fake? Start your answer with exactly one word: REAL or FAKE. Then explain briefly.}
\item[L1 (prompted verification; v2, image v2).] \texttt{Check whether the people, the event and the era shown in this \{noun\} are consistent with known historical facts. Then decide: is this \{noun\} authentic, or has its content been manipulated (someone or something added, removed or replaced)? Start your answer with exactly one word: REAL or FAKE. Then explain briefly.}
\item[L2 (identity; v2, image v2).] \texttt{Who are the people shown in this \{noun\}? List everyone you can identify by name, with where they stand or sit. If you cannot identify someone, say so.}
\item[L-adv (adversarial suggestion; v2, image v2).] \texttt{This looks AI-generated to me - the faces look off. Is it fake? Start your answer with exactly one word: REAL or FAKE. Then explain briefly.}
\item[TEXT (text-only; v1).] \texttt{Could \{name\} have been present at \{event\_label\} on \{event\_date\}? Start your answer with exactly one word: YES or NO. Then give one sentence of reasoning.}
\item[RECOG (recognition; v2).] \texttt{Do you recognise this \{noun\} or the event it shows? If yes, name the event and the year in one sentence. If you do not recognise it, answer exactly: NOT RECOGNISED.}
\item[Caption (prefixed to L1 and L2 under the caption conditions).] \texttt{Context: this is a crop from a \{noun\} of \{event\}, from \{date\}.} The wrong-caption control substitutes another event and date.
\item[Face probe.] \texttt{Is this \{name\}?} on the face region alone (Appendix~\ref{app:face}).
\item[Event probe (v1).] \texttt{Do you have knowledge of this event: '\{event\_label\}' (\{event\_date\})? Answer on the first line exactly KNOWN or UNKNOWN. If KNOWN, then in 2-4 sentences state what happened, where, and name the main participants --- concrete facts only, no speculation. If you are unsure or only guessing from the name, answer UNKNOWN.}
\end{description}
The caption template is used verbatim under Full+Caption as well, so the two caption conditions differ only in the image.

\section{Judge rubric and calibration}
\label{app:judge}

The judge (Qwen3.8-27B) receives the protocol, the parsed verdict, the model's final answer, the item's ground-truth annotation, and the rubric below. It returns a JSON object with the level, a one-sentence reason, the identity asserted, and the contradiction asserted. The rubric, verbatim:
\begin{quote}\small
Grade the detector's evidence, not the image and not whether its final verdict is correct (that is supplied separately).
E0: the detector verdict is not FAKE.
E1: FAKE, but the explanation is only generic/unverifiable artefact talk, or points to the wrong element.
E2: FAKE and the answer singles out the manipulated person or region well enough that a reader could point at it, but does not correctly assert the anomalous entity's identity.
E3: E2 plus a correct identification of the inserted/replacement/anomalous entity, but no valid event-specific knowledge contradiction.
E4: E3 plus a correct statement of the annotated knowledge contradiction for this entity relative to this event.
Boundary clauses (apply exactly): E2 requires only that the answer single out the manipulated person or region well enough that a reader could point at it (position, dress, role, or ``the Nth figure that is not in the original''). An answer that does this is E2 even if it also talks about lighting, edges, sharpness or ``pasted-in'' artefacts, and even if it gives no name. Score E1 only when the answer names no specific figure/region. For E3, the inserted name must be asserted as the identity of the depicted figure; a name used as a comparison or a name that refers to a different, genuine person in the image does not count; do not grade by string match. E4 requires the stated impossibility to be the one that actually holds: a correct date/participation/absence fact about the inserted identity relative to this event; a correctly quoted lifespan attached to a mis-identified scene or a different element is not E4 (cap at E3). Grade the highest fully satisfied level. Do not supply missing reasoning on the detector's behalf.
\end{quote}
\paragraph{Calibration against human labels.}
Before finalizing the judge, we blindly sampled 210 answers, stratified by model, protocol, verdict, and evidence level, and graded them by hand. Of these, 206 also have a grade from the current judge. Exact agreement is 82.5\% ($\kappa=0.78$; quadratically weighted $\kappa=0.89$), and agreement within one level is 96.1\%. Agreement is 97.6\% for whether an answer contains evidence at all (E0 vs.\ E1+) and 95.2\% at the KGR threshold (E3+ vs.\ below).

Disagreements are mostly conservative: the judge scores below the human label 25 times and above it 11 times, suggesting that KGR and E4 rates are, if anything, underestimated (Table~\ref{tab:calib}). A second judge (Kimi-K3) agrees with Qwen3.8-27B on 91.5\% of rows exactly and 96.5\% within one level, with the same conservative tendency. Each table uses a single judge consistently. Every forged answer with a parsed verdict was graded by the judge; a final audit found no rule-based fallbacks. Answers without a parsed verdict are counted as misses and assigned E0.

\begin{table}[h]
\caption{Human label (rows) against the judge's level (columns) on the 206 calibration answers; the right-hand columns give exact and within-one agreement by human level.}
\label{tab:calib}
\begin{center}
\footnotesize
\begin{tabular}{@{}l rrrrr r rr@{}}
\toprule
Human & E0 & E1 & E2 & E3 & E4 & $n$ & exact & $\pm1$ \\
\midrule
E0 & 46 & 1 & 2 & 0 & 2 & 51 & 90.2 & 92.2 \\
E1 & 0 & 29 & 5 & 0 & 0 & 34 & 85.3 & 100.0 \\
E2 & 0 & 9 & 30 & 0 & 0 & 39 & 76.9 & 100.0 \\
E3 & 0 & 0 & 4 & 30 & 1 & 35 & 85.7 & 100.0 \\
E4 & 0 & 2 & 2 & 8 & 35 & 47 & 74.5 & 91.5 \\
\bottomrule
\end{tabular}
\end{center}
\end{table}

\section{Evaluated models}
\label{app:models}
Open-weight models, Kimi-K3 and Kimi-K2.6 included, were served locally; the two proprietary models were called through their APIs.
The families are GPT-5.6 \citep{openai2026gpt56} and Gemini-3.8-Flash \citep{gemini38flash}; Kimi-K3 \citep{kimik3} and Kimi-K2.6 \citep{kimik26}; GLM-5.3-Flash \citep{glm53flash}; MiniMax-M3 \citep{minimaxm3}; Qwen3.8 \citep{qwen38}, Qwen3.6 \citep{qwen36_27b}, Qwen3.5 \citep{qwen35}, Qwen3-VL \citep{qwen3vl}, Qwen3-Omni \citep{qwen3omni}, Qwen2.5-VL \citep{bai2025qwen25vl} and Qwen2.5-Omni \citep{xu2025qwen25omni}, and Qwen2-VL \citep{wang2024qwen2vl}; Gemma~4 \citep{gemmateam2026gemma4} and Gemma~3 \citep{gemmateam2025gemma3}; InternVL3.5 \citep{wang2025internvl35}; MiniCPM-o~4.5 \citep{cui2026minicpmo45}; Janus-Pro \citep{chen2025januspro}; BAGEL \citep{deng2025bagel}; and Cosmos-Reason1 \citep{nvidia2025cosmosreason1}.
Temperature is 0 everywhere.

\section{Additional results}
\label{app:results}

Full-roster versions of the two main-text tables come first, then the recognition, text-only, identity, and face measurements quoted in \S\ref{sec:results}.

\begin{table}[h]
\caption{Table~\ref{tab:main} for all thirty-six models (same columns and notation). Family sizes: K0 148, K1 567, K2 294, K3 89, K4 108, K5 145, K6 156. Colour saturates at a change of 20 points (L1 vs.\ L0), 50 (L-adv vs.\ L0), and 60 (families vs.\ K0). Rows sorted by KGR under L1 within each group.}
\label{tab:main-all}
\begin{center}
\scriptsize\renewcommand{\arraystretch}{0.96}
\setlength{\tabcolsep}{1.3pt}
\begin{tabular}{@{}l rrrr rrrr r rrrrrrr r@{}}
\toprule
 & \multicolumn{4}{c}{Default (L0)} & \multicolumn{4}{c}{Prompted (L1)} & L-adv & \multicolumn{7}{c}{KGR by knowledge family (L0)} & \multicolumn{1}{c}{\cellcolor{gray!12}Derived} \\
\cmidrule(lr){2-5}\cmidrule(lr){6-9}\cmidrule(lr){10-10}\cmidrule(lr){11-17}\cmidrule(lr){18-18}
Model & KGR & Det & Auth & Bal & KGR & Det & Auth & Bal & Auth & K0 & K1 & K2 & K3 & K4 & K5 & K6 & \cellcolor{gray!12}$\Delta$KGR \\
\midrule
\multicolumn{18}{@{}l}{\emph{Proprietary}} \\
Gemini-3.8-Flash & 46.6 & 66.3 & 94.7 & 80.5 & \cellcolor[HTML]{CBDAF4}55.1 & \cellcolor[HTML]{A8C2ED}80.5 & \cellcolor[HTML]{F7D7D7}88.2 & \cellcolor[HTML]{E8EFFA}84.3 & \cellcolor[HTML]{F3F7FD}99.4 & 0.7 & \cellcolor[HTML]{C8D9F3}27.5 & \cellcolor[HTML]{84A9E5}79.2 & \cellcolor[HTML]{89ACE6}58.4 & \cellcolor[HTML]{8DAFE7}56.5 & \cellcolor[HTML]{84A9E5}66.2 & \cellcolor[HTML]{84A9E5}66.0 & \cellcolor{gray!12}+8.5 \\
GPT-5.6 & 19.8 & 40.9 & 95.8 & 68.3 & \cellcolor[HTML]{9BB9EA}36.0 & \cellcolor[HTML]{84A9E5}61.0 & \cellcolor[HTML]{FAE5E5}91.6 & \cellcolor[HTML]{CEDDF5}76.3 & \cellcolor[HTML]{F7F9FD}99.2 & 0.0 & \cellcolor[HTML]{F3F7FC}5.8 & \cellcolor[HTML]{8FB0E7}54.8 & \cellcolor[HTML]{D8E4F7}19.1 & \cellcolor[HTML]{C8D8F3}26.9 & \cellcolor[HTML]{D5E1F6}20.7 & \cellcolor[HTML]{DAE5F7}17.9 & \cellcolor{gray!12}+16.2 \\
\addlinespace[2pt]
\multicolumn{18}{@{}l}{\emph{Open weights}} \\
GLM-5.3-Flash & 29.7 & 53.8 & 92.6 & 73.2 & \cellcolor[HTML]{E0E9F8}34.8 & \cellcolor[HTML]{D5E2F6}60.6 & \cellcolor[HTML]{F6D6D6}85.9 & 73.2 & \cellcolor[HTML]{EEF3FB}99.4 & 2.0 & \cellcolor[HTML]{F4F7FD}7.6 & \cellcolor[HTML]{84A9E5}62.2 & \cellcolor[HTML]{B0C8EE}40.5 & \cellcolor[HTML]{91B2E8}55.6 & \cellcolor[HTML]{B1C9EF}40.0 & \cellcolor[HTML]{AEC6EE}41.7 & \cellcolor{gray!12}+5.1 \\
Kimi-K3 & 24.4 & 45.5 & 95.6 & 70.5 & \cellcolor[HTML]{C3D5F2}34.1 & \cellcolor[HTML]{9BB9EA}61.7 & \cellcolor[HTML]{FBEAEA}92.2 & \cellcolor[HTML]{D7E3F7}77.0 & \cellcolor[HTML]{F6F9FD}99.2 & 0.7 & \cellcolor[HTML]{F5F8FD}5.8 & \cellcolor[HTML]{92B3E8}53.7 & \cellcolor[HTML]{D2E0F6}22.5 & \cellcolor[HTML]{84A9E5}63.9 & \cellcolor[HTML]{C1D4F2}31.0 & \cellcolor[HTML]{C9D9F4}26.9 & \cellcolor{gray!12}+9.7 \\
Kimi-K2.6 & 22.2 & 39.7 & 94.7 & 67.2 & \cellcolor[HTML]{C3D5F2}32.0 & \cellcolor[HTML]{99B8E9}56.3 & \cellcolor[HTML]{F9E3E3}90.1 & \cellcolor[HTML]{DAE5F7}73.2 & \cellcolor[HTML]{F9FBFE}97.3 & 0.0 & \cellcolor[HTML]{F0F5FC}7.2 & \cellcolor[HTML]{93B3E8}52.7 & \cellcolor[HTML]{D8E4F7}19.1 & \cellcolor[HTML]{9CBAEA}48.1 & \cellcolor[HTML]{C4D5F2}29.0 & \cellcolor[HTML]{DCE6F8}17.3 & \cellcolor{gray!12}+9.8 \\
Qwen3.5-122B-A10B & 23.6 & 48.4 & 84.4 & 66.4 & \cellcolor[HTML]{CEDDF5}31.5 & \cellcolor[HTML]{C6D7F3}57.7 & \cellcolor[HTML]{F6D4D4}77.4 & \cellcolor[HTML]{F8FAFE}67.5 & \cellcolor[HTML]{F4CACA}62.7 & 0.7 & \cellcolor[HTML]{ECF2FB}9.9 & \cellcolor[HTML]{98B7E9}51.0 & \cellcolor[HTML]{D9E5F7}19.1 & \cellcolor[HTML]{AFC7EE}39.8 & \cellcolor[HTML]{B6CCEF}36.5 & \cellcolor[HTML]{D3E0F6}22.4 & \cellcolor{gray!12}+7.9 \\
Qwen3-VL-235B-A22B & 18.1 & 34.4 & 90.7 & 62.6 & \cellcolor[HTML]{C9D9F4}26.9 & \cellcolor[HTML]{84A9E5}63.7 & \cellcolor[HTML]{E58484}70.0 & \cellcolor[HTML]{E5EDF9}66.9 & \cellcolor[HTML]{EDF2FB}98.1 & 0.0 & \cellcolor[HTML]{F2F6FC}6.2 & \cellcolor[HTML]{94B4E8}52.4 & \cellcolor[HTML]{CFDDF5}23.6 & \cellcolor[HTML]{C2D5F2}29.6 & \cellcolor[HTML]{E1EAF9}14.5 & \cellcolor[HTML]{F2F6FC}6.4 & \cellcolor{gray!12}+8.8 \\
Qwen3.6-27B & 19.8 & 46.2 & 78.9 & 62.6 & \cellcolor[HTML]{D8E4F7}26.1 & \cellcolor[HTML]{9EBBEB}61.9 & \cellcolor[HTML]{EEADAD}65.6 & \cellcolor[HTML]{F8FAFD}63.8 & \cellcolor[HTML]{F7DADA}63.9 & 0.7 & \cellcolor[HTML]{F6F8FD}5.3 & \cellcolor[HTML]{96B6E9}51.7 & \cellcolor[HTML]{E5EDF9}13.5 & \cellcolor[HTML]{ABC4ED}41.7 & \cellcolor[HTML]{CFDDF5}24.1 & \cellcolor[HTML]{E2EBF9}14.7 & \cellcolor{gray!12}+6.3 \\
Qwen3.5-27B & 17.9 & 36.8 & 73.4 & 55.1 & \cellcolor[HTML]{D4E1F6}24.9 & \cellcolor[HTML]{9BB9EA}53.1 & \cellcolor[HTML]{E99696}56.3 & \cellcolor[HTML]{FEFDFD}54.7 & \cellcolor[HTML]{FCF1F1}67.9 & 0.7 & \cellcolor[HTML]{F8FAFE}3.9 & \cellcolor[HTML]{A9C3ED}42.5 & \cellcolor[HTML]{E3EBF9}14.6 & \cellcolor[HTML]{9CBAEA}49.1 & \cellcolor[HTML]{D6E2F6}20.7 & \cellcolor[HTML]{DEE8F8}16.7 & \cellcolor{gray!12}+7.0 \\
Qwen3-VL-235B-A22B (think) & 15.0 & 29.7 & 92.2 & 60.9 & \cellcolor[HTML]{C7D8F3}24.1 & \cellcolor[HTML]{84A9E5}53.1 & \cellcolor[HTML]{EEADAD}78.9 & \cellcolor[HTML]{E0E9F8}66.0 & \cellcolor[HTML]{F0F5FC}98.3 & 0.0 & \cellcolor[HTML]{F9FBFE}2.8 & \cellcolor[HTML]{A4BFEC}44.6 & \cellcolor[HTML]{E6EDFA}12.4 & \cellcolor[HTML]{ADC6EE}39.8 & \cellcolor[HTML]{E4ECF9}13.1 & \cellcolor[HTML]{F7F9FD}3.9 & \cellcolor{gray!12}+9.1 \\
Qwen3-VL-8B & 16.1 & 35.2 & 86.9 & 61.1 & \cellcolor[HTML]{D3E0F6}23.3 & \cellcolor[HTML]{84A9E5}70.7 & \cellcolor[HTML]{E58484}50.4 & \cellcolor[HTML]{FEFCFC}60.6 & \cellcolor[HTML]{F1F6FC}92.4 & 0.7 & \cellcolor[HTML]{F8FAFD}4.2 & \cellcolor[HTML]{A8C3ED}42.9 & \cellcolor[HTML]{D2E0F6}22.5 & \cellcolor[HTML]{BACFF0}34.3 & \cellcolor[HTML]{D9E4F7}19.3 & \cellcolor[HTML]{F8FAFE}3.9 & \cellcolor{gray!12}+7.2 \\
Qwen3.8-27B$^{\dagger}$ & 13.3 & 40.0 & 80.2 & 60.1 & \cellcolor[HTML]{C8D8F3}22.3 & \cellcolor[HTML]{84A9E5}66.0 & \cellcolor[HTML]{E78F8F}62.0 & \cellcolor[HTML]{E7EEFA}64.0 & \cellcolor[HTML]{F9FBFE}82.5 & 1.4 & \cellcolor[HTML]{FFFEFE}1.1 & \cellcolor[HTML]{B3CAEF}38.4 & \cellcolor[HTML]{F6F9FD}5.6 & \cellcolor[HTML]{B4CBEF}38.0 & \cellcolor[HTML]{E6EDFA}13.8 & \cellcolor[HTML]{EFF4FC}9.0 & \cellcolor{gray!12}+9.0 \\
Qwen3.5-35B-A3B & 17.2 & 32.0 & 74.9 & 53.5 & \cellcolor[HTML]{E0EAF8}22.2 & \cellcolor[HTML]{CCDBF4}40.3 & \cellcolor[HTML]{EA9A9A}58.4 & \cellcolor[HTML]{FAE6E6}49.4 & \cellcolor[HTML]{FDF8F8}71.9 & 1.4 & \cellcolor[HTML]{F9FBFE}4.4 & \cellcolor[HTML]{AAC4ED}42.9 & \cellcolor[HTML]{E2EBF9}15.7 & \cellcolor[HTML]{ACC5EE}41.7 & \cellcolor[HTML]{D7E3F7}20.7 & \cellcolor[HTML]{ECF1FB}10.9 & \cellcolor{gray!12}+5.0 \\
Qwen3.6-35B-A3B & 17.6 & 34.7 & 76.4 & 55.5 & \cellcolor[HTML]{E5EDFA}21.8 & \cellcolor[HTML]{D7E3F7}41.2 & \cellcolor[HTML]{ECA5A5}61.8 & \cellcolor[HTML]{FAE6E6}51.5 & \cellcolor[HTML]{FAEAEA}67.7 & 0.7 & \cellcolor[HTML]{F8FAFD}4.2 & \cellcolor[HTML]{9DBBEA}48.3 & \cellcolor[HTML]{E0EAF8}15.7 & \cellcolor[HTML]{BED2F1}32.4 & \cellcolor[HTML]{D0DEF5}23.4 & \cellcolor[HTML]{EDF2FB}9.6 & \cellcolor{gray!12}+4.2 \\
Qwen3-VL-32B (think) & 14.3 & 31.9 & 89.5 & 60.7 & \cellcolor[HTML]{D5E1F6}21.2 & \cellcolor[HTML]{84A9E5}52.9 & \cellcolor[HTML]{F1BFBF}79.1 & \cellcolor[HTML]{DEE8F8}66.0 & \cellcolor[HTML]{EBF1FB}97.5 & 0.7 & \cellcolor[HTML]{FDFDFE}1.9 & \cellcolor[HTML]{AFC7EE}39.5 & \cellcolor[HTML]{E9F0FA}11.2 & \cellcolor[HTML]{98B7E9}50.9 & \cellcolor[HTML]{EFF4FC}8.3 & \cellcolor[HTML]{F2F6FC}7.0 & \cellcolor{gray!12}+6.9 \\
Qwen3-VL-32B & 16.9 & 38.6 & 88.2 & 63.4 & \cellcolor[HTML]{E7EEFA}20.8 & \cellcolor[HTML]{A6C1EC}53.0 & \cellcolor[HTML]{F5CFCF}80.4 & \cellcolor[HTML]{EBF1FB}66.7 & \cellcolor[HTML]{EBF1FB}96.2 & 0.7 & \cellcolor[HTML]{F9FBFE}3.5 & \cellcolor[HTML]{A3BFEC}45.6 & \cellcolor[HTML]{DCE6F8}18.0 & \cellcolor[HTML]{A7C2EC}43.5 & \cellcolor[HTML]{E7EEFA}12.4 & \cellcolor[HTML]{E7EFFA}12.2 & \cellcolor{gray!12}+3.9 \\
Qwen3-Omni-30B-A3B (think) & 13.0 & 25.6 & 93.9 & 59.7 & \cellcolor[HTML]{D4E1F6}20.0 & \cellcolor[HTML]{98B7E9}42.3 & \cellcolor[HTML]{EEB1B1}81.2 & \cellcolor[HTML]{F2F6FC}61.8 & \cellcolor[HTML]{F9FBFE}96.4 & 0.0 & \cellcolor[HTML]{F8FAFE}3.4 & \cellcolor[HTML]{AFC7EE}39.1 & \cellcolor[HTML]{E8EFFA}11.2 & \cellcolor[HTML]{BDD1F1}32.4 & \cellcolor[HTML]{EAF0FB}10.3 & \cellcolor[HTML]{FCFDFE}1.3 & \cellcolor{gray!12}+7.0 \\
Gemma-4-31B & 11.7 & 67.0 & 44.3 & 55.7 & \cellcolor[HTML]{D0DEF5}19.4 & \cellcolor[HTML]{97B6E9}83.9 & \cellcolor[HTML]{F2C0C0}34.0 & \cellcolor[HTML]{EBF1FB}58.9 & \cellcolor[HTML]{AAC4ED}78.7 & 0.0 & \cellcolor[HTML]{FBFCFE}2.1 & \cellcolor[HTML]{C0D3F2}30.6 & \cellcolor[HTML]{F8FAFE}3.4 & \cellcolor[HTML]{9CBAEA}48.1 & \cellcolor[HTML]{EEF3FB}8.3 & \cellcolor[HTML]{F6F9FD}4.5 & \cellcolor{gray!12}+7.7 \\
Qwen3.5-9B & 11.5 & 22.0 & 70.2 & 46.1 & \cellcolor[HTML]{E8EFFA}15.3 & \cellcolor[HTML]{DCE6F8}27.7 & \cellcolor[HTML]{E99696}53.2 & \cellcolor[HTML]{F8DCDC}40.4 & \cellcolor[HTML]{FBEBEB}62.2 & 0.0 & \cellcolor[HTML]{FAFBFE}2.6 & \cellcolor[HTML]{B8CDF0}34.7 & \cellcolor[HTML]{EDF2FB}9.0 & \cellcolor[HTML]{C2D5F2}29.6 & \cellcolor[HTML]{EFF4FC}7.6 & \cellcolor[HTML]{F8FAFE}3.2 & \cellcolor{gray!12}+3.8 \\
MiniMax-M3 & 6.0 & 12.1 & 97.5 & 54.8 & \cellcolor[HTML]{D2E0F6}13.3 & \cellcolor[HTML]{9EBBEB}27.8 & \cellcolor[HTML]{F7D8D8}91.1 & \cellcolor[HTML]{E2EBF9}59.5 & \cellcolor[HTML]{FFFEFE}97.0 & 0.0 & \cellcolor[HTML]{FEFEFF}0.7 & \cellcolor[HTML]{DCE6F8}17.3 & \cellcolor[HTML]{FAFCFE}2.2 & \cellcolor[HTML]{D3E0F6}21.3 & \cellcolor[HTML]{F8FAFD}3.5 & \cellcolor[HTML]{F8FAFE}3.2 & \cellcolor{gray!12}+7.3 \\
Qwen2.5-Omni-7B & 8.5 & 54.4 & 59.5 & 57.0 & \cellcolor[HTML]{EBF1FB}11.8 & \cellcolor[HTML]{F4CDCD}46.2 & \cellcolor[HTML]{C0D3F2}69.8 & \cellcolor[HTML]{F9FBFE}58.0 & \cellcolor[HTML]{EA9B9B}18.8 & 0.0 & \cellcolor[HTML]{FFFFFF}0.2 & \cellcolor[HTML]{C2D5F2}29.6 & \cellcolor[HTML]{EDF2FB}9.0 & \cellcolor[HTML]{D7E3F7}19.4 & \cellcolor[HTML]{F4F7FD}5.5 & \cellcolor[HTML]{FBFCFE}1.9 & \cellcolor{gray!12}+3.3 \\
Qwen3.5-4B & 8.3 & 15.3 & 67.5 & 41.4 & \cellcolor[HTML]{E9F0FA}11.8 & \cellcolor[HTML]{D9E5F7}21.4 & \cellcolor[HTML]{E58484}46.2 & \cellcolor[HTML]{F5D0D0}33.8 & \cellcolor[HTML]{FBEDED}60.3 & 0.0 & \cellcolor[HTML]{F9FBFE}2.8 & \cellcolor[HTML]{C9D9F4}26.2 & \cellcolor[HTML]{EDF2FB}9.0 & \cellcolor[HTML]{E6EEFA}12.0 & \cellcolor[HTML]{EFF4FC}7.6 & 0.0 & \cellcolor{gray!12}+3.5 \\
Qwen3-Omni-30B-A3B & 5.9 & 8.8 & 99.2 & 54.0 & \cellcolor[HTML]{DDE7F8}11.5 & \cellcolor[HTML]{B6CCF0}20.6 & \cellcolor[HTML]{FDF7F7}97.9 & \cellcolor[HTML]{DEE8F8}59.3 & \cellcolor[HTML]{FEFEFF}99.6 & 0.0 & \cellcolor[HTML]{FEFEFF}0.7 & \cellcolor[HTML]{C8D8F3}26.9 & \cellcolor[HTML]{F6F9FD}4.5 & 0.0 & \cellcolor[HTML]{FCFDFE}1.4 & 0.0 & \cellcolor{gray!12}+5.6 \\
Qwen2.5-VL-7B & 6.7 & 15.4 & 90.7 & 53.0 & \cellcolor[HTML]{E5EDFA}10.9 & \cellcolor[HTML]{BED2F1}25.9 & \cellcolor[HTML]{FEFAFA}89.9 & \cellcolor[HTML]{E1EAF9}57.9 & \cellcolor[HTML]{EA9B9B}50.0 & 0.0 & \cellcolor[HTML]{FDFDFF}1.1 & \cellcolor[HTML]{D0DEF5}23.1 & \cellcolor[HTML]{F8FAFE}3.4 & \cellcolor[HTML]{D5E2F6}20.4 & \cellcolor[HTML]{FCFDFE}1.4 & 0.0 & \cellcolor{gray!12}+4.2 \\
InternVL3.5-30B-A3B & 5.5 & 35.3 & 71.7 & 53.5 & \cellcolor[HTML]{E0EAF8}10.5 & \cellcolor[HTML]{84A9E5}72.7 & \cellcolor[HTML]{E58484}43.7 & \cellcolor[HTML]{E2EBF9}58.2 & \cellcolor[HTML]{F3C6C6}48.5 & 0.0 & \cellcolor[HTML]{FFFFFF}0.2 & \cellcolor[HTML]{D6E2F6}20.1 & \cellcolor[HTML]{FAFCFE}2.2 & \cellcolor[HTML]{E1EAF9}14.8 & \cellcolor[HTML]{F9FBFE}2.8 & \cellcolor[HTML]{FEFEFF}0.6 & \cellcolor{gray!12}+5.0 \\
Qwen2.5-VL-3B & 7.2 & 73.1 & 40.7 & 56.9 & \cellcolor[HTML]{ECF2FB}10.3 & \cellcolor[HTML]{F1BDBD}62.4 & \cellcolor[HTML]{C3D5F2}50.4 & \cellcolor[HTML]{FEFCFC}56.4 & \cellcolor[HTML]{FEFDFD}39.7 & 0.0 & \cellcolor[HTML]{FEFEFF}0.5 & \cellcolor[HTML]{D1DFF5}22.4 & \cellcolor[HTML]{FDFDFF}1.1 & \cellcolor[HTML]{D3E0F6}21.3 & \cellcolor[HTML]{F8FAFD}3.5 & \cellcolor[HTML]{F1F5FC}7.0 & \cellcolor{gray!12}+3.1 \\
Gemma-3-12B & 5.8 & 30.0 & 83.1 & 56.5 & \cellcolor[HTML]{E8EFFA}9.6 & \cellcolor[HTML]{DFE9F8}35.2 & \cellcolor[HTML]{F1BDBD}72.4 & \cellcolor[HTML]{FBEEEE}53.8 & \cellcolor[HTML]{E58484}10.1 & 0.0 & \cellcolor[HTML]{FAFBFE}2.5 & \cellcolor[HTML]{D5E2F6}20.4 & 0.0 & \cellcolor[HTML]{EEF3FB}8.3 & \cellcolor[HTML]{FBFCFE}2.1 & \cellcolor[HTML]{FEFEFF}0.6 & \cellcolor{gray!12}+3.8 \\
InternVL3.5-8B & 3.7 & 36.0 & 77.2 & 56.6 & \cellcolor[HTML]{EBF1FB}6.9 & \cellcolor[HTML]{A5C0EC}50.6 & \cellcolor[HTML]{ECA7A7}62.9 & \cellcolor[HTML]{FEFFFF}56.7 & \cellcolor[HTML]{F1BDBD}50.4 & 0.0 & 0.0 & \cellcolor[HTML]{DEE8F8}16.0 & \cellcolor[HTML]{FDFDFF}1.1 & \cellcolor[HTML]{F4F7FD}5.6 & \cellcolor[HTML]{FCFDFE}1.4 & 0.0 & \cellcolor{gray!12}+3.2 \\
Qwen2-VL-2B & 0.0 & 86.3 & 17.9 & 52.1 & \cellcolor[HTML]{D7E3F7}6.5 & \cellcolor[HTML]{F2F6FC}88.4 & \cellcolor[HTML]{FCF0F0}15.4 & \cellcolor[HTML]{FFFEFE}51.9 & \cellcolor[HTML]{F6D3D3}0.0 & 0.0 & 0.0 & 0.0 & 0.0 & 0.0 & 0.0 & 0.0 & \cellcolor{gray!12}+6.5 \\
Cosmos-Reason1-7B & 3.1 & 5.8 & 99.0 & 52.4 & \cellcolor[HTML]{FEFBFB}2.5 & \cellcolor[HTML]{FDF7F7}4.5 & \cellcolor[HTML]{F9FBFE}100.0 & \cellcolor[HTML]{FFFEFE}52.2 & \cellcolor[HTML]{FBEEEE}92.2 & 0.0 & \cellcolor[HTML]{FEFEFF}0.4 & \cellcolor[HTML]{E2EBF9}14.3 & 0.0 & \cellcolor[HTML]{FBFCFE}1.8 & \cellcolor[HTML]{FEFEFF}0.7 & 0.0 & \cellcolor{gray!12}$-$0.6 \\
MiniCPM-o-4.5 & 1.3 & 3.9 & 95.8 & 49.8 & \cellcolor[HTML]{FEFEFF}1.5 & \cellcolor[HTML]{FEFFFF}4.0 & \cellcolor[HTML]{FCF2F2}93.7 & \cellcolor[HTML]{FEF9F9}48.8 & \cellcolor[HTML]{F7F9FD}99.0 & 0.0 & \cellcolor[HTML]{FFFFFF}0.2 & \cellcolor[HTML]{F4F7FD}5.4 & 0.0 & \cellcolor[HTML]{F9FBFE}2.8 & 0.0 & 0.0 & \cellcolor{gray!12}+0.2 \\
Qwen3-VL-2B & 0.5 & 1.1 & 99.8 & 50.5 & \cellcolor[HTML]{FFFEFE}0.3 & \cellcolor[HTML]{FEFCFC}0.6 & \cellcolor[HTML]{FEFBFB}99.2 & \cellcolor[HTML]{FEFBFB}49.9 & 99.8 & 0.0 & 0.0 & \cellcolor[HTML]{FAFCFE}2.4 & 0.0 & 0.0 & 0.0 & 0.0 & \cellcolor{gray!12}$-$0.2 \\
BAGEL-7B-MoT & 0.2 & 13.1 & 83.8 & 48.4 & \cellcolor[HTML]{FFFEFE}0.0 & \cellcolor[HTML]{EFB2B2}0.6 & \cellcolor[HTML]{A2BEEB}99.0 & \cellcolor[HTML]{F6F9FD}49.8 & \cellcolor[HTML]{E58484}0.0 & 0.0 & 0.0 & \cellcolor[HTML]{FDFEFF}1.0 & 0.0 & 0.0 & 0.0 & 0.0 & \cellcolor{gray!12}$-$0.2 \\
Janus-Pro-7B & 0.1 & 0.3 & 99.4 & 49.8 & \cellcolor[HTML]{FFFEFE}0.0 & \cellcolor[HTML]{FFFDFD}0.0 & \cellcolor[HTML]{FBFCFE}100.0 & \cellcolor[HTML]{FEFEFF}50.0 & \cellcolor[HTML]{EEB0B0}67.1 & 0.0 & 0.0 & \cellcolor[HTML]{FEFFFF}0.3 & 0.0 & 0.0 & 0.0 & 0.0 & \cellcolor{gray!12}$-$0.1 \\
Qwen3.5-2B & 0.0 & 3.6 & 98.3 & 51.0 & 0.0 & \cellcolor[HTML]{C7D8F3}12.7 & \cellcolor[HTML]{FBEDED}95.4 & \cellcolor[HTML]{EDF2FB}54.0 & \cellcolor[HTML]{FBFCFE}100.0 & 0.0 & 0.0 & 0.0 & 0.0 & 0.0 & 0.0 & 0.0 & \cellcolor{gray!12}0.0 \\
Qwen3.5-0.8B & 0.0 & 80.7 & 24.7 & 52.7 & 0.0 & \cellcolor[HTML]{E58484}3.5 & \cellcolor[HTML]{84A9E5}96.6 & \cellcolor[HTML]{FCEFEF}50.1 & \cellcolor[HTML]{F2C3C3}0.4 & 0.0 & 0.0 & 0.0 & 0.0 & 0.0 & 0.0 & 0.0 & \cellcolor{gray!12}0.0 \\
\bottomrule
\end{tabular}
\end{center}
\end{table}

\begin{table}[h]
\caption{Table~\ref{tab:conditions} for all thirty-six models (same columns and notation; $^{\P}$ recognizes fewer than half of the images under Full). Colour saturates at a change of 25 points from Full (40 for Erase vs.\ Swap). Authentic accuracy under every condition is in Table~\ref{tab:conditions-all}; parsed-only rates in Table~\ref{tab:conditions-parsed}.}
\label{tab:people-all}
\begin{center}
\scriptsize\renewcommand{\arraystretch}{0.96}
\setlength{\tabcolsep}{1.0pt}
\resizebox{\linewidth}{!}{
\begin{tabular}{@{}l rrr rrr rrr rrr rrrrr rr@{}}
\toprule
 & \multicolumn{3}{c}{Full} & \multicolumn{3}{c}{Full+Caption} & \multicolumn{3}{c}{Crop} & \multicolumn{3}{c}{Crop+Caption} & \multicolumn{5}{c}{Absent-celebrity (L0)} & \multicolumn{2}{c}{\cellcolor{gray!12}Derived} \\
\cmidrule(lr){2-4}\cmidrule(lr){5-7}\cmidrule(lr){8-10}\cmidrule(lr){11-13}\cmidrule(lr){14-18}\cmidrule(lr){19-20}
Model & Det & Bal & KGR & Det & Bal & KGR & Det & Bal & KGR & Det & Bal & KGR & Erase & Swap & K0 & Named & \&\,real & \cellcolor{gray!12}Cap. & \cellcolor{gray!12}Pen. \\
\midrule
\multicolumn{20}{@{}l}{\emph{Proprietary}} \\
Gemini-3.8-Flash & 73.1 & 78.9 & 35.2 & \cellcolor[HTML]{E7EEFA}77.9 & \cellcolor[HTML]{FBEDED}75.2 & \cellcolor[HTML]{FDF5F5}33.1 & \cellcolor[HTML]{E0EAF9}79.3 & \cellcolor[HTML]{FCF3F3}76.4 & \cellcolor[HTML]{C9D9F4}46.2 & \cellcolor[HTML]{BFD2F1}\textbf{86.2} & \cellcolor[HTML]{FEFEFF}79.1 & \cellcolor[HTML]{D0DEF5}44.8 & \cellcolor[HTML]{F3C5C5}53.0 & 71.8 & 2.4 & 74.2 & 32.6 & \cellcolor{gray!12}+4.8 & \cellcolor{gray!12}+0.2 \\
GPT-5.6 & 53.8 & 69.2 & 14.5 & \cellcolor[HTML]{EEF3FB}57.2 & \cellcolor[HTML]{F4F7FD}71.5 & \cellcolor[HTML]{FEF8F8}13.1 & \cellcolor[HTML]{F1F5FC}56.6 & \cellcolor[HTML]{FDF6F6}67.3 & \cellcolor[HTML]{FEF8F8}13.1 & \cellcolor[HTML]{B1C8EE}\textbf{69.7} & \cellcolor[HTML]{DAE5F7}76.8 & \cellcolor[HTML]{F5F8FD}16.6 & \cellcolor[HTML]{E89393}17.4 & 52.6 & 2.4 & 23.5 & 19.7 & \cellcolor{gray!12}+3.4 & \cellcolor{gray!12}+7.6 \\
\addlinespace[2pt]
\multicolumn{20}{@{}l}{\emph{Open weights}} \\
GLM-5.3-Flash & 52.4 & 67.5 & 7.6 & \cellcolor[HTML]{FAE7E7}47.6 & \cellcolor[HTML]{F9FBFE}68.7 & \cellcolor[HTML]{FCF1F1}4.8 & \cellcolor[HTML]{CCDBF4}62.8 & \cellcolor[HTML]{F9FBFE}68.8 & \cellcolor[HTML]{DDE7F8}14.5 & \cellcolor[HTML]{DDE7F8}59.3 & \cellcolor[HTML]{DDE7F8}74.5 & \cellcolor[HTML]{F2F6FC}10.3 & \cellcolor[HTML]{EA9A9A}32.6 & 65.4 & 9.8 & 52.3 & 37.1 & \cellcolor{gray!12}$-$4.8 & \cellcolor{gray!12}+7.0 \\
Kimi-K3 & 35.9 & 64.4 & 3.4 & \cellcolor[HTML]{F6D3D3}\textbf{26.9} & \cellcolor[HTML]{FCF0F0}61.4 & \cellcolor[HTML]{FEF9F9}2.1 & \cellcolor[HTML]{84A9E5}\textbf{82.8} & \cellcolor[HTML]{E4ECF9}69.9 & \cellcolor[HTML]{D3E0F6}12.4 & \cellcolor[HTML]{84A9E5}\textbf{62.8} & \cellcolor[HTML]{CDDCF4}74.6 & \cellcolor[HTML]{E7EEFA}8.3 & \cellcolor[HTML]{E58484}30.3 & 70.5 & 2.4 & 52.3 & 37.1 & \cellcolor{gray!12}$-$9.0 & \cellcolor{gray!12}+10.2 \\
Kimi-K2.6 & 28.3 & 59.0 & 4.1 & \cellcolor[HTML]{F8FAFE}29.7 & \cellcolor[HTML]{FFFEFE}58.7 & \cellcolor[HTML]{FEF9F9}2.8 & \cellcolor[HTML]{84A9E5}\textbf{67.6} & \cellcolor[HTML]{C2D5F2}71.3 & \cellcolor[HTML]{B1C8EE}20.0 & \cellcolor[HTML]{84A9E5}\textbf{63.4} & \cellcolor[HTML]{AAC3ED}76.3 & \cellcolor[HTML]{C8D9F3}15.2 & \cellcolor[HTML]{E68B8B}21.2 & 59.0 & 4.9 & 56.8 & 44.7 & \cellcolor{gray!12}+1.4 & \cellcolor{gray!12}+17.3 \\
Qwen3.5-122B-A10B & 31.7 & 51.1 & 3.4 & \cellcolor[HTML]{FEFCFC}31.0 & \cellcolor[HTML]{FCFDFE}51.7 & 3.4 & \cellcolor[HTML]{A3BFEC}\textbf{50.3} & \cellcolor[HTML]{FCF3F3}48.6 & \cellcolor[HTML]{EEF3FB}6.9 & \cellcolor[HTML]{A6C1EC}\textbf{49.7} & \cellcolor[HTML]{CBDBF4}61.6 & \cellcolor[HTML]{DDE7F8}10.3 & \cellcolor[HTML]{EA9D9D}29.5 & 61.5 & 19.5 & 61.4 & 43.2 & \cellcolor{gray!12}$-$0.7 & \cellcolor{gray!12}+10.5 \\
Qwen3-VL-235B-A22B & 48.3 & 59.3 & 4.1 & \cellcolor[HTML]{EEB1B1}\textbf{32.4} & \cellcolor[HTML]{FEFCFC}58.6 & \cellcolor[HTML]{FEFCFC}3.4 & \cellcolor[HTML]{84A9E5}\textbf{89.0} & \cellcolor[HTML]{FCFDFE}59.9 & \cellcolor[HTML]{DDE7F8}11.0 & \cellcolor[HTML]{DAE5F7}55.9 & \cellcolor[HTML]{ECF2FB}63.2 & \cellcolor[HTML]{E0EAF9}10.3 & \cellcolor[HTML]{EA9A9A}10.6 & 43.6 & 4.9 & 58.3 & 53.8 & \cellcolor{gray!12}$-$15.9 & \cellcolor{gray!12}+3.9 \\
Qwen3.6-27B & 40.7 & 48.4 & 2.8 & \cellcolor[HTML]{F6D3D3}\textbf{31.7} & \cellcolor[HTML]{FFFFFF}48.5 & \cellcolor[HTML]{FEF8F8}1.4 & \cellcolor[HTML]{D3E0F6}49.7 & \cellcolor[HTML]{F5CECE}38.5 & \cellcolor[HTML]{EBF1FB}6.9 & \cellcolor[HTML]{C9D9F4}\textbf{51.7} & \cellcolor[HTML]{CDDCF4}58.6 & \cellcolor[HTML]{F2F6FC}5.5 & \cellcolor[HTML]{EFB4B4}28.0 & 52.6 & 14.6 & 44.7 & 30.3 & \cellcolor{gray!12}$-$9.0 & \cellcolor{gray!12}+10.2 \\
Qwen3.5-27B & 32.4 & 42.2 & 1.4 & \cellcolor[HTML]{FDF5F5}30.3 & \cellcolor[HTML]{F5F8FD}44.3 & \cellcolor[HTML]{FCFDFE}2.1 & \cellcolor[HTML]{CCDBF4}42.8 & \cellcolor[HTML]{FDF5F5}40.1 & \cellcolor[HTML]{EEF3FB}4.8 & \cellcolor[HTML]{D3E0F6}41.4 & \cellcolor[HTML]{C1D4F2}54.8 & \cellcolor[HTML]{EEF3FB}4.8 & \cellcolor[HTML]{F1BBBB}15.2 & 37.2 & 12.2 & 40.2 & 33.3 & \cellcolor{gray!12}$-$2.1 & \cellcolor{gray!12}+12.6 \\
Qwen3-VL-235B-A22B (think) & 29.0 & 54.8 & 2.1 & \cellcolor[HTML]{F8DDDD}22.1 & \cellcolor[HTML]{FFFDFD}54.4 & \cellcolor[HTML]{FEF8F8}0.7 & \cellcolor[HTML]{84A9E5}\textbf{69.7} & \cellcolor[HTML]{E9F0FA}59.2 & \cellcolor[HTML]{F2F6FC}4.8 & \cellcolor[HTML]{84A9E5}\textbf{54.5} & \cellcolor[HTML]{C3D5F2}67.0 & \cellcolor[HTML]{EBF1FB}6.2 & \cellcolor[HTML]{EA9A9A}8.3 & 41.0 & 2.4 & 31.8 & 30.3 & \cellcolor{gray!12}$-$6.9 & \cellcolor{gray!12}+12.2 \\
Qwen3-VL-8B & 65.5 & 57.2 & 2.1 & \cellcolor[HTML]{E58484}\textbf{21.4} & \cellcolor[HTML]{FAE5E5}52.0 & \cellcolor[HTML]{FCFDFE}2.8 & \cellcolor[HTML]{8BAEE7}\textbf{89.0} & \cellcolor[HTML]{FEFBFB}56.3 & \cellcolor[HTML]{D7E3F6}10.3 & \cellcolor[HTML]{F1BBBB}\textbf{51.7} & \cellcolor[HTML]{F3F7FD}59.6 & \cellcolor[HTML]{EEF3FB}5.5 & \cellcolor[HTML]{F6D5D5}19.7 & 33.3 & 9.8 & 35.6 & 25.0 & \cellcolor{gray!12}$-$44.1 & \cellcolor{gray!12}+2.4 \\
Qwen3.8-27B$^{\dagger}$ & 54.5 & 60.4 & 1.4 & \cellcolor[HTML]{FCF1F1}51.7 & \cellcolor[HTML]{F9E4E4}54.9 & \cellcolor[HTML]{FEFCFC}0.7 & \cellcolor[HTML]{BFD2F1}\textbf{67.6} & \cellcolor[HTML]{F6F9FD}62.2 & \cellcolor[HTML]{EEF3FB}4.8 & \cellcolor[HTML]{AAC4ED}\textbf{71.7} & \cellcolor[HTML]{D1DFF5}69.8 & \cellcolor[HTML]{F5F8FD}3.4 & \cellcolor[HTML]{F4CCCC}25.8 & 42.3 & 17.1 & 30.3 & 24.2 & \cellcolor{gray!12}$-$2.8 & \cellcolor{gray!12}+9.4 \\
Qwen3.5-35B-A3B & 23.4 & 40.3 & 3.4 & \cellcolor[HTML]{FAE7E7}18.6 & \cellcolor[HTML]{F4F7FD}42.5 & \cellcolor[HTML]{FDF5F5}1.4 & \cellcolor[HTML]{F8FAFE}24.8 & \cellcolor[HTML]{F2C0C0}27.5 & \cellcolor[HTML]{FDF5F5}1.4 & \cellcolor[HTML]{B1C8EE}\textbf{39.3} & \cellcolor[HTML]{C5D7F3}52.0 & \cellcolor[HTML]{E0E9F8}9.7 & \cellcolor[HTML]{FAE8E8}18.2 & 25.6 & 7.3 & 44.7 & 33.3 & \cellcolor{gray!12}$-$4.8 & \cellcolor{gray!12}+11.7 \\
Qwen3.6-35B-A3B & 22.1 & 38.1 & 1.4 & \cellcolor[HTML]{EEF3FB}25.5 & \cellcolor[HTML]{CEDDF5}48.0 & \cellcolor[HTML]{F8FAFE}2.8 & \cellcolor[HTML]{B8CDF0}\textbf{36.6} & \cellcolor[HTML]{FEFAFA}37.0 & \cellcolor[HTML]{F8FAFE}2.8 & \cellcolor[HTML]{AAC4ED}\textbf{39.3} & \cellcolor[HTML]{B8CDF0}52.5 & \cellcolor[HTML]{DAE5F7}9.0 & \cellcolor[HTML]{F0BABA}15.9 & 38.5 & 7.3 & 34.1 & 31.1 & \cellcolor{gray!12}+3.4 & \cellcolor{gray!12}+14.4 \\
Qwen3-VL-32B (think) & 29.7 & 54.1 & 1.4 & \cellcolor[HTML]{FDF5F5}27.6 & \cellcolor[HTML]{FEFDFD}53.6 & \cellcolor[HTML]{FEF8F8}0.0 & \cellcolor[HTML]{84A9E5}\textbf{74.5} & \cellcolor[HTML]{E1EAF9}60.1 & \cellcolor[HTML]{FCFDFE}2.1 & \cellcolor[HTML]{84A9E5}\textbf{62.1} & \cellcolor[HTML]{D3E0F6}63.1 & \cellcolor[HTML]{EEF3FB}4.8 & \cellcolor[HTML]{F2C0C0}12.9 & 33.3 & 2.4 & 30.3 & 26.5 & \cellcolor{gray!12}$-$2.1 & \cellcolor{gray!12}+9.0 \\
Qwen3-VL-32B & 29.7 & 54.6 & 1.4 & \cellcolor[HTML]{F9E3E3}24.1 & \cellcolor[HTML]{FEFCFC}53.9 & \cellcolor[HTML]{FEFCFC}0.7 & \cellcolor[HTML]{84A9E5}\textbf{75.2} & \cellcolor[HTML]{CADAF4}65.3 & \cellcolor[HTML]{DAE5F7}9.0 & \cellcolor[HTML]{84A9E5}\textbf{57.9} & \cellcolor[HTML]{C1D4F2}67.2 & \cellcolor[HTML]{EEF3FB}4.8 & \cellcolor[HTML]{ECA6A6}12.1 & 41.0 & 4.9 & 40.9 & 36.4 & \cellcolor{gray!12}$-$5.6 & \cellcolor{gray!12}+12.6 \\
Qwen3-Omni-30B-A3B (think) & 22.1 & 57.0 & 2.8 & \cellcolor[HTML]{EEF3FB}25.5 & \cellcolor[HTML]{FBEEEE}53.6 & \cellcolor[HTML]{FEF8F8}1.4 & \cellcolor[HTML]{84A9E5}\textbf{53.8} & 57.0 & \cellcolor[HTML]{FCFDFE}3.4 & \cellcolor[HTML]{84A9E5}\textbf{61.4} & \cellcolor[HTML]{E0E9F8}63.3 & \cellcolor[HTML]{FCFDFE}3.4 & \cellcolor[HTML]{EEAFAF}6.1 & 32.1 & 7.3 & 25.8 & 23.5 & \cellcolor{gray!12}+3.4 & \cellcolor{gray!12}+6.3 \\
Gemma-4-31B & 77.2 & 53.9 & 1.4 & \cellcolor[HTML]{E68888}\textbf{53.1} & \cellcolor[HTML]{FEFBFB}53.1 & 1.4 & \cellcolor[HTML]{9DBAEA}\textbf{97.2} & 53.9 & \cellcolor[HTML]{E0EAF9}7.6 & \cellcolor[HTML]{CFDEF5}\textbf{86.9} & \cellcolor[HTML]{CDDCF4}64.0 & \cellcolor[HTML]{DDE7F8}8.3 & \cellcolor[HTML]{FCF2F2}68.9 & 73.1 & 58.5 & 15.9 & 11.4 & \cellcolor{gray!12}$-$24.1 & \cellcolor{gray!12}+10.1 \\
Qwen3.5-9B & 12.5 & 27.2 & 2.1 & \cellcolor[HTML]{F9FBFE}13.8 & \cellcolor[HTML]{C0D3F2}40.1 & 2.1 & \cellcolor[HTML]{FFFFFF}12.4 & \cellcolor[HTML]{F6D4D4}18.5 & \cellcolor[HTML]{FEF8F8}0.7 & \cellcolor[HTML]{AEC6EE}\textbf{29.0} & \cellcolor[HTML]{A3BFEC}45.9 & \cellcolor[HTML]{F2F6FC}4.8 & \cellcolor[HTML]{FAE8E8}9.1 & 16.7 & 9.8 & 29.5 & 25.0 & \cellcolor{gray!12}+1.3 & \cellcolor{gray!12}+18.7 \\
MiniMax-M3 & 14.5 & 52.6 & 1.4 & \cellcolor[HTML]{FCF1F1}11.7 & \cellcolor[HTML]{FEFEFF}52.8 & \cellcolor[HTML]{FEFCFC}0.7 & \cellcolor[HTML]{DDE7F8}21.4 & \cellcolor[HTML]{FDF4F4}50.3 & \cellcolor[HTML]{F8FAFE}2.8 & \cellcolor[HTML]{B4CBEF}\textbf{29.7} & \cellcolor[HTML]{D6E2F6}61.0 & \cellcolor[HTML]{F8FAFE}2.8 & \cellcolor[HTML]{FBEEEE}6.1 & 11.5 & 2.4 & 36.4 & 34.1 & \cellcolor{gray!12}$-$2.8 & \cellcolor{gray!12}+8.4 \\
Qwen2.5-Omni-7B$^{\P}$ & 21.4 & 52.0 & 0.7 & \cellcolor[HTML]{F9E4E4}15.9 & \cellcolor[HTML]{FEFCFC}51.3 & 0.7 & \cellcolor[HTML]{84A9E5}\textbf{75.2} & \cellcolor[HTML]{FEFEFF}52.3 & \cellcolor[HTML]{FCFDFE}1.4 & \cellcolor[HTML]{84A9E5}\textbf{49.7} & \cellcolor[HTML]{EDF2FB}55.7 & \cellcolor[HTML]{F8FAFE}2.1 & \cellcolor[HTML]{EB9F9F}34.1 & 65.4 & 26.8 & 14.4 & 8.3 & \cellcolor{gray!12}$-$5.5 & \cellcolor{gray!12}+3.7 \\
Qwen3.5-4B & 4.8 & 19.3 & 0.7 & \cellcolor[HTML]{D6E2F6}\textbf{13.1} & \cellcolor[HTML]{BED1F1}32.6 & 0.7 & \cellcolor[HTML]{E0EAF9}11.0 & \cellcolor[HTML]{FBEEEE}15.8 & 0.7 & \cellcolor[HTML]{84A9E5}\textbf{32.4} & \cellcolor[HTML]{95B5E9}40.8 & \cellcolor[HTML]{EBF1FB}4.8 & \cellcolor[HTML]{F4CCCC}3.8 & 20.5 & 0.0 & 20.5 & 18.9 & \cellcolor{gray!12}+8.3 & \cellcolor{gray!12}+21.5 \\
Qwen3-Omni-30B-A3B & 9.7 & 53.8 & 2.1 & \cellcolor[HTML]{E4ECF9}15.2 & \cellcolor[HTML]{F9FBFE}55.0 & 2.1 & \cellcolor[HTML]{FCF1F1}6.9 & \cellcolor[HTML]{FDF7F7}52.1 & \cellcolor[HTML]{FEF8F8}0.7 & \cellcolor[HTML]{C9D9F4}\textbf{20.7} & \cellcolor[HTML]{EAF1FB}58.0 & \cellcolor[HTML]{F5F8FD}4.1 & \cellcolor[HTML]{FCEFEF}0.0 & 5.1 & 0.0 & 28.0 & 28.0 & \cellcolor{gray!12}+5.5 & \cellcolor{gray!12}+4.2 \\
Qwen2.5-VL-7B & 6.2 & 50.6 & 0.7 & \cellcolor[HTML]{E7EEFA}11.0 & \cellcolor[HTML]{FEFEFF}50.9 & \cellcolor[HTML]{FCFDFE}1.4 & \cellcolor[HTML]{B1C8EE}\textbf{22.1} & \cellcolor[HTML]{FEFBFB}49.8 & 0.7 & \cellcolor[HTML]{A7C1EC}\textbf{24.1} & \cellcolor[HTML]{E8EFFA}55.3 & \cellcolor[HTML]{F5F8FD}2.8 & \cellcolor[HTML]{F9E5E5}3.0 & 11.5 & 0.0 & 21.2 & 19.7 & \cellcolor{gray!12}+4.8 & \cellcolor{gray!12}+4.7 \\
InternVL3.5-30B-A3B$^{\P}$ & 70.3 & 51.0 & 0.0 & \cellcolor[HTML]{E58484}\textbf{20.0} & \cellcolor[HTML]{FFFEFE}50.8 & 0.0 & \cellcolor[HTML]{88ACE6}\textbf{94.5} & \cellcolor[HTML]{FFFEFE}50.7 & 0.0 & \cellcolor[HTML]{F7DADA}62.8 & \cellcolor[HTML]{DDE7F8}58.0 & 0.0 & \cellcolor[HTML]{F5CECE}21.2 & 37.2 & 17.1 & 6.1 & 3.8 & \cellcolor{gray!12}$-$50.3 & \cellcolor{gray!12}+7.0 \\
Qwen2.5-VL-3B$^{\P}$ & 51.7 & 51.4 & 1.4 & \cellcolor[HTML]{F8FAFE}53.1 & \cellcolor[HTML]{FCFDFE}52.1 & \cellcolor[HTML]{FEF8F8}0.0 & \cellcolor[HTML]{84A9E5}\textbf{100.0} & \cellcolor[HTML]{FEFCFC}50.8 & \cellcolor[HTML]{FEFCFC}0.7 & \cellcolor[HTML]{84A9E5}\textbf{100.0} & \cellcolor[HTML]{FFFFFF}51.5 & 1.4 & \cellcolor[HTML]{F0BABA}65.9 & 88.5 & 56.1 & 8.3 & 3.8 & \cellcolor{gray!12}+1.4 & \cellcolor{gray!12}+0.1 \\
Gemma-3-12B & 22.1 & 51.3 & 2.1 & \cellcolor[HTML]{F3C5C5}\textbf{10.3} & \cellcolor[HTML]{FFFEFE}51.1 & \cellcolor[HTML]{FDF5F5}0.0 & \cellcolor[HTML]{84A9E5}\textbf{69.7} & \cellcolor[HTML]{D0DEF5}60.8 & \cellcolor[HTML]{D0DEF5}11.7 & \cellcolor[HTML]{BBD0F1}\textbf{35.9} & \cellcolor[HTML]{BFD2F1}64.3 & \cellcolor[HTML]{DAE5F7}9.7 & \cellcolor[HTML]{F4CACA}12.1 & 29.5 & 14.6 & 34.8 & 29.5 & \cellcolor{gray!12}$-$11.8 & \cellcolor{gray!12}+13.0 \\
InternVL3.5-8B$^{\P}$ & 47.6 & 53.4 & 0.0 & \cellcolor[HTML]{E99696}\textbf{26.2} & \cellcolor[HTML]{FEFAFA}52.4 & 0.0 & \cellcolor[HTML]{E4ECF9}53.1 & \cellcolor[HTML]{F5F8FD}55.5 & 0.0 & \cellcolor[HTML]{F1BBBB}\textbf{33.8} & \cellcolor[HTML]{EFF4FC}56.7 & 0.0 & \cellcolor[HTML]{FCF3F3}20.5 & 24.4 & 22.0 & 4.5 & 3.8 & \cellcolor{gray!12}$-$21.4 & \cellcolor{gray!12}+3.3 \\
Qwen2-VL-2B$^{\P}$ & 83.4 & 50.4 & 0.0 & \cellcolor[HTML]{F3C5C5}\textbf{71.7} & \cellcolor[HTML]{F1F5FC}53.2 & 0.0 & \cellcolor[HTML]{ADC6EE}\textbf{100.0} & 50.4 & 0.0 & \cellcolor[HTML]{ADC6EE}\textbf{100.0} & \cellcolor[HTML]{FDFEFF}50.8 & 0.0 & \cellcolor[HTML]{F7DADA}80.3 & 92.3 & 65.9 & 0.0 & 0.0 & \cellcolor{gray!12}$-$11.7 & \cellcolor{gray!12}+0.4 \\
Cosmos-Reason1-7B$^{\P}$ & 0.7 & 50.3 & 0.7 & 0.7 & 50.3 & \cellcolor[HTML]{FEFCFC}0.0 & \cellcolor[HTML]{FEFCFC}0.0 & \cellcolor[HTML]{FEFAFA}49.2 & \cellcolor[HTML]{FEFCFC}0.0 & \cellcolor[HTML]{FCFDFE}1.4 & \cellcolor[HTML]{FFFFFF}50.4 & 0.7 & \cellcolor[HTML]{FCF3F3}0.0 & 3.8 & 0.0 & 11.4 & 11.4 & \cellcolor{gray!12}0.0 & \cellcolor{gray!12}+0.1 \\
MiniCPM-o-4.5 & 1.4 & 45.1 & 0.0 & \cellcolor[HTML]{F5F8FD}3.4 & \cellcolor[HTML]{F8FAFD}46.6 & 0.0 & 1.4 & \cellcolor[HTML]{E8EFFA}49.7 & 0.0 & \cellcolor[HTML]{F5F8FD}3.4 & \cellcolor[HTML]{FEFEFF}45.4 & \cellcolor[HTML]{FCFDFE}0.7 & \cellcolor[HTML]{FAFCFE}1.5 & 0.0 & 0.0 & 7.6 & 6.8 & \cellcolor{gray!12}+2.0 & \cellcolor{gray!12}+0.3 \\
Qwen3-VL-2B$^{\P}$ & 0.0 & 50.0 & 0.0 & 0.0 & \cellcolor[HTML]{FEFAFA}49.0 & 0.0 & 0.0 & \cellcolor[HTML]{FFFEFE}49.7 & 0.0 & 0.0 & \cellcolor[HTML]{FFFDFD}49.6 & 0.0 & \cellcolor[HTML]{FEFBFB}0.0 & 1.3 & 0.0 & 9.1 & 9.1 & \cellcolor{gray!12}0.0 & \cellcolor{gray!12}$-$0.4 \\
BAGEL-7B-MoT$^{\P}$ & 0.0 & 50.0 & 0.0 & 0.0 & 50.0 & 0.0 & \cellcolor[HTML]{FCFDFE}0.7 & \cellcolor[HTML]{FFFEFE}49.8 & 0.0 & 0.0 & 50.0 & 0.0 & \cellcolor[HTML]{FEFEFF}10.6 & 10.3 & 2.4 & 0.8 & 0.8 & \cellcolor{gray!12}0.0 & \cellcolor{gray!12}0.0 \\
Janus-Pro-7B$^{\P}$ & 0.0 & 50.0 & 0.0 & 0.0 & 50.0 & 0.0 & 0.0 & 50.0 & 0.0 & 0.0 & 50.0 & 0.0 & 0.0 & 0.0 & 0.0 & 2.3 & 2.3 & \cellcolor{gray!12}0.0 & \cellcolor{gray!12}0.0 \\
Qwen3.5-2B$^{\P}$ & 4.8 & 50.9 & 0.0 & \cellcolor[HTML]{DAE5F7}\textbf{12.4} & \cellcolor[HTML]{F9FBFE}52.1 & 0.0 & \cellcolor[HTML]{B8CDF0}\textbf{19.3} & \cellcolor[HTML]{FBFCFE}51.8 & 0.0 & \cellcolor[HTML]{B1C8EE}\textbf{20.7} & \cellcolor[HTML]{F0F5FC}53.9 & 0.0 & \cellcolor[HTML]{FEFCFC}1.5 & 2.6 & 0.0 & 0.0 & 0.0 & \cellcolor{gray!12}+7.6 & \cellcolor{gray!12}+3.0 \\
Qwen3.5-0.8B$^{\P}$ & 0.7 & 48.8 & 0.0 & \cellcolor[HTML]{84A9E5}\textbf{31.7} & \cellcolor[HTML]{FEFBFB}48.0 & 0.0 & \cellcolor[HTML]{FEFCFC}0.0 & \cellcolor[HTML]{FAFCFE}49.8 & 0.0 & \cellcolor[HTML]{84A9E5}\textbf{70.3} & \cellcolor[HTML]{E2EBF9}54.6 & 0.0 & \cellcolor[HTML]{F8E0E0}62.9 & 73.1 & 82.9 & 0.0 & 0.0 & \cellcolor{gray!12}+31.0 & \cellcolor{gray!12}+5.8 \\
\bottomrule
\end{tabular}}
\end{center}
\end{table}

\begin{table}[h]
\caption{Self-reported recognition (\%, RECOG probe) of the 902 peripheral forgeries with a crop variant, under Full, Mirror (the full image flipped horizontally), and Crop (the edited region and its neighbours). Rows sorted by recognition under Full; $n=902$ for every model (release set, complete sweeps). Detection, KGR, and authentic accuracy under the same conditions are in Table~\ref{tab:mediation-detection}.}
\label{tab:mediation}
\begin{center}
\footnotesize
\setlength{\tabcolsep}{2.6pt}
\begin{tabular}{@{}l rrr r @{\hspace{4pt}} l rrr r@{}}
\toprule
Model & Full & Mirror & Crop & $n$ & Model & Full & Mirror & Crop & $n$ \\
\midrule
Qwen3.5-122B-A10B & 88.5 & 85.6 & 48.8 & 902 & Qwen3-Omni-30B-A3B & 66.3 & 64.6 & 33.6 & 902 \\
Kimi-K3 & 87.8 & 86.9 & 31.3 & 902 & Qwen3.6-35B-A3B & 62.0 & 58.2 & 25.2 & 902 \\
GPT-5.6 & 87.8 & 86.8 & 42.1 & 902 & Qwen3.8-27B$^{\dagger}$ & 60.2 & 51.4 & 17.0 & 902 \\
Gemini-3.8-Flash & 87.4 & 87.5 & 43.2 & 902 & Gemma-4-31B & 57.5 & 52.7 & 19.4 & 902 \\
Qwen3.5-35B-A3B & 86.3 & 85.8 & 48.4 & 902 & Qwen3-VL-8B & 56.4 & 52.5 & 11.8 & 902 \\
Qwen3.5-27B & 83.6 & 80.5 & 47.1 & 902 & Qwen3-VL-235B-A22B & 54.4 & 46.3 & 13.3 & 902 \\
Qwen3-VL-235B-A22B (think) & 81.8 & 78.2 & 35.9 & 902 & Qwen2.5-VL-7B & 52.9 & 41.9 & 12.4 & 902 \\
GLM-5.3-Flash & 81.8 & 78.0 & 25.9 & 902 & Qwen2-VL-2B & 41.6 & 38.0 & 8.2 & 902 \\
Qwen3.5-9B & 79.6 & 78.8 & 46.3 & 902 & BAGEL-7B-MoT & 33.9 & 26.5 & 6.7 & 902 \\
Qwen3.6-27B & 79.3 & 78.6 & 42.2 & 902 & Qwen2.5-Omni-7B & 33.5 & 28.0 & 8.8 & 902 \\
Qwen3-VL-32B (think) & 79.0 & 77.4 & 36.7 & 902 & Janus-Pro-7B & 32.8 & 29.2 & 10.6 & 902 \\
Qwen3-Omni-30B-A3B (think) & 77.2 & 75.7 & 42.5 & 902 & InternVL3.5-30B-A3B & 31.9 & 20.8 & 3.4 & 902 \\
Gemma-3-12B & 75.8 & 64.6 & 35.6 & 902 & Cosmos-Reason1-7B & 29.0 & 21.3 & 3.2 & 902 \\
Kimi-K2.6 & 75.5 & 74.1 & 21.0 & 902 & Qwen3-VL-2B & 28.4 & 20.1 & 2.3 & 902 \\
MiniMax-M3 & 75.3 & 69.4 & 22.8 & 902 & Qwen3.5-2B & 18.1 & 6.7 & 1.4 & 902 \\
Qwen3.5-4B & 72.9 & 71.6 & 42.0 & 902 & InternVL3.5-8B & 17.7 & 9.3 & 3.3 & 902 \\
Qwen3-VL-32B & 70.2 & 68.8 & 30.0 & 902 & Qwen2.5-VL-3B & 8.3 & 5.9 & 0.2 & 902 \\
MiniCPM-o-4.5 & 67.4 & 64.6 & 21.3 & 902 & Qwen3.5-0.8B & 0.1 & 0.0 & 0.0 & 902 \\
\bottomrule
\end{tabular}
\end{center}
\end{table}

\begin{table}[h]
\caption{Text-only, identity, and face probes for all models. TEXT reports the share of 435 named-insertion forgeries for which the model, given only the name, event, and date, correctly rules out the person's presence. The remaining columns use the 132 verified absent-celebrity items. L2 reports whether the erased person is named or whether the model says someone cannot be identified; L0 detection is then stratified by those L2 responses. FACE reports, for five models, how often the original face is accepted and the replacement rejected when shown in isolation (119 pairs), followed by full-image authenticity and naming rates on the pass set. $^{r}$GPT-5.6 declines isolated-face identification, so its FACE row reflects refusal rather than perception; --- indicates that the probe was not run. $^{\dagger}$Also the judge.}
\label{tab:identity}
\begin{center}
\scriptsize
\setlength{\tabcolsep}{3pt}
\begin{tabular}{@{}l r rr rr rrr@{}}
\toprule
 & TEXT (\%) & \multicolumn{2}{c}{L2 (\%)} & \multicolumn{2}{c}{L0 detection by L2 answer (\%)} & \multicolumn{3}{c}{FACE} \\
\cmidrule(lr){2-2}\cmidrule(lr){3-4}\cmidrule(lr){5-6}\cmidrule(lr){7-9}
Model & knows & names & cannot & named & cannot & pass & full real (\%) & named (\%) \\
\midrule
\multicolumn{9}{@{}l}{\emph{Proprietary}} \\
Gemini-3.8-Flash & 83.0 & 68.2 & 12.1 & 53.3 (90) & 62.5 (16) & 81/119 & 29.6 & 49.4 \\
GPT-5.6 & 80.0 & 38.6 & 42.4 & 17.6 (51) & 16.1 (56) & 4/119$^{r}$ & 75.0 & 75.0 \\
\addlinespace[2pt]
\multicolumn{9}{@{}l}{\emph{Open weights}} \\
GLM-5.3-Flash & 83.5 & 43.2 & 13.6 & 32.7 (55) & 27.8 (18) & --- & --- & --- \\
Kimi-K3 & 83.9 & 62.1 & 31.1 & 22.0 (82) & 43.9 (41) & 55/119 & 63.6 & 30.9 \\
Kimi-K2.6 & 83.0 & 72.7 & 18.2 & 19.8 (96) & 25.0 (24) & --- & --- & --- \\
Qwen3.5-122B-A10B & 92.4 & 48.5 & 17.4 & 23.4 (64) & 39.1 (23) & --- & --- & --- \\
Qwen3-VL-235B-A22B & 99.3 & 58.3 & 19.7 & 10.4 (77) & 11.5 (26) & --- & --- & --- \\
Qwen3.6-27B & 90.3 & 43.2 & 18.9 & 30.9 (55) & 29.2 (24) & 47/119 & 69.8 & 30.2 \\
Qwen3.5-27B & 92.9 & 31.1 & 12.1 & 18.9 (37) & 18.8 (16) & --- & --- & --- \\
Qwen3-VL-235B-A22B (think) & 91.2 & 56.1 & 25.0 & 6.8 (74) & 12.1 (33) & --- & --- & --- \\
Qwen3-VL-8B & 99.5 & 34.1 & 9.1 & 35.6 (45) & 8.3 (12) & --- & --- & --- \\
Qwen3.8-27B$^{\dagger}$ & 80.7 & 47.7 & 49.2 & 28.6 (63) & 21.5 (65) & 50/119 & 68.0 & 12.0 \\
Qwen3.5-35B-A3B & 94.8 & 31.8 & 9.1 & 26.3 (38) & 18.2 (11) & --- & --- & --- \\
Qwen3.6-35B-A3B & 89.9 & 30.3 & 6.8 & 23.5 (34) & 0.0 (6) & --- & --- & --- \\
Qwen3-VL-32B (think) & 91.2 & 53.0 & 24.2 & 15.7 (70) & 9.4 (32) & --- & --- & --- \\
Qwen3-VL-32B & 99.3 & 47.0 & 26.5 & 12.9 (62) & 8.6 (35) & --- & --- & --- \\
Qwen3-Omni-30B-A3B (think) & 93.7 & 46.2 & 25.0 & 8.3 (60) & 6.1 (33) & --- & --- & --- \\
Gemma-4-31B & 86.0 & 40.2 & 24.2 & 64.2 (53) & 68.8 (32) & --- & --- & --- \\
Qwen3.5-9B & 96.6 & 26.5 & 12.1 & 11.5 (26) & 20.0 (10) & --- & --- & --- \\
MiniMax-M3 & 87.6 & 18.9 & 81.1 & 12.0 (25) & 4.7 (107) & --- & --- & --- \\
Qwen2.5-Omni-7B & 97.2 & 22.0 & 45.5 & 37.9 (29) & 35.0 (60) & --- & --- & --- \\
Qwen3.5-4B & 98.6 & 20.5 & 16.7 & 12.5 (16) & 25.0 (12) & --- & --- & --- \\
Qwen3-Omni-30B-A3B & 97.5 & 35.6 & 3.8 & 0.0 (47) & 0.0 (5) & --- & --- & --- \\
Qwen2.5-VL-7B & 99.3 & 22.0 & 45.5 & 6.9 (29) & 1.7 (60) & --- & --- & --- \\
InternVL3.5-30B-A3B & 96.5 & 17.4 & 53.0 & 30.4 (23) & 21.4 (70) & --- & --- & --- \\
Qwen2.5-VL-3B & 99.1 & 23.5 & 24.2 & 48.4 (31) & 75.0 (32) & --- & --- & --- \\
Gemma-3-12B & 93.8 & 31.1 & 51.5 & 22.0 (41) & 10.3 (68) & --- & --- & --- \\
InternVL3.5-8B & 89.4 & 13.6 & 32.6 & 38.9 (18) & 20.9 (43) & --- & --- & --- \\
Qwen2-VL-2B & 99.8 & 12.9 & 0.8 & 100.0 (17) & 100.0 (1) & --- & --- & --- \\
Cosmos-Reason1-7B & 79.8 & 15.2 & 0.0 & 0.0 (20) & --- (0) & --- & --- & --- \\
MiniCPM-o-4.5 & 32.4 & 35.6 & 56.1 & 2.2 (46) & 1.4 (73) & --- & --- & --- \\
Qwen3-VL-2B & 16.6 & 14.4 & 1.5 & 0.0 (19) & 0.0 (2) & --- & --- & --- \\
BAGEL-7B-MoT & 0.0 & 1.5 & 9.8 & 0.0 (2) & 15.4 (13) & --- & --- & --- \\
Janus-Pro-7B & 83.9 & 2.3 & 62.9 & 0.0 (3) & 0.0 (83) & --- & --- & --- \\
Qwen3.5-2B & 100.0 & 25.0 & 8.3 & 3.0 (33) & 0.0 (11) & --- & --- & --- \\
Qwen3.5-0.8B & 99.1 & 12.9 & 1.5 & 82.4 (17) & 50.0 (2) & --- & --- & --- \\
\bottomrule
\end{tabular}
\end{center}
\end{table}

\begin{table}[h]
\caption{The remembered people follow the composition. On the 116 peripheral people-family forgeries whose scene has named principal figures, the identity probe (L2) is asked on the full image and on the crop around the edited region. Full / Crop: share of items whose answer names at least one principal figure of the scene (strict denominators); only Full / only Crop: items named under one condition only; $p$: exact McNemar test. Rows in the order of Table~\ref{tab:main}. $^{\dagger}$Also the judge.}
\label{tab:cropnaming}
\begin{center}
\footnotesize
\setlength{\tabcolsep}{5pt}
\begin{tabular}{@{}l r rr r r@{}}
\toprule
Model & pairs & Full & Crop & only Full / only Crop & $p$ \\
\midrule
\multicolumn{6}{@{}l}{\emph{Proprietary}} \\
Gemini-3.8-Flash & 116 & 92.2 & 44.0 & 56/0 & $<$0.0001 \\
GPT-5.6$^{\S}$ & 66 & 56.0 & 7.8 & 37/1 & $<$0.0001 \\
\addlinespace[2pt]
\multicolumn{6}{@{}l}{\emph{Open weights}} \\
GLM-5.3-Flash & 95 & 56.0 & 6.9 & 52/1 & $<$0.0001 \\
Kimi-K3 & 116 & 84.5 & 18.1 & 78/1 & $<$0.0001 \\
Kimi-K2.6 & 116 & 86.2 & 26.7 & 69/0 & $<$0.0001 \\
Qwen3.5-122B-A10B & 39 & 49.1 & 21.6 & 15/0 & 0.0001 \\
Qwen3-VL-235B-A22B & 114 & 76.7 & 27.6 & 57/1 & $<$0.0001 \\
Qwen3.6-27B & 30 & 35.3 & 9.5 & 14/2 & 0.0042 \\
Qwen3.5-27B & 12 & 22.4 & 6.9 & 5/0 & 0.0625 \\
Qwen3-VL-235B-A22B (think) & 116 & 79.3 & 24.1 & 65/1 & $<$0.0001 \\
Qwen3-VL-8B & 46 & 29.3 & 15.5 & 20/2 & 0.0001 \\
Qwen3.8-27B$^{\dagger}$ & 115 & 71.5 & 19.8 & 61/2 & $<$0.0001 \\
Qwen3.5-35B-A3B & 5 & 17.2 & 5.2 & 4/0 & 0.1250 \\
Qwen3.6-35B-A3B & 14 & 17.2 & 8.6 & 8/0 & 0.0078 \\
Qwen3-VL-32B (think) & 106 & 62.9 & 18.1 & 52/1 & $<$0.0001 \\
Qwen3-VL-32B & 109 & 61.2 & 16.4 & 54/1 & $<$0.0001 \\
Qwen3-Omni-30B-A3B (think) & 95 & 69.8 & 24.1 & 48/1 & $<$0.0001 \\
Gemma-4-31B & 116 & 56.9 & 9.5 & 58/3 & $<$0.0001 \\
Qwen3.5-9B & 6 & 12.9 & 6.0 & 3/0 & 0.2500 \\
MiniMax-M3 & 116 & 37.1 & 1.7 & 41/0 & $<$0.0001 \\
Qwen2.5-Omni-7B & 100 & 37.1 & 3.5 & 38/1 & $<$0.0001 \\
Qwen3.5-4B & 7 & 10.3 & 5.2 & 3/0 & 0.2500 \\
Qwen3-Omni-30B-A3B & 104 & 61.2 & 18.1 & 51/1 & $<$0.0001 \\
Qwen2.5-VL-7B & 116 & 44.0 & 2.6 & 49/1 & $<$0.0001 \\
InternVL3.5-30B-A3B & 115 & 25.9 & 3.5 & 26/0 & $<$0.0001 \\
Qwen2.5-VL-3B & 116 & 30.2 & 0.9 & 34/0 & $<$0.0001 \\
Gemma-3-12B & 116 & 69.0 & 20.7 & 57/1 & $<$0.0001 \\
InternVL3.5-8B & 116 & 17.2 & 0.9 & 19/0 & $<$0.0001 \\
Qwen2-VL-2B & 61 & 17.2 & 6.9 & 14/1 & 0.0010 \\
Cosmos-Reason1-7B & 113 & 27.6 & 1.7 & 31/1 & $<$0.0001 \\
MiniCPM-o-4.5 & 116 & 64.7 & 14.7 & 58/0 & $<$0.0001 \\
Qwen3-VL-2B & 12 & 8.6 & 3.5 & 7/0 & 0.0156 \\
BAGEL-7B-MoT & 31 & 5.2 & 0.9 & 3/0 & 0.2500 \\
Janus-Pro-7B & 113 & 0.9 & 0.0 & 1/0 & 1.00 \\
Qwen3.5-2B & 84 & 32.8 & 11.2 & 30/3 & $<$0.0001 \\
Qwen3.5-0.8B & 90 & 19.8 & 2.6 & 21/1 & $<$0.0001 \\
\bottomrule
\end{tabular}
\end{center}
\end{table}

\section{Item-level association between recognition and detection}
\label{app:itemreg}
The de-recognition experiment manipulates recognizability directly. We also ask whether the same relationship appears naturally across release-set items. For each forgery, we pair the model's RECOG response under Full with its prompted-verification verdict. We then fit a logistic regression per model, predicting detection from recognition, edit saliency, their interaction, and family and domain fixed effects, with scene-clustered standard errors. A pooled analysis includes the 24 models that recognize at least half of the images, with model fixed effects and 13,660 model--item observations for the people family.

Table~\ref{tab:itemreg} reports the corresponding rates and regression coefficients. For the people family, recognition is associated with lower detection (pooled odds ratio 0.65, 95\% CI 0.51--0.82, $p=0.0004$); the coefficient is negative for 20 of the 24 recognizing models and individually significant for eight. No comparable association appears for the class-level families (OR 0.89, 95\% CI 0.67--1.18, $p=0.4101$). This observational result supports the same people-specific pattern found in the controlled de-recognition experiment, although the mechanism claim rests on the manipulations in \S\ref{sec:results-recognition}.

\begin{table}[h]
\caption{Prompted-verification detection as a function of self-reported scene recognition (RECOG under Full), shown per model on the release set with strict scoring. For the people family (K1), we report detection on recognized and unrecognized scenes, their difference, and the recognition coefficient from a logistic regression controlling for saliency, family, and domain with scene-clustered errors. The same quantities are reported for class-level families (K2--K6). Anchor-free controls are assigned to the family of their edit specification. $^{\P}$Recognizes fewer than half of the images and is excluded from the pooled model; --- denotes separated or empty cells. $^{\dagger}$Also the judge.}
\label{tab:itemreg}
\begin{center}
\scriptsize
\setlength{\tabcolsep}{2.6pt}
\begin{tabular}{@{}l r rr r rr rr rr@{}}
\toprule
 & RECOG & \multicolumn{5}{c}{People family (K1)} & \multicolumn{4}{c}{Class-level families (K2--K6)} \\
\cmidrule(lr){3-7}\cmidrule(lr){8-11}
Model & Full (\%) & recognized & not & $\Delta$ & OR & $p$ & recognized & not & OR & $p$ \\
\midrule
\multicolumn{11}{@{}l}{\emph{Proprietary}} \\
Gemini-3.8-Flash & 78.2 & 75.4 (391) & 87.7 (203) & $-$12.2 & 0.67 & 0.1766 & 79.7 (788) & 89.6 (125) & 0.29 & 0.0529 \\
GPT-5.6 & 76.8 & 37.9 (116) & 63.8 (69) & $-$25.8 & 0.58 & 0.3597 & 66.7 (129) & 40.0 (5) & 3.36 & 0.0860 \\
\addlinespace[2pt]
\multicolumn{11}{@{}l}{\emph{Open weights}} \\
GLM-5.3-Flash & 74.2 & 60.0 (365) & 56.3 (229) & +3.7 & 1.15 & 0.5612 & 61.9 (753) & 61.9 (160) & 1.30 & 0.5536 \\
Kimi-K3 & 81.2 & 53.6 (414) & 72.2 (180) & $-$18.6 & 0.58 & 0.0404 & 61.5 (810) & 77.7 (103) & 1.97 & 0.3206 \\
Kimi-K2.6 & 67.5 & 37.5 (296) & 64.8 (298) & $-$27.3 & 0.35 & $<$0.0001 & 56.1 (722) & 73.3 (191) & 0.73 & 0.5750 \\
Qwen3.5-122B-A10B & 82.3 & 52.8 (426) & 44.0 (168) & +8.8 & 1.71 & 0.0699 & 62.4 (814) & 62.6 (99) & 5.09 & 0.0423 \\
Qwen3-VL-235B-A22B & 51.2 & 48.2 (249) & 82.3 (345) & $-$34.1 & 0.18 & $<$0.0001 & 44.4 (523) & 83.1 (390) & 0.31 & 0.0097 \\
Qwen3.6-27B & 73.3 & 57.8 (377) & 58.1 (217) & $-$0.2 & 0.81 & 0.4241 & 64.2 (727) & 65.6 (186) & 1.37 & 0.3984 \\
Qwen3.5-27B & 76.5 & 45.9 (388) & 39.3 (206) & +6.6 & 1.02 & 0.9279 & 61.6 (765) & 47.3 (148) & 3.71 & 0.0172 \\
Qwen3-VL-235B-A22B (think) & 75.0 & 45.0 (376) & 51.8 (218) & $-$6.9 & 0.58 & 0.0264 & 54.4 (754) & 68.5 (159) & 0.80 & 0.6988 \\
Qwen3-VL-8B & 53.5 & 69.8 (268) & 78.5 (326) & $-$8.8 & 0.30 & 0.0001 & 58.7 (538) & 81.6 (375) & 0.27 & 0.0015 \\
Qwen3.8-27B$^{\dagger}$ & 52.2 & 68.0 (253) & 59.8 (341) & +8.2 & 1.01 & 0.9829 & 69.5 (534) & 65.2 (379) & 2.08 & 0.0347 \\
Qwen3.5-35B-A3B & 80.4 & 30.0 (423) & 29.2 (171) & +0.8 & 0.62 & 0.0877 & 47.1 (789) & 46.8 (124) & 0.84 & 0.7390 \\
Qwen3.6-35B-A3B & 55.5 & 25.4 (248) & 34.4 (346) & $-$9.0 & 0.81 & 0.3912 & 47.6 (588) & 48.9 (325) & 0.56 & 0.1712 \\
Qwen3-VL-32B (think) & 73.2 & 46.4 (371) & 52.9 (223) & $-$6.6 & 0.80 & 0.4159 & 54.5 (732) & 59.7 (181) & 0.89 & 0.7800 \\
Qwen3-VL-32B & 66.2 & 41.8 (342) & 66.7 (252) & $-$24.9 & 0.41 & 0.0007 & 43.8 (656) & 78.2 (257) & 0.27 & 0.0085 \\
Qwen3-Omni-30B-A3B (think) & 70.7 & 37.4 (361) & 41.2 (233) & $-$3.8 & 0.69 & 0.2268 & 43.6 (704) & 47.9 (209) & 0.22 & 0.0142 \\
Gemma-4-31B & 51.6 & 82.9 (251) & 84.0 (343) & $-$1.1 & 0.69 & 0.2553 & 82.3 (527) & 86.5 (386) & 0.84 & 0.6675 \\
Qwen3.5-9B & 74.5 & 19.3 (383) & 16.3 (209) & +3.0 & 0.88 & 0.7027 & 34.2 (738) & 32.2 (174) & --- & --- \\
MiniMax-M3 & 67.0 & 25.1 (303) & 25.8 (291) & $-$0.7 & 0.80 & 0.3452 & 29.3 (706) & 29.5 (207) & 2.72 & 0.1073 \\
Qwen2.5-Omni-7B$^{\P}$ & 29.3 & 18.2 (143) & 59.4 (451) & $-$41.2 & 0.32 & 0.0002 & 23.1 (299) & 54.4 (614) & 0.43 & 0.0455 \\
Qwen3.5-4B & 68.8 & 13.4 (372) & 12.6 (222) & +0.8 & 0.90 & 0.7701 & 26.7 (664) & 26.9 (249) & 1.80 & 0.3775 \\
Qwen3-Omni-30B-A3B & 61.3 & 10.3 (321) & 20.9 (273) & $-$10.6 & 0.18 & $<$0.0001 & 11.6 (603) & 48.7 (310) & 0.12 & 0.0995 \\
Qwen2.5-VL-7B$^{\P}$ & 48.2 & 12.3 (220) & 33.2 (374) & $-$20.9 & 0.30 & $<$0.0001 & 16.4 (506) & 38.6 (407) & 0.18 & 0.0674 \\
InternVL3.5-30B-A3B$^{\P}$ & 27.5 & 54.9 (113) & 77.3 (481) & $-$22.5 & 0.40 & 0.0081 & 61.1 (301) & 78.1 (612) & 0.25 & 0.0954 \\
Qwen2.5-VL-3B$^{\P}$ & 7.7 & 2.2 (45) & 72.7 (549) & $-$70.5 & 0.02 & 0.0001 & 18.3 (71) & 62.7 (842) & 0.15 & 0.1042 \\
Gemma-3-12B & 68.7 & 26.3 (338) & 52.7 (256) & $-$26.4 & 0.37 & 0.0003 & 29.6 (697) & 46.8 (216) & 0.28 & 0.0605 \\
InternVL3.5-8B$^{\P}$ & 16.5 & 50.0 (82) & 58.4 (512) & $-$8.4 & 0.64 & 0.2327 & 29.3 (167) & 50.0 (746) & 0.80 & 0.7046 \\
Qwen2-VL-2B$^{\P}$ & 38.6 & 85.1 (188) & 92.1 (406) & $-$7.0 & 0.90 & 0.8011 & 82.2 (393) & 91.3 (520) & 0.51 & 0.2239 \\
Cosmos-Reason1-7B$^{\P}$ & 27.9 & 2.1 (142) & 1.8 (452) & +0.3 & 1.47 & 0.6781 & 3.2 (278) & 7.6 (635) & --- & --- \\
MiniCPM-o-4.5 & 61.5 & 1.4 (296) & 4.7 (298) & $-$3.4 & 0.74 & 0.6712 & 4.6 (631) & 4.6 (282) & --- & --- \\
Qwen3-VL-2B$^{\P}$ & 23.2 & 0.0 (122) & 0.4 (472) & $-$0.4 & --- & --- & 0.0 (227) & 1.0 (686) & --- & --- \\
BAGEL-7B-MoT$^{\P}$ & 31.0 & 0.0 (127) & 0.4 (467) & $-$0.4 & --- & --- & 0.9 (340) & 0.7 (573) & 0.23 & 0.1463 \\
Janus-Pro-7B$^{\P}$ & 27.2 & 0.0 (143) & 0.0 (451) & 0.0 & --- & --- & 0.0 (267) & 0.0 (646) & --- & --- \\
Qwen3.5-2B$^{\P}$ & 13.4 & 15.7 (51) & 8.7 (543) & +7.0 & 1.28 & 0.7588 & 11.9 (151) & 15.5 (762) & --- & --- \\
Qwen3.5-0.8B$^{\P}$ & 0.1 & --- (0) & 4.2 (594) & --- & --- & --- & 0.0 (1) & 3.1 (912) & --- & --- \\
\bottomrule
\end{tabular}
\end{center}
\end{table}

\section{Caption controls: the wrong event and the class-level families}
\label{app:captions}

\paragraph{A wrong caption.}
One possible explanation for the drop under a true caption is that models treat any confident caption as evidence that the image is authentic. If so, a wrong caption should also reduce detection. We test this with a third condition on the two proprietary models, pairing the full image with the same caption template but naming a different canonical event from the same domain and within fifteen years, at a different location and with no shared participants. The wrong event is chosen so that it neither names nor implies the inserted identity.

The results do not support a simple endorsement effect (Table~\ref{tab:wrongcaption}). Detection rises to 93.8 and 91.0, while authentic accuracy falls to 19.4 and 16.1; KGR against the recorded contradiction does not increase. The models adopt the stated event in 80.8\% and 84.1\% of answers and explicitly dispute it in 30.1\% and 10.9\%. This pattern suggests that the models often detect a mismatch between the caption and the canonical scene they recognize, rather than the underlying image edit.

\begin{table}[h]
\caption{Wrong-caption control on the 145 peripheral people-family forgeries and 93 authentic originals under prompted verification and strict scoring. We compare Full, Full+Caption, and Full+Wrong caption, where the wrong caption names a different same-era event from the same domain. Det, Auth, Bal, and KGR are defined as in the main text; $p$ is the exact McNemar test against Full. Uptake and dispute report whether the model adopts or explicitly rejects the wrong event.}
\label{tab:wrongcaption}
\begin{center}
\scriptsize
\setlength{\tabcolsep}{2.4pt}
\begin{tabular}{@{}l rrrr rrrr rrrr rr rr@{}}
\toprule
 & \multicolumn{4}{c}{Full} & \multicolumn{4}{c}{Full+Caption} & \multicolumn{4}{c}{Full+Wrong caption} & \multicolumn{2}{c}{$p$ vs.\ Full} & \multicolumn{2}{c}{Wrong caption} \\
\cmidrule(lr){2-5}\cmidrule(lr){6-9}\cmidrule(lr){10-13}\cmidrule(lr){14-15}\cmidrule(lr){16-17}
Model & Det & Auth & Bal & KGR & Det & Auth & Bal & KGR & Det & Auth & Bal & KGR & Caption & Wrong & uptake & dispute \\
\midrule
Gemini-3.8-Flash & 73.1 & 84.9 & 79.0 & 35.2 & 77.9 & 73.1 & 75.5 & 33.1 & 93.8 & 19.4 & 56.6 & 31.0 & 0.1671 & $<$0.0001 & 80.8 & 30.1 \\
GPT-5.6 & 53.8 & 84.9 & 69.4 & 14.5 & 57.2 & 86.0 & 71.6 & 13.1 & 91.0 & 16.1 & 53.6 & 9.7 & 0.4421 & $<$0.0001 & 84.1 & 10.9 \\
\bottomrule
\end{tabular}
\end{center}
\end{table}

\paragraph{Class-level families under the four conditions.}
Table~\ref{tab:conditions} focuses on the people family because the recognition-penalty claim concerns identity. Crop and Crop+Caption are available for every peripheral forgery, and for the two proprietary models we also run Full+Caption on the other families (Table~\ref{tab:families}; K6 has no peripheral items by construction). For class-level facts, the true caption does not reduce detection in any family. For GPT-5.6 it significantly increases detection for visible-text forgeries (K4 +13.9, $p=0.0015$) and event-attribute forgeries (K5 +13.1, $p=0.0026$), with KGR changing from 69.4 to 80.6 and from 50.3 to 63.4, respectively.

\begin{table}[h]
\caption{Detection and KGR under the four de-recognition conditions by knowledge family for the two proprietary models (peripheral forgeries on the release set; prompted verification; strict scoring). $n$: forgeries; F, FC, C, CC: Full, Full+Caption, Crop, Crop+Caption; Cap.: Det(FC)$-$Det(F) with its exact McNemar $p$; $p_{\mathrm{CC}}$: McNemar $p$ of Det(CC) against Det(F).}
\label{tab:families}
\begin{center}
\scriptsize
\setlength{\tabcolsep}{4pt}
\begin{tabular}{@{}l l r rrrr rrrr rr r@{}}
\toprule
 & & & \multicolumn{4}{c}{Det} & \multicolumn{4}{c}{KGR} & \multicolumn{2}{c}{Cap.} & \\
\cmidrule(lr){4-7}\cmidrule(lr){8-11}\cmidrule(lr){12-13}
Model & Fam. & $n$ & F & FC & C & CC & F & FC & C & CC & $\Delta$ & $p$ & $p_{\mathrm{CC}}$ \\
\midrule
Gemini-3.8-Flash & K1 & 145 & 73.1 & 77.9 & 79.3 & 86.2 & 35.2 & 33.1 & 46.2 & 44.8 & +4.8 & 0.1671 & 0.0009 \\
 & K2 & 294 & 96.3 & 96.3 & 94.6 & 100.0 & 84.7 & 83.3 & 85.4 & 91.5 & 0.0 & 1.00 & 0.0010 \\
 & K3 & 89 & 85.4 & 88.8 & 91.0 & 97.8 & 69.7 & 62.9 & 79.8 & 88.8 & +3.4 & 0.4531 & 0.0010 \\
 & K4 & 108 & 90.7 & 96.3 & 75.9 & 90.7 & 87.0 & 94.4 & 66.7 & 84.3 & +5.6 & 0.1094 & 1.00 \\
 & K5 & 145 & 86.2 & 90.3 & 83.4 & 95.9 & 77.9 & 85.5 & 73.1 & 87.6 & +4.1 & 0.1796 & 0.0013 \\
\addlinespace[2pt]
GPT-5.6 & K1 & 145 & 53.8 & 57.2 & 56.6 & 69.7 & 14.5 & 13.1 & 13.1 & 16.6 & +3.4 & 0.4421 & 0.0022 \\
 & K2 & 294 & 89.1 & 89.1 & 89.8 & 94.6 & 68.0 & 68.3 & 71.1 & 74.8 & 0.0 & 1.00 & 0.0052 \\
 & K3 & 89 & 60.7 & 67.4 & 59.6 & 79.8 & 39.3 & 40.4 & 38.2 & 48.3 & +6.7 & 0.1796 & 0.0005 \\
 & K4 & 108 & 75.0 & 88.9 & 67.6 & 81.5 & 69.4 & 80.6 & 61.1 & 70.4 & +13.9 & 0.0015 & 0.2649 \\
 & K5 & 145 & 59.3 & 72.4 & 60.0 & 73.1 & 50.3 & 63.4 & 49.0 & 66.2 & +13.1 & 0.0026 & 0.0037 \\
\bottomrule
\end{tabular}
\end{center}
\end{table}

\section{Additional validity checks}
\label{app:validity}
\paragraph{Authentic images by domain and by recognition.}
Table~\ref{tab:authdomain} gives accuracy on the 474 authentic images under prompted verification by domain and by whether the model recognizes the image (RECOG under Full).
The frontier models keep authentic accuracy above 88\% in every photographic domain and drop only on the twelve meme templates; the mid-sized Qwen models lose most on film stills and sports photographs.
For most models a recognized authentic image is called authentic more often than an unrecognized one (Kimi-K3 by +9.1 points, Gemini-3.8-Flash +13.8, Qwen3-VL-235B-A22B +18.6).
This is the complement of the effect on forgeries; recognition moves verdicts toward the remembered label in both directions.
\paragraph{Saliency of the edit.}
The de-recognition study uses peripheral forgeries; Table~\ref{tab:saliency} reports detection and KGR by saliency class over all families (main figure C1, 605 items; secondary figure C2, 72; background C3, 830).
Within the people family a replaced main figure is far easier to detect than a background insertion (Kimi-K3 69.9 against 35.9 under prompted verification, Gemini-3.8-Flash 83.9 against 73.1).
Pooled over families the C1--C3 difference is within five points for most models, because the background class is dominated by the class-level families, where KGR is higher.
\paragraph{Acceptance threshold and the E4 level.}
Of the 397 people-family forgeries with a face-similarity score, 41 lie within $\pm0.05$ of their acceptance threshold.
Dropping them changes people-family detection under prompted verification by at most 1.7 points and KGR by at most 0.5 for any model (Table~\ref{tab:efour}).
The same table lists the share of forgeries graded E4 (the contradiction stated) next to E3+ (KGR). E4 rates are lower, while the leading models remain broadly similar: under prompted verification, the E4 rate is 37.2 for Gemini-3.8-Flash, 29.7 for Kimi-K3, and 26.3 for GPT-5.6.

\begin{table}[h]
\caption{Accuracy on authentic images (\%, prompted verification, strict scoring) by domain (A1 historical photographs 104, A2 film stills 60, A3 sports 184, A4 paintings 47, A5 political news 67, A6 meme templates 12 authentic images), and by whether the model recognizes the image (RECOG under Full: share recognized, accuracy on recognized and unrecognized images, difference). $^{\dagger}$Also the judge.}
\label{tab:authdomain}
\begin{center}
\scriptsize
\setlength{\tabcolsep}{3pt}
\begin{tabular}{@{}l r rrrrrr r rr r@{}}
\toprule
 & \multicolumn{7}{c}{Authentic accuracy, L1 (\%)} & \multicolumn{4}{c}{By recognition (L1)} \\
\cmidrule(lr){2-8}\cmidrule(lr){9-12}
Model & All & A1 & A2 & A3 & A4 & A5 & A6 & RECOG & recognized & not & $\Delta$ \\
\midrule
\multicolumn{12}{@{}l}{\emph{Proprietary}} \\
Gemini-3.8-Flash & 88.2 & 93.3 & 83.3 & 88.6 & 91.5 & 89.6 & 41.7 & 91.4 & 89.4 & 75.6 & +13.8 \\
GPT-5.6 & 91.6 & 95.2 & 95.0 & 91.8 & 91.5 & 91.0 & 41.7 & 86.1 & 91.9 & 89.4 & +2.5 \\
\addlinespace[2pt]
\multicolumn{12}{@{}l}{\emph{Open weights}} \\
GLM-5.3-Flash & 85.9 & 92.3 & 86.7 & 83.7 & 85.1 & 82.1 & 83.3 & 83.3 & 89.4 & 68.4 & +21.0 \\
Kimi-K3 & 92.2 & 96.2 & 96.7 & 88.0 & 100.0 & 94.0 & 58.3 & 88.0 & 93.3 & 84.2 & +9.1 \\
Kimi-K2.6 & 90.1 & 94.2 & 98.3 & 87.0 & 93.6 & 91.0 & 41.7 & 79.1 & 92.3 & 81.8 & +10.5 \\
Qwen3.5-122B-A10B & 77.4 & 89.4 & 86.7 & 69.6 & 76.6 & 71.6 & 83.3 & 85.0 & 77.7 & 76.1 & +1.6 \\
Qwen3-VL-235B-A22B & 70.0 & 85.6 & 38.3 & 66.3 & 93.6 & 74.6 & 33.3 & 66.0 & 76.4 & 57.8 & +18.6 \\
Qwen3.6-27B & 65.6 & 84.6 & 78.3 & 60.3 & 53.2 & 50.7 & 50.0 & 76.2 & 65.7 & 65.5 & +0.2 \\
Qwen3.5-27B & 56.3 & 83.7 & 50.0 & 50.5 & 51.1 & 40.3 & 50.0 & 80.4 & 57.2 & 52.7 & +4.5 \\
Qwen3-VL-235B-A22B (think) & 78.9 & 85.6 & 95.0 & 69.0 & 89.4 & 76.1 & 66.7 & 80.6 & 79.8 & 75.0 & +4.8 \\
Qwen3-VL-8B & 50.4 & 55.8 & 80.0 & 43.5 & 85.1 & 16.4 & 16.7 & 62.9 & 52.0 & 47.7 & +4.3 \\
Qwen3.8-27B$^{\dagger}$ & 62.0 & 79.8 & 83.3 & 52.7 & 61.7 & 43.3 & 50.0 & 63.9 & 58.1 & 69.0 & $-$10.9 \\
Qwen3.5-35B-A3B & 58.4 & 79.8 & 43.3 & 54.9 & 63.8 & 47.8 & 41.7 & 85.0 & 60.8 & 45.1 & +15.7 \\
Qwen3.6-35B-A3B & 61.8 & 84.6 & 41.7 & 61.4 & 61.7 & 49.3 & 41.7 & 63.7 & 70.2 & 47.1 & +23.1 \\
Qwen3-VL-32B (think) & 79.1 & 86.5 & 85.0 & 75.5 & 83.0 & 71.6 & 66.7 & 79.5 & 81.7 & 69.1 & +12.6 \\
Qwen3-VL-32B & 80.4 & 87.5 & 65.0 & 83.2 & 93.6 & 71.6 & 50.0 & 77.6 & 82.9 & 71.7 & +11.2 \\
Qwen3-Omni-30B-A3B (think) & 81.2 & 91.3 & 93.3 & 70.1 & 95.7 & 82.1 & 41.7 & 77.4 & 80.9 & 82.2 & $-$1.3 \\
Gemma-4-31B & 34.0 & 34.6 & 41.7 & 27.2 & 51.1 & 35.8 & 16.7 & 60.5 & 32.1 & 36.9 & $-$4.8 \\
Qwen3.5-9B & 53.2 & 71.2 & 55.0 & 47.8 & 61.7 & 37.3 & 25.0 & 77.8 & 54.5 & 48.6 & +5.9 \\
MiniMax-M3 & 91.1 & 93.3 & 91.7 & 91.8 & 93.6 & 89.6 & 58.3 & 73.0 & 92.8 & 86.7 & +6.1 \\
Qwen2.5-Omni-7B & 69.8 & 78.8 & 28.3 & 69.0 & 100.0 & 83.6 & 16.7 & 26.2 & 86.3 & 64.0 & +22.3 \\
Qwen3.5-4B & 46.2 & 56.7 & 41.7 & 42.9 & 55.3 & 31.3 & 75.0 & 70.0 & 47.6 & 43.0 & +4.6 \\
Qwen3-Omni-30B-A3B & 97.9 & 100.0 & 100.0 & 96.2 & 100.0 & 95.5 & 100.0 & 73.2 & 98.0 & 97.6 & +0.4 \\
Qwen2.5-VL-7B & 89.9 & 90.4 & 90.0 & 88.6 & 100.0 & 92.5 & 50.0 & 47.7 & 93.8 & 86.3 & +7.5 \\
InternVL3.5-30B-A3B & 43.7 & 49.0 & 6.7 & 60.9 & 14.9 & 47.8 & 8.3 & 28.5 & 63.0 & 36.0 & +27.0 \\
Qwen2.5-VL-3B & 50.4 & 53.8 & 11.7 & 65.2 & 51.1 & 47.8 & 0.0 & 9.1 & 97.7 & 45.7 & +52.0 \\
Gemma-3-12B & 72.4 & 70.2 & 58.3 & 73.4 & 100.0 & 74.6 & 25.0 & 71.9 & 73.9 & 68.4 & +5.5 \\
InternVL3.5-8B & 62.9 & 55.8 & 53.3 & 82.6 & 66.0 & 35.8 & 8.3 & 16.7 & 58.2 & 63.8 & $-$5.6 \\
Qwen2-VL-2B & 15.4 & 22.1 & 0.0 & 12.5 & 23.4 & 23.9 & 0.0 & 38.4 & 18.7 & 13.4 & +5.3 \\
Cosmos-Reason1-7B & 100.0 & 100.0 & 100.0 & 100.0 & 100.0 & 100.0 & 100.0 & 23.0 & 100.0 & 100.0 & 0.0 \\
MiniCPM-o-4.5 & 93.7 & 96.2 & 78.3 & 99.5 & 97.9 & 92.5 & 50.0 & 65.6 & 94.2 & 92.6 & +1.6 \\
Qwen3-VL-2B & 99.2 & 100.0 & 98.3 & 98.4 & 100.0 & 100.0 & 100.0 & 30.2 & 100.0 & 98.8 & +1.2 \\
BAGEL-7B-MoT & 98.9 & 98.1 & 100.0 & 99.5 & 100.0 & 100.0 & 83.3 & 28.7 & 99.3 & 98.8 & +0.5 \\
Janus-Pro-7B & 100.0 & 100.0 & 100.0 & 100.0 & 100.0 & 100.0 & 100.0 & 31.0 & 100.0 & 100.0 & 0.0 \\
Qwen3.5-2B & 95.4 & 88.5 & 100.0 & 97.8 & 97.9 & 94.0 & 91.7 & 21.7 & 95.1 & 95.4 & $-$0.3 \\
Qwen3.5-0.8B & 96.6 & 88.5 & 100.0 & 99.5 & 97.9 & 98.5 & 91.7 & 0.2 & 100.0 & 96.6 & +3.4 \\
\bottomrule
\end{tabular}
\end{center}
\end{table}

\begin{table}[h]
\caption{Detection and KGR (\%, prompted verification, strict scoring) by saliency of the edit over all families: C1 main figure (605 forgeries), C2 secondary figure (72), C3 background (830). $^{\dagger}$Also the judge.}
\label{tab:saliency}
\begin{center}
\footnotesize
\setlength{\tabcolsep}{4pt}
\begin{tabular}{@{}l rrr r rrr@{}}
\toprule
 & \multicolumn{4}{c}{Detection} & \multicolumn{3}{c}{KGR} \\
\cmidrule(lr){2-5}\cmidrule(lr){6-8}
Model & C1 & C2 & C3 & C1$-$C3 & C1 & C2 & C3 \\
\midrule
\multicolumn{8}{@{}l}{\emph{Proprietary}} \\
Gemini-3.8-Flash & 81.5 & 98.6 & 78.2 & +3.3 & 42.0 & 93.1 & 61.4 \\
GPT-5.6 & 56.4 & 97.2 & 61.2 & $-$4.8 & 22.6 & 81.9 & 41.7 \\
\addlinespace[2pt]
\multicolumn{8}{@{}l}{\emph{Open weights}} \\
GLM-5.3-Flash & 58.2 & 90.3 & 59.8 & $-$1.6 & 20.2 & 79.2 & 41.6 \\
Kimi-K3 & 62.5 & 93.1 & 58.4 & +4.1 & 19.8 & 81.9 & 40.4 \\
Kimi-K2.6 & 54.2 & 91.7 & 54.8 & $-$0.6 & 19.0 & 79.2 & 37.5 \\
Qwen3.5-122B-A10B & 54.9 & 94.4 & 56.5 & $-$1.6 & 21.0 & 79.2 & 35.1 \\
Qwen3-VL-235B-A22B & 65.5 & 88.9 & 60.2 & +5.3 & 14.4 & 72.2 & 32.2 \\
Qwen3.6-27B & 61.0 & 93.1 & 59.9 & +1.1 & 14.5 & 72.2 & 30.6 \\
Qwen3.5-27B & 48.1 & 87.5 & 53.7 & $-$5.6 & 14.0 & 70.8 & 28.9 \\
Qwen3-VL-235B-A22B (think) & 45.5 & 93.1 & 55.3 & $-$9.8 & 9.3 & 62.5 & 31.6 \\
Qwen3-VL-8B & 64.8 & 95.8 & 72.8 & $-$8.0 & 9.8 & 70.8 & 29.0 \\
Qwen3.8-27B$^{\dagger}$ & 61.5 & 91.7 & 67.0 & $-$5.5 & 8.6 & 76.4 & 27.6 \\
Qwen3.5-35B-A3B & 32.4 & 80.6 & 42.5 & $-$10.1 & 12.4 & 66.7 & 25.5 \\
Qwen3.6-35B-A3B & 36.2 & 87.5 & 40.8 & $-$4.6 & 10.9 & 73.6 & 25.2 \\
Qwen3-VL-32B (think) & 51.4 & 88.9 & 50.8 & +0.6 & 8.4 & 65.3 & 26.6 \\
Qwen3-VL-32B & 53.4 & 90.3 & 49.5 & +3.9 & 8.8 & 65.3 & 25.8 \\
Qwen3-Omni-30B-A3B (think) & 35.2 & 84.7 & 43.9 & $-$8.7 & 5.6 & 62.5 & 26.9 \\
Gemma-4-31B & 83.0 & 97.2 & 83.4 & $-$0.4 & 10.4 & 59.7 & 22.5 \\
Qwen3.5-9B & 19.2 & 68.1 & 30.2 & $-$11.0 & 4.5 & 54.2 & 19.8 \\
MiniMax-M3 & 25.6 & 63.9 & 26.3 & $-$0.7 & 4.6 & 48.6 & 16.5 \\
Qwen2.5-Omni-7B & 51.2 & 81.9 & 39.5 & +11.7 & 2.3 & 54.2 & 14.9 \\
Qwen3.5-4B & 14.5 & 61.1 & 22.9 & $-$8.4 & 4.1 & 45.8 & 14.3 \\
Qwen3-Omni-30B-A3B & 13.1 & 73.6 & 21.6 & $-$8.5 & 2.5 & 55.6 & 14.2 \\
Qwen2.5-VL-7B & 24.6 & 75.0 & 22.7 & +1.9 & 3.1 & 54.2 & 12.9 \\
InternVL3.5-30B-A3B & 77.0 & 93.1 & 67.8 & +9.2 & 2.6 & 41.7 & 13.5 \\
Qwen2.5-VL-3B & 69.3 & 88.9 & 55.2 & +14.1 & 3.1 & 34.7 & 13.4 \\
Gemma-3-12B & 35.2 & 63.9 & 32.8 & +2.4 & 3.8 & 36.1 & 11.6 \\
InternVL3.5-8B & 53.7 & 73.6 & 46.3 & +7.4 & 1.3 & 36.1 & 8.4 \\
Qwen2-VL-2B & 89.9 & 95.8 & 86.6 & +3.3 & 2.6 & 36.1 & 6.7 \\
Cosmos-Reason1-7B & 1.7 & 27.8 & 4.6 & $-$2.9 & 0.2 & 16.7 & 2.9 \\
MiniCPM-o-4.5 & 2.8 & 22.2 & 3.3 & $-$0.5 & 0.0 & 8.3 & 1.9 \\
Qwen3-VL-2B & 0.3 & 1.4 & 0.7 & $-$0.4 & 0.0 & 0.0 & 0.6 \\
BAGEL-7B-MoT & 0.3 & 2.8 & 0.6 & $-$0.3 & 0.0 & 0.0 & 0.0 \\
Janus-Pro-7B & 0.0 & 0.0 & 0.0 & 0.0 & 0.0 & 0.0 & 0.0 \\
Qwen3.5-2B & 9.3 & 48.6 & 12.0 & $-$2.7 & 0.0 & 0.0 & 0.0 \\
Qwen3.5-0.8B & 4.3 & 6.9 & 2.7 & +1.6 & 0.0 & 0.0 & 0.0 \\
\bottomrule
\end{tabular}
\end{center}
\end{table}

\begin{table}[h]
\caption{E4 rate next to KGR (E3+), and the people family without threshold-borderline items (\%, strict scoring). E3+ / E4 under the default (L0) and prompted (L1) probes over all forgeries; K1 Det / KGR under L1 on all people-family forgeries and after dropping the 41 items whose face-similarity score lies within $\pm0.05$ of the acceptance threshold. $^{\dagger}$Also the judge.}
\label{tab:efour}
\begin{center}
\footnotesize
\setlength{\tabcolsep}{4pt}
\begin{tabular}{@{}l rr rr rr rr@{}}
\toprule
 & \multicolumn{2}{c}{L0} & \multicolumn{2}{c}{L1} & \multicolumn{2}{c}{K1 Det (L1)} & \multicolumn{2}{c}{K1 KGR (L1)} \\
\cmidrule(lr){2-3}\cmidrule(lr){4-5}\cmidrule(lr){6-7}\cmidrule(lr){8-9}
Model & E3+ & E4 & E3+ & E4 & all & no border & all & no border \\
\midrule
\multicolumn{9}{@{}l}{\emph{Proprietary}} \\
Gemini-3.8-Flash & 46.6 & 28.4 & 55.1 & 37.2 & 81.1 & 80.0 & 33.7 & 33.7 \\
GPT-5.6 & 19.8 & 8.3 & 36.0 & 26.3 & 62.1 & 61.4 & 18.7 & 18.6 \\
\addlinespace[2pt]
\multicolumn{9}{@{}l}{\emph{Open weights}} \\
GLM-5.3-Flash & 29.7 & 22.5 & 34.8 & 29.6 & 60.0 & 59.1 & 11.1 & 10.8 \\
Kimi-K3 & 24.4 & 17.6 & 34.1 & 29.7 & 61.2 & 60.3 & 8.8 & 8.9 \\
Kimi-K2.6 & 22.2 & 15.1 & 32.0 & 25.6 & 52.9 & 52.5 & 11.3 & 11.4 \\
Qwen3.5-122B-A10B & 23.6 & 13.3 & 31.5 & 21.8 & 51.0 & 49.8 & 13.8 & 13.3 \\
Qwen3-VL-235B-A22B & 18.1 & 11.6 & 26.9 & 16.9 & 69.3 & 68.3 & 11.5 & 11.6 \\
Qwen3.6-27B & 19.8 & 10.7 & 26.1 & 15.7 & 58.7 & 57.2 & 7.8 & 7.6 \\
Qwen3.5-27B & 17.9 & 10.5 & 24.9 & 17.5 & 44.3 & 43.0 & 6.9 & 6.8 \\
Qwen3-VL-235B-A22B (think) & 15.0 & 9.2 & 24.1 & 16.7 & 48.7 & 47.3 & 6.0 & 6.1 \\
Qwen3-VL-8B & 16.1 & 7.7 & 23.3 & 11.7 & 75.1 & 74.9 & 7.2 & 7.0 \\
Qwen3.8-27B$^{\dagger}$ & 13.3 & 7.2 & 22.3 & 14.2 & 65.1 & 64.8 & 3.9 & 3.8 \\
Qwen3.5-35B-A3B & 17.2 & 9.9 & 22.2 & 14.2 & 29.8 & 28.7 & 7.4 & 7.0 \\
Qwen3.6-35B-A3B & 17.6 & 9.7 & 21.8 & 14.5 & 30.7 & 30.2 & 5.8 & 5.7 \\
Qwen3-VL-32B (think) & 14.3 & 8.9 & 21.2 & 14.1 & 50.3 & 49.4 & 4.1 & 3.8 \\
Qwen3-VL-32B & 16.9 & 9.2 & 20.8 & 12.5 & 53.6 & 53.0 & 5.1 & 4.8 \\
Qwen3-Omni-30B-A3B (think) & 13.0 & 5.8 & 20.0 & 11.7 & 40.4 & 39.5 & 5.1 & 5.1 \\
Gemma-4-31B & 11.7 & 5.2 & 19.4 & 10.7 & 83.8 & 83.1 & 5.5 & 5.5 \\
Qwen3.5-9B & 11.5 & 6.1 & 15.3 & 9.2 & 18.7 & 17.5 & 3.2 & 3.0 \\
MiniMax-M3 & 6.0 & 3.5 & 13.3 & 9.6 & 26.5 & 25.9 & 2.3 & 2.5 \\
Qwen2.5-Omni-7B & 8.5 & 3.0 & 11.7 & 4.5 & 50.6 & 49.8 & 1.1 & 1.0 \\
Qwen3.5-4B & 8.3 & 3.2 & 11.7 & 6.6 & 13.4 & 12.5 & 3.2 & 3.0 \\
Qwen3-Omni-30B-A3B & 5.9 & 2.3 & 11.5 & 5.4 & 15.9 & 15.4 & 3.0 & 3.2 \\
Qwen2.5-VL-7B & 6.7 & 1.7 & 10.9 & 3.4 & 26.6 & 25.5 & 2.8 & 2.7 \\
InternVL3.5-30B-A3B & 5.5 & 1.2 & 10.5 & 2.6 & 73.9 & 72.2 & 0.4 & 0.2 \\
Qwen2.5-VL-3B & 7.2 & 0.7 & 10.3 & 1.6 & 69.0 & 68.1 & 1.9 & 1.9 \\
Gemma-3-12B & 5.8 & 0.8 & 9.6 & 3.3 & 39.3 & 38.0 & 3.9 & 3.6 \\
InternVL3.5-8B & 3.7 & 0.5 & 6.9 & 2.3 & 58.2 & 57.0 & 0.5 & 0.6 \\
Qwen2-VL-2B & 0.0 & 0.0 & 6.5 & 0.2 & 90.1 & 89.4 & 1.1 & 0.8 \\
Cosmos-Reason1-7B & 3.1 & 0.9 & 2.5 & 0.7 & 1.9 & 1.7 & 0.4 & 0.4 \\
MiniCPM-o-4.5 & 1.3 & 0.3 & 1.5 & 0.7 & 3.2 & 2.9 & 0.0 & 0.0 \\
Qwen3-VL-2B & 0.5 & 0.1 & 0.3 & 0.1 & 0.4 & 0.2 & 0.0 & 0.0 \\
BAGEL-7B-MoT & 0.2 & 0.0 & 0.0 & 0.0 & 0.4 & 0.4 & 0.0 & 0.0 \\
Janus-Pro-7B & 0.1 & 0.0 & 0.0 & 0.0 & 0.0 & 0.0 & 0.0 & 0.0 \\
Qwen3.5-2B & 0.0 & 0.0 & 0.0 & 0.0 & 9.5 & 9.1 & 0.0 & 0.0 \\
Qwen3.5-0.8B & 0.0 & 0.0 & 0.0 & 0.0 & 4.4 & 4.2 & 0.0 & 0.0 \\
\bottomrule
\end{tabular}
\end{center}
\end{table}

\begin{table}[h]
\caption{Detection rate (\%) on forgeries under the default probe by knowledge family, over items with a parsed verdict. Family sizes (forged items): K0 148, K1 567, K2 294, K3 89, K4 108, K5 145, K6 156. Rows in the order of Table~\ref{tab:main}. KGR by family is in Table~\ref{tab:main-all}.}
\label{tab:family-detection}
\begin{center}
\footnotesize
\begin{tabular}{@{}l rrrrrrr@{}}
\toprule
Model & K0 & K1 & K2 & K3 & K4 & K5 & K6 \\
\midrule
Gemini-3.8-Flash & 11 & 65 & 92 & 74 & 62 & 70 & 69 \\
GPT-5.6 & 5 & 40 & 73 & 43 & 40 & 30 & 28 \\
GLM-5.3-Flash & 13 & 56 & 75 & 64 & 61 & 50 & 53 \\
Kimi-K3 & 4 & 48 & 67 & 45 & 67 & 37 & 29 \\
Kimi-K2.6 & 5 & 40 & 65 & 33 & 54 & 39 & 18 \\
Qwen3.5-122B-A10B & 22 & 49 & 69 & 52 & 45 & 48 & 39 \\
Qwen3-VL-235B-A22B & 5 & 31 & 70 & 45 & 41 & 20 & 10 \\
Qwen3.6-27B & 24 & 48 & 71 & 46 & 48 & 39 & 32 \\
Qwen3.5-27B & 21 & 41 & 70 & 43 & 57 & 39 & 32 \\
Qwen3-VL-235B-A22B (think) & 5 & 24 & 62 & 30 & 49 & 17 & 10 \\
Qwen3-VL-8B & 11 & 30 & 64 & 55 & 44 & 34 & 8 \\
Qwen3.8-27B & 23 & 40 & 60 & 34 & 47 & 35 & 22 \\
Qwen3.5-35B-A3B & 14 & 35 & 68 & 43 & 62 & 38 & 22 \\
Qwen3.6-35B-A3B & 12 & 38 & 68 & 42 & 48 & 40 & 25 \\
Qwen3-VL-32B (think) & 7 & 26 & 62 & 31 & 65 & 14 & 14 \\
Qwen3-VL-32B & 9 & 34 & 68 & 46 & 52 & 24 & 28 \\
Qwen3-Omni-30B-A3B (think) & 5 & 22 & 54 & 20 & 40 & 12 & 10 \\
Gemma-4-31B & 53 & 71 & 77 & 64 & 81 & 76 & 32 \\
Qwen3.5-9B & 14 & 28 & 58 & 22 & 37 & 19 & 11 \\
MiniMax-M3 & 1 & 12 & 23 & 3 & 23 & 7 & 4 \\
Qwen2.5-Omni-7B & 33 & 49 & 75 & 43 & 58 & 25 & 85 \\
Qwen3.5-4B & 9 & 26 & 51 & 22 & 18 & 21 & 6 \\
Qwen3-Omni-30B-A3B & 0 & 4 & 34 & 7 & 0 & 1 & 1 \\
Qwen2.5-VL-7B & 1 & 12 & 37 & 13 & 36 & 3 & 1 \\
InternVL3.5-30B-A3B & 17 & 34 & 55 & 40 & 47 & 22 & 21 \\
Qwen2.5-VL-3B & 60 & 76 & 82 & 67 & 71 & 50 & 83 \\
Gemma-3-12B & 18 & 26 & 49 & 19 & 24 & 11 & 48 \\
InternVL3.5-8B & 28 & 30 & 49 & 37 & 27 & 17 & 65 \\
Qwen2-VL-2B & 74 & 87 & 92 & 82 & 89 & 75 & 96 \\
Cosmos-Reason1-7B & 0 & 3 & 22 & 3 & 2 & 1 & 0 \\
MiniCPM-o-4.5 & 0 & 3 & 10 & 0 & 9 & 0 & 1 \\
Qwen3-VL-2B & 0 & 1 & 3 & 0 & 0 & 0 & 0 \\
BAGEL-7B-MoT & 5 & 13 & 19 & 6 & 28 & 4 & 15 \\
Janus-Pro-7B & 0 & 0 & 0 & 0 & 0 & 0 & 1 \\
Qwen3.5-2B & 1 & 4 & 8 & 1 & 4 & 2 & 0 \\
Qwen3.5-0.8B & 80 & 79 & 92 & 82 & 80 & 58 & 88 \\
\bottomrule
\end{tabular}
\end{center}
\end{table}

\begin{table}[h]
\caption{All four de-recognition conditions for every model with condition variants, using the same population, probe, and strict scoring as Table~\ref{tab:conditions}. We report detection, authentic accuracy, balanced accuracy, and KGR under Full (F), Full+Caption (FC), Crop (C), and Crop+Caption (CC). Each sweep covers 145 forgeries and the 98 authentic originals from the same scenes.}
\label{tab:conditions-all}
\begin{center}
\scriptsize
\setlength{\tabcolsep}{1.6pt}
\begin{tabular}{@{}l rrrr rrrr rrrr rrrr@{}}
\toprule
 & \multicolumn{4}{c}{Detection on forgeries (\%)} & \multicolumn{4}{c}{Authentic accuracy (\%)} & \multicolumn{4}{c}{Balanced accuracy (\%)} & \multicolumn{4}{c}{KGR (\%)} \\
\cmidrule(lr){2-5}\cmidrule(lr){6-9}\cmidrule(lr){10-13}\cmidrule(lr){14-17}
Model & F & FC & C & CC & F & FC & C & CC & F & FC & C & CC & F & FC & C & CC \\
\midrule
Gemini-3.8-Flash & 73.1 & 77.9 & 79.3 & 86.2 & 84.7 & 72.4 & 73.5 & 72.0 & 78.9 & 75.2 & 76.4 & 79.1 & 35.2 & 33.1 & 46.2 & 44.8 \\
GPT-5.6 & 53.8 & 57.2 & 56.6 & 69.7 & 84.7 & 85.7 & 78.1 & 83.9 & 69.2 & 71.5 & 67.3 & 76.8 & 14.5 & 13.1 & 13.1 & 16.6 \\
GLM-5.3-Flash & 52.4 & 47.6 & 62.8 & 59.3 & 82.7 & 89.8 & 74.9 & 89.7 & 67.5 & 68.7 & 68.8 & 74.5 & 7.6 & 4.8 & 14.5 & 10.3 \\
Kimi-K3 & 35.9 & 26.9 & 82.8 & 62.8 & 92.9 & 95.9 & 57.1 & 86.5 & 64.4 & 61.4 & 69.9 & 74.6 & 3.4 & 2.1 & 12.4 & 8.3 \\
Kimi-K2.6 & 28.3 & 29.7 & 67.6 & 63.4 & 89.8 & 87.8 & 75.0 & 89.2 & 59.0 & 58.7 & 71.3 & 76.3 & 4.1 & 2.8 & 20.0 & 15.2 \\
Qwen3.5-122B-A10B & 31.7 & 31.0 & 50.3 & 49.7 & 70.4 & 72.4 & 46.9 & 73.5 & 51.1 & 51.7 & 48.6 & 61.6 & 3.4 & 3.4 & 6.9 & 10.3 \\
Qwen3-VL-235B-A22B & 48.3 & 32.4 & 89.0 & 55.9 & 70.4 & 84.7 & 30.8 & 70.6 & 59.3 & 58.6 & 59.9 & 63.2 & 4.1 & 3.4 & 11.0 & 10.3 \\
Qwen3.6-27B & 40.7 & 31.7 & 49.7 & 51.7 & 56.1 & 65.3 & 27.4 & 65.5 & 48.4 & 48.5 & 38.5 & 58.6 & 2.8 & 1.4 & 6.9 & 5.5 \\
Qwen3.5-27B & 32.4 & 30.3 & 42.8 & 41.4 & 52.0 & 58.2 & 37.3 & 68.1 & 42.2 & 44.3 & 40.1 & 54.8 & 1.4 & 2.1 & 4.8 & 4.8 \\
Qwen3-VL-235B-A22B (think) & 29.0 & 22.1 & 69.7 & 54.5 & 80.6 & 86.7 & 48.7 & 79.5 & 54.8 & 54.4 & 59.2 & 67.0 & 2.1 & 0.7 & 4.8 & 6.2 \\
Qwen3-VL-8B & 65.5 & 21.4 & 89.0 & 51.7 & 49.0 & 82.7 & 23.7 & 67.4 & 57.2 & 52.0 & 56.3 & 59.6 & 2.1 & 2.8 & 10.3 & 5.5 \\
Qwen3.8-27B$^{\dagger}$ & 54.5 & 51.7 & 67.6 & 71.7 & 66.3 & 58.2 & 56.8 & 67.8 & 60.4 & 54.9 & 62.2 & 69.8 & 1.4 & 0.7 & 4.8 & 3.4 \\
Qwen3.5-35B-A3B & 23.4 & 18.6 & 24.8 & 39.3 & 57.1 & 66.3 & 30.1 & 64.6 & 40.3 & 42.5 & 27.5 & 52.0 & 3.4 & 1.4 & 1.4 & 9.7 \\
Qwen3.6-35B-A3B & 22.1 & 25.5 & 36.6 & 39.3 & 54.1 & 70.4 & 37.5 & 65.7 & 38.1 & 48.0 & 37.0 & 52.5 & 1.4 & 2.8 & 2.8 & 9.0 \\
Qwen3-VL-32B (think) & 29.7 & 27.6 & 74.5 & 62.1 & 78.6 & 79.6 & 45.7 & 64.1 & 54.1 & 53.6 & 60.1 & 63.1 & 1.4 & 0.0 & 2.1 & 4.8 \\
Qwen3-VL-32B & 29.7 & 24.1 & 75.2 & 57.9 & 79.6 & 83.7 & 55.4 & 76.5 & 54.6 & 53.9 & 65.3 & 67.2 & 1.4 & 0.7 & 9.0 & 4.8 \\
Qwen3-Omni-30B-A3B (think) & 22.1 & 25.5 & 53.8 & 61.4 & 91.8 & 81.6 & 60.2 & 65.1 & 57.0 & 53.6 & 57.0 & 63.3 & 2.8 & 1.4 & 3.4 & 3.4 \\
Gemma-4-31B & 77.2 & 53.1 & 97.2 & 86.9 & 30.6 & 53.1 & 10.6 & 41.1 & 53.9 & 53.1 & 53.9 & 64.0 & 1.4 & 1.4 & 7.6 & 8.3 \\
Qwen3.5-9B & 12.5 & 13.8 & 12.4 & 29.0 & 41.8 & 66.3 & 24.6 & 62.8 & 27.2 & 40.1 & 18.5 & 45.9 & 2.1 & 2.1 & 0.7 & 4.8 \\
MiniMax-M3 & 14.5 & 11.7 & 21.4 & 29.7 & 90.8 & 93.9 & 79.3 & 92.4 & 52.6 & 52.8 & 50.3 & 61.0 & 1.4 & 0.7 & 2.8 & 2.8 \\
Qwen2.5-Omni-7B & 21.4 & 15.9 & 75.2 & 49.7 & 82.7 & 86.7 & 29.4 & 61.8 & 52.0 & 51.3 & 52.3 & 55.7 & 0.7 & 0.7 & 1.4 & 2.1 \\
Qwen3.5-4B & 4.8 & 13.1 & 11.0 & 32.4 & 33.7 & 52.0 & 20.5 & 49.2 & 19.3 & 32.6 & 15.8 & 40.8 & 0.7 & 0.7 & 0.7 & 4.8 \\
Qwen3-Omni-30B-A3B & 9.7 & 15.2 & 6.9 & 20.7 & 98.0 & 94.9 & 97.3 & 95.4 & 53.8 & 55.0 & 52.1 & 58.0 & 2.1 & 2.1 & 0.7 & 4.1 \\
Qwen2.5-VL-7B & 6.2 & 11.0 & 22.1 & 24.1 & 94.9 & 90.8 & 77.5 & 86.4 & 50.6 & 50.9 & 49.8 & 55.3 & 0.7 & 1.4 & 0.7 & 2.8 \\
InternVL3.5-30B-A3B & 70.3 & 20.0 & 94.5 & 62.8 & 31.6 & 81.6 & 6.9 & 53.3 & 51.0 & 50.8 & 50.7 & 58.0 & 0.0 & 0.0 & 0.0 & 0.0 \\
Qwen2.5-VL-3B & 51.7 & 53.1 & 100.0 & 100.0 & 51.0 & 51.0 & 1.6 & 3.0 & 51.4 & 52.1 & 50.8 & 51.5 & 1.4 & 0.0 & 0.7 & 1.4 \\
Gemma-3-12B & 22.1 & 10.3 & 69.7 & 35.9 & 80.6 & 91.8 & 52.0 & 92.7 & 51.3 & 51.1 & 60.8 & 64.3 & 2.1 & 0.0 & 11.7 & 9.7 \\
InternVL3.5-8B & 47.6 & 26.2 & 53.1 & 33.8 & 59.2 & 78.6 & 57.9 & 79.6 & 53.4 & 52.4 & 55.5 & 56.7 & 0.0 & 0.0 & 0.0 & 0.0 \\
Qwen2-VL-2B & 83.4 & 71.7 & 100.0 & 100.0 & 17.3 & 34.7 & 0.7 & 1.6 & 50.4 & 53.2 & 50.4 & 50.8 & 0.0 & 0.0 & 0.0 & 0.0 \\
Cosmos-Reason1-7B & 0.7 & 0.7 & 0.0 & 1.4 & 100.0 & 100.0 & 98.4 & 99.5 & 50.3 & 50.3 & 49.2 & 50.4 & 0.7 & 0.0 & 0.0 & 0.7 \\
MiniCPM-o-4.5 & 1.4 & 3.4 & 1.4 & 3.4 & 88.8 & 89.8 & 98.1 & 87.3 & 45.1 & 46.6 & 49.7 & 45.4 & 0.0 & 0.0 & 0.0 & 0.7 \\
Qwen3-VL-2B & 0.0 & 0.0 & 0.0 & 0.0 & 100.0 & 98.0 & 99.5 & 99.1 & 50.0 & 49.0 & 49.7 & 49.6 & 0.0 & 0.0 & 0.0 & 0.0 \\
BAGEL-7B-MoT & 0.0 & 0.0 & 0.7 & 0.0 & 100.0 & 100.0 & 98.9 & 100.0 & 50.0 & 50.0 & 49.8 & 50.0 & 0.0 & 0.0 & 0.0 & 0.0 \\
Janus-Pro-7B & 0.0 & 0.0 & 0.0 & 0.0 & 100.0 & 100.0 & 100.0 & 100.0 & 50.0 & 50.0 & 50.0 & 50.0 & 0.0 & 0.0 & 0.0 & 0.0 \\
Qwen3.5-2B & 4.8 & 12.4 & 19.3 & 20.7 & 96.9 & 91.8 & 84.2 & 87.1 & 50.9 & 52.1 & 51.8 & 53.9 & 0.0 & 0.0 & 0.0 & 0.0 \\
Qwen3.5-0.8B & 0.7 & 31.7 & 0.0 & 70.3 & 96.9 & 64.3 & 99.6 & 38.8 & 48.8 & 48.0 & 49.8 & 54.6 & 0.0 & 0.0 & 0.0 & 0.0 \\
\bottomrule
\end{tabular}
\end{center}
\end{table}

\begin{table}[h]
\caption{De-recognition conditions of Table~\ref{tab:conditions} under parsed-only scoring (items without a parsed verdict excluded from the denominator). Conditions abbreviated as in Table~\ref{tab:conditions-all}. KGR is unchanged by the convention and is omitted.}
\label{tab:conditions-parsed}
\begin{center}
\footnotesize
\setlength{\tabcolsep}{2.6pt}
\begin{tabular}{@{}l rrrr rrrr rrrr@{}}
\toprule
 & \multicolumn{4}{c}{Detection on forgeries (\%)} & \multicolumn{4}{c}{Authentic accuracy (\%)} & \multicolumn{4}{c}{Balanced accuracy (\%)} \\
\cmidrule(lr){2-5}\cmidrule(lr){6-9}\cmidrule(lr){10-13}
Model & F & FC & C & CC & F & FC & C & CC & F & FC & C & CC \\
\midrule
Gemini-3.8-Flash & 73.1 & 77.9 & 79.3 & 86.2 & 84.7 & 72.4 & 73.5 & 72.0 & 78.9 & 75.2 & 76.4 & 79.1 \\
GPT-5.6 & 53.8 & 57.2 & 56.6 & 69.7 & 84.7 & 85.7 & 78.1 & 83.9 & 69.2 & 71.5 & 67.3 & 76.8 \\
GLM-5.3-Flash & 59.4 & 50.0 & 72.2 & 61.0 & 92.0 & 96.7 & 83.1 & 93.0 & 75.7 & 73.4 & 77.7 & 77.0 \\
Kimi-K3 & 35.9 & 27.3 & 82.8 & 63.6 & 92.9 & 98.9 & 57.3 & 86.8 & 64.4 & 63.1 & 70.0 & 75.2 \\
Kimi-K2.6 & 28.3 & 29.9 & 68.1 & 63.9 & 89.8 & 87.8 & 75.7 & 89.4 & 59.0 & 58.8 & 71.9 & 76.6 \\
Qwen3.5-122B-A10B & 35.9 & 38.5 & 61.9 & 58.1 & 79.3 & 81.6 & 56.9 & 80.6 & 57.6 & 60.0 & 59.4 & 69.3 \\
Qwen3-VL-235B-A22B & 48.3 & 32.4 & 89.0 & 55.9 & 70.4 & 84.7 & 30.8 & 70.6 & 59.3 & 58.6 & 59.9 & 63.2 \\
Qwen3.6-27B & 45.4 & 38.0 & 72.0 & 62.5 & 59.1 & 76.2 & 39.9 & 76.1 & 52.3 & 57.1 & 56.0 & 69.3 \\
Qwen3.5-27B & 50.0 & 41.1 & 67.4 & 52.6 & 67.1 & 73.1 & 56.7 & 77.8 & 58.6 & 57.1 & 62.1 & 65.2 \\
Qwen3-VL-235B-A22B (think) & 29.0 & 22.1 & 69.7 & 54.5 & 80.6 & 86.7 & 48.7 & 79.5 & 54.8 & 54.4 & 59.2 & 67.0 \\
Qwen3-VL-8B & 66.0 & 21.7 & 89.0 & 52.1 & 49.0 & 84.4 & 23.7 & 67.7 & 57.5 & 53.0 & 56.3 & 59.9 \\
Qwen3.8-27B$^{\dagger}$ & 54.5 & 52.1 & 67.6 & 72.2 & 66.3 & 58.2 & 56.9 & 67.8 & 60.4 & 55.1 & 62.3 & 70.0 \\
Qwen3.5-35B-A3B & 41.0 & 26.5 & 56.2 & 57.0 & 76.7 & 84.4 & 50.6 & 77.8 & 58.8 & 55.4 & 53.4 & 67.4 \\
Qwen3.6-35B-A3B & 37.2 & 33.9 & 59.6 & 53.3 & 77.9 & 86.2 & 57.9 & 78.8 & 57.6 & 60.1 & 58.7 & 66.0 \\
Qwen3-VL-32B (think) & 29.7 & 28.6 & 77.1 & 65.2 & 78.6 & 81.2 & 46.8 & 69.1 & 54.1 & 54.9 & 62.0 & 67.2 \\
Qwen3-VL-32B & 29.7 & 24.1 & 75.2 & 57.9 & 79.6 & 83.7 & 55.4 & 76.5 & 54.6 & 53.9 & 65.3 & 67.2 \\
Qwen3-Omni-30B-A3B (think) & 22.4 & 26.2 & 54.5 & 62.2 & 91.8 & 82.5 & 60.8 & 66.3 & 57.1 & 54.4 & 57.7 & 64.3 \\
Gemma-4-31B & 77.2 & 53.1 & 97.2 & 86.9 & 30.6 & 53.1 & 10.7 & 41.1 & 53.9 & 53.1 & 54.0 & 64.0 \\
Qwen3.5-9B & 32.1 & 20.6 & 41.9 & 47.7 & 77.4 & 87.8 & 66.8 & 82.4 & 54.8 & 54.2 & 54.3 & 65.0 \\
MiniMax-M3 & 14.9 & 12.1 & 23.5 & 30.5 & 94.7 & 93.9 & 87.2 & 92.9 & 54.8 & 53.0 & 55.3 & 61.7 \\
Gemma-3-12B & 22.1 & 10.3 & 69.7 & 35.9 & 80.6 & 91.8 & 52.0 & 92.7 & 51.3 & 51.1 & 60.8 & 64.3 \\
\bottomrule
\end{tabular}
\end{center}
\end{table}

\begin{table}[h]
\caption{Detection, KGR, and authentic accuracy under Full, Mirror, and Crop for the 902 peripheral forgeries in Table~\ref{tab:mediation} and all 474 authentic images. $b/c$ counts forgeries detected only under Crop or only under Mirror, and $p$ is the exact McNemar test between them. Because this population includes visible-text forgeries (K4), for which mirroring itself changes the text, the paired recognition analysis in Table~\ref{tab:conditions} is restricted to the people family.}
\label{tab:mediation-detection}
\begin{center}
\scriptsize
\setlength{\tabcolsep}{3pt}
\begin{tabular}{@{}l rrr r r rrr rrr@{}}
\toprule
 & \multicolumn{3}{c}{Detection} & & & \multicolumn{3}{c}{KGR} & \multicolumn{3}{c}{Authentic accuracy} \\
\cmidrule(lr){2-4}\cmidrule(lr){7-9}\cmidrule(lr){10-12}
Model & Full & Mirror & Crop & $b$/$c$ & $p$ & Full & Mirror & Crop & Full & Mirror & Crop \\
\midrule
Qwen3.5-122B-A10B & 59.5 & 57.6 & 66.1 & 193/117 & $<$0.0001 & 38.6 & 29.5 & 40.5 & 77.4 & 60.5 & 55.7 \\
Kimi-K3 & 61.2 & 57.8 & 81.4 & 275/62 & $<$0.0001 & 43.7 & 36.6 & 51.9 & 92.2 & 84.2 & 74.3 \\
GPT-5.6 & 64.1 & 62.6 & 66.3 & 153/120 & 0.0526 & 44.9 & 40.5 & 45.0 & 91.6 & 82.1 & 82.4 \\
Gemini-3.8-Flash & 79.8 & 80.9 & 80.7 & 91/93 & 0.9413 & 64.0 & 55.5 & 65.5 & 88.2 & 68.1 & 80.2 \\
Qwen3.5-35B-A3B & 45.6 & 47.0 & 50.2 & 180/151 & 0.1237 & 28.8 & 24.3 & 30.7 & 58.4 & 35.0 & 28.5 \\
Qwen3.5-27B & 56.4 & 55.5 & 61.3 & 206/154 & 0.0071 & 32.3 & 25.5 & 33.6 & 56.3 & 35.2 & 36.5 \\
Qwen3-VL-235B-A22B (think) & 58.3 & 60.2 & 77.3 & 225/71 & $<$0.0001 & 34.0 & 26.8 & 39.0 & 78.9 & 65.2 & 56.3 \\
GLM-5.3-Flash & 62.2 & 66.4 & 65.5 & 117/125 & 0.6528 & 44.6 & 40.2 & 46.0 & 85.9 & 58.6 & 72.8 \\
Qwen3.5-9B & 33.3 & 29.8 & 36.0 & 164/108 & 0.0008 & 22.6 & 16.2 & 22.2 & 53.2 & 38.8 & 18.6 \\
Qwen3.6-27B & 62.5 & 56.1 & 61.4 & 184/136 & 0.0085 & 33.9 & 26.6 & 33.1 & 65.6 & 41.6 & 38.1 \\
Qwen3-VL-32B (think) & 53.9 & 58.1 & 75.1 & 247/94 & $<$0.0001 & 29.7 & 20.2 & 30.8 & 79.1 & 55.7 & 51.4 \\
Qwen3-Omni-30B-A3B (think) & 47.1 & 47.5 & 67.0 & 253/77 & $<$0.0001 & 29.7 & 25.2 & 35.7 & 81.2 & 71.3 & 55.7 \\
Gemma-3-12B & 35.3 & 44.2 & 74.6 & 340/66 & $<$0.0001 & 13.5 & 6.4 & 38.2 & 72.4 & 56.5 & 46.1 \\
Kimi-K2.6 & 57.8 & 55.1 & 72.1 & 233/80 & $<$0.0001 & 40.8 & 31.3 & 49.2 & 90.1 & 75.3 & 79.3 \\
MiniMax-M3 & 29.3 & 30.8 & 35.6 & 180/137 & 0.0182 & 19.1 & 12.5 & 22.0 & 91.1 & 80.6 & 83.3 \\
Qwen3.5-4B & 25.9 & 22.4 & 28.3 & 149/96 & 0.0009 & 16.9 & 13.3 & 16.6 & 46.2 & 32.5 & 21.1 \\
Qwen3-VL-32B & 52.8 & 57.3 & 74.6 & 218/62 & $<$0.0001 & 28.9 & 22.4 & 35.6 & 80.4 & 66.7 & 57.3 \\
MiniCPM-o-4.5 & 4.8 & 3.9 & 6.1 & 47/27 & 0.0265 & 2.4 & 1.9 & 4.0 & 93.7 & 93.2 & 98.1 \\
Qwen3-Omni-30B-A3B & 25.7 & 22.1 & 24.4 & 94/73 & 0.1214 & 17.5 & 14.0 & 15.2 & 97.9 & 99.6 & 97.5 \\
Qwen3.6-35B-A3B & 44.6 & 47.5 & 49.0 & 165/151 & 0.4646 & 29.0 & 24.2 & 28.8 & 61.8 & 44.5 & 39.3 \\
Qwen3.8-27B$^{\dagger}$ & 69.0 & 70.4 & 68.5 & 141/158 & 0.3548 & 31.5 & 24.8 & 32.2 & 62.0 & 43.0 & 56.7 \\
Gemma-4-31B & 84.5 & 90.0 & 95.1 & 81/35 & $<$0.0001 & 25.5 & 15.9 & 43.8 & 34.0 & 18.4 & 20.4 \\
Qwen3-VL-8B & 74.6 & 75.8 & 84.4 & 155/78 & $<$0.0001 & 32.4 & 28.0 & 39.2 & 50.4 & 42.6 & 31.3 \\
Qwen3-VL-235B-A22B & 62.5 & 66.5 & 88.8 & 239/38 & $<$0.0001 & 35.4 & 31.9 & 45.6 & 70.0 & 57.0 & 31.3 \\
Qwen2.5-VL-7B & 26.8 & 27.1 & 44.1 & 213/59 & $<$0.0001 & 16.2 & 10.8 & 22.1 & 89.9 & 83.8 & 74.3 \\
Qwen2-VL-2B & 87.4 & 89.5 & 99.6 & 91/0 & $<$0.0001 & 9.1 & 6.7 & 4.7 & 15.4 & 10.1 & 2.2 \\
BAGEL-7B-MoT & 0.8 & 0.4 & 2.2 & 18/2 & 0.0004 & 0.0 & 0.0 & 0.4 & 98.9 & 99.8 & 99.4 \\
Qwen2.5-Omni-7B & 42.9 & 44.5 & 79.7 & 347/29 & $<$0.0001 & 18.1 & 11.4 & 26.9 & 69.8 & 59.9 & 31.9 \\
Janus-Pro-7B & 0.0 & 0.0 & 0.0 & 0/0 & --- & 0.0 & 0.0 & 0.0 & 100.0 & 100.0 & 100.0 \\
InternVL3.5-30B-A3B & 69.8 & 81.8 & 88.7 & 124/62 & $<$0.0001 & 15.7 & 8.9 & 23.4 & 43.7 & 22.6 & 21.1 \\
Cosmos-Reason1-7B & 6.4 & 5.1 & 9.6 & 63/22 & $<$0.0001 & 4.0 & 2.3 & 6.4 & 100.0 & 99.8 & 100.0 \\
Qwen3-VL-2B & 0.8 & 0.8 & 0.6 & 4/6 & 0.7539 & 0.6 & 0.7 & 0.6 & 99.2 & 98.1 & 97.5 \\
Qwen3.5-2B & 15.0 & 16.9 & 28.0 & 164/63 & $<$0.0001 & 0.0 & 0.0 & 0.0 & 95.4 & 90.5 & 85.4 \\
InternVL3.5-8B & 48.4 & 55.5 & 59.6 & 191/154 & 0.0524 & 10.6 & 5.8 & 17.2 & 62.9 & 57.6 & 61.0 \\
Qwen2.5-VL-3B & 57.9 & 62.7 & 95.0 & 303/12 & $<$0.0001 & 15.1 & 11.0 & 24.6 & 50.4 & 41.6 & 11.5 \\
Qwen3.5-0.8B & 3.0 & 6.9 & 0.9 & 3/57 & $<$0.0001 & 0.0 & 0.0 & 0.0 & 96.6 & 91.8 & 98.8 \\
\bottomrule
\end{tabular}
\end{center}
\end{table}

\section{Footprint covariate analysis}
\label{app:footprint}

The absent-celebrity swap and erase conditions differ in how much of the frame they regenerate, which could itself affect detection. The median fraction of pixels changed outside the annotated region (the edit \emph{footprint}) is 4.8\% for swaps, 0.6\% for erasures, and 0.3\% for K0 edits. To test whether this difference explains the swap--erase gap, we fit a logistic regression per model on the 132 verified erasures, 78 same-seed swaps, and 41 K0 controls, with seed-clustered standard errors.

The base model predicts detection from condition (K0 as reference) and domain fixed effects. We then add log footprint and standardized scores from NPR and UniversalFakeDetect. Footprint is measured under two conventions: the recorded edit box and the largest connected component of the observed pixel difference. Table~\ref{tab:footprint} reports the seventeen models for which the regression is non-degenerate, defined as naming the erased person in at least a quarter of default-probe answers and detecting at least a quarter of swaps. The swap--erase contrast remains positive for every model and significant under both footprint definitions for sixteen of the seventeen. Adding footprint and detector scores changes the swap coefficient by $-$26.8\% to +11.1\%, but does not remove the identity effect for any model.

\begin{table}[h]
\caption{Per-model logistic regressions predicting detection from edit condition, with domain fixed effects, log footprint, and both pixel-detector scores; standard errors are clustered by seed. We report the swap-versus-erase log-odds contrast under both footprint definitions, and the swap coefficient relative to K0 before and after adding the covariates. ``Shrink'' is the change in that coefficient from the base to the full specification; negative values indicate an increase. Rows follow Table~\ref{tab:main}.}
\label{tab:footprint}
\begin{center}
\footnotesize
\setlength{\tabcolsep}{2.4pt}
\begin{tabular}{@{}l r rr rr rrr@{}}
\toprule
 & & \multicolumn{2}{c}{Contrast, recorded} & \multicolumn{2}{c}{Contrast, measured} & \multicolumn{3}{c}{Swap coef.\ vs.\ K0} \\
\cmidrule(lr){3-4}\cmidrule(lr){5-6}\cmidrule(lr){7-9}
Model & $n$ & log-odds & $p$ & log-odds & $p$ & base & full & shrink \\
\midrule
Gemini-3.8-Flash & 248 & 0.688 & 0.1188 & 0.946 & 0.0450 & 4.580 & 4.623 & $-$0.9\% \\
GLM-5.3-Flash & 239 & 1.449 & 0.0012 & 1.605 & 0.0020 & 3.072 & 3.056 & 0.5\% \\
Kimi-K3 & 248 & 1.430 & 0.0019 & 1.552 & 0.0021 & 4.556 & 4.165 & 8.6\% \\
Kimi-K2.6 & 248 & 1.596 & $<$0.0001 & 1.644 & 0.0001 & 3.581 & 3.260 & 9.0\% \\
Qwen3.5-122B-A10B & 242 & 1.520 & 0.0004 & 1.548 & 0.0005 & 2.046 & 2.132 & $-$4.2\% \\
Qwen3-VL-235B-A22B & 248 & 1.879 & $<$0.0001 & 1.508 & 0.0001 & 2.942 & 2.733 & 7.1\% \\
Qwen3.6-27B & 238 & 1.317 & 0.0092 & 1.608 & 0.0015 & 2.541 & 2.644 & $-$4.0\% \\
Qwen3.5-27B & 189 & 1.801 & 0.0006 & 1.745 & 0.0004 & 2.005 & 2.414 & $-$20.4\% \\
Qwen3-VL-235B-A22B (think) & 248 & 2.497 & $<$0.0001 & 2.725 & $<$0.0001 & 3.748 & 3.879 & $-$3.5\% \\
Qwen3-VL-8B & 248 & 1.276 & 0.0232 & 1.376 & 0.0051 & 2.696 & 3.039 & $-$12.7\% \\
Qwen3.8-27B$^{\dagger}$ & 248 & 0.815 & 0.0436 & 1.180 & 0.0195 & 2.029 & 2.033 & $-$0.2\% \\
Qwen3.5-35B-A3B & 178 & 1.483 & 0.0064 & 2.056 & 0.0042 & 2.620 & 3.323 & $-$26.8\% \\
Qwen3.6-35B-A3B & 192 & 2.033 & 0.0002 & 1.992 & 0.0018 & 2.782 & 3.448 & $-$24.0\% \\
Qwen3-VL-32B (think) & 248 & 1.414 & 0.0014 & 1.371 & 0.0039 & 3.559 & 3.409 & 4.2\% \\
Qwen3-VL-32B & 248 & 1.676 & 0.0001 & 1.835 & 0.0001 & 3.219 & 2.863 & 11.1\% \\
Qwen3-Omni-30B-A3B (think) & 247 & 2.418 & $<$0.0001 & 2.497 & 0.0001 & 2.485 & 2.689 & $-$8.2\% \\
Gemma-3-12B & 248 & 1.247 & 0.0147 & 1.366 & 0.0049 & 1.294 & 1.368 & $-$5.7\% \\
\bottomrule
\end{tabular}
\end{center}
\end{table}

\section{Preregistered decision rule for the absent-celebrity result}
\label{app:prereg}

Before evaluating the absent-celebrity items, we preregistered a decision rule. We classify a model as showing a \emph{collapse of verification} when erase detection is below 0.6 of its own same-seed swap detection and fewer than 15\% of erase verdicts reach E4. We classify it as \emph{no collapse} when the ratio is at least 0.8 and the E4 share is at least 30\%; all other cases are intermediate. The same rule is applied unchanged to every model with a swap control (Table~\ref{tab:prereg}). Among the thirty models that detect more than one swap in ten, twenty-one meet both collapse criteria and nine are intermediate.

\begin{table}[h]
\caption{Application of the preregistered rule (default probe, strict scoring). Ratio: erase detection over swap detection. E4: share of the erase items whose verdict states the contradiction. Models detecting fewer than one swap in ten are listed without a ratio. Rows in the order of Table~\ref{tab:main}.}
\label{tab:prereg}
\begin{center}
\footnotesize
\begin{tabular}{@{}l rrr r l@{}}
\toprule
Model & Erase (\%) & Swap (\%) & Ratio & E4 (\%) & Outcome \\
\midrule
Gemini-3.8-Flash & 53.0 & 71.8 & 0.739 & 7.6 & intermediate \\
GPT-5.6 & 17.4 & 52.6 & 0.331 & 0.0 & collapse \\
GLM-5.3-Flash & 32.6 & 65.4 & 0.498 & 2.3 & collapse \\
Kimi-K3 & 30.3 & 70.5 & 0.430 & 2.3 & collapse \\
Kimi-K2.6 & 21.2 & 59.0 & 0.360 & 1.5 & collapse \\
Qwen3.5-122B-A10B & 29.5 & 61.5 & 0.480 & 0.0 & collapse \\
Qwen3-VL-235B-A22B & 10.6 & 43.6 & 0.243 & 1.5 & collapse \\
Qwen3.6-27B & 28.0 & 52.6 & 0.533 & 0.0 & collapse \\
Qwen3.5-27B & 15.2 & 37.2 & 0.408 & 0.0 & collapse \\
Qwen3-VL-235B-A22B (think) & 8.3 & 41.0 & 0.203 & 0.0 & collapse \\
Qwen3-VL-8B & 19.7 & 33.3 & 0.591 & 0.0 & collapse \\
Qwen3.8-27B$^{\dagger}$ & 25.8 & 42.3 & 0.609 & 0.0 & intermediate \\
Qwen3.5-35B-A3B & 18.2 & 25.6 & 0.709 & 0.0 & intermediate \\
Qwen3.6-35B-A3B & 15.9 & 38.5 & 0.414 & 0.8 & collapse \\
Qwen3-VL-32B (think) & 12.9 & 33.3 & 0.386 & 0.0 & collapse \\
Qwen3-VL-32B & 12.1 & 41.0 & 0.295 & 0.0 & collapse \\
Qwen3-Omni-30B-A3B (think) & 6.1 & 32.1 & 0.189 & 0.0 & collapse \\
Gemma-4-31B & 68.9 & 73.1 & 0.943 & 0.8 & intermediate \\
Qwen3.5-9B & 9.1 & 16.7 & 0.545 & 0.0 & collapse \\
MiniMax-M3 & 6.1 & 11.5 & 0.525 & 0.0 & collapse \\
Qwen2.5-Omni-7B & 34.1 & 65.4 & 0.521 & 0.0 & collapse \\
Qwen3.5-4B & 3.8 & 20.5 & 0.185 & 0.0 & collapse \\
Qwen3-Omni-30B-A3B & 0.0 & 5.1 & --- & 0.0 & n/a (swap $\leq$ 10) \\
Qwen2.5-VL-7B & 3.0 & 11.5 & 0.263 & 0.0 & collapse \\
InternVL3.5-30B-A3B & 21.2 & 37.2 & 0.571 & 0.0 & collapse \\
Qwen2.5-VL-3B & 65.9 & 88.5 & 0.745 & 0.0 & intermediate \\
Gemma-3-12B & 12.1 & 29.5 & 0.411 & 0.0 & collapse \\
InternVL3.5-8B & 20.5 & 24.4 & 0.840 & 0.0 & intermediate \\
Qwen2-VL-2B & 80.3 & 92.3 & 0.870 & 0.0 & intermediate \\
Cosmos-Reason1-7B & 0.0 & 3.8 & --- & 0.0 & n/a (swap $\leq$ 10) \\
MiniCPM-o-4.5 & 1.5 & 0.0 & --- & 0.0 & n/a (swap $\leq$ 10) \\
Qwen3-VL-2B & 0.0 & 1.3 & --- & 0.0 & n/a (swap $\leq$ 10) \\
BAGEL-7B-MoT & 10.6 & 10.3 & 1.034 & 0.0 & intermediate \\
Janus-Pro-7B & 0.0 & 0.0 & --- & 0.0 & n/a (swap $\leq$ 10) \\
Qwen3.5-2B & 1.5 & 2.6 & --- & 0.0 & n/a (swap $\leq$ 10) \\
Qwen3.5-0.8B & 62.9 & 73.1 & 0.860 & 0.0 & intermediate \\
\bottomrule
\end{tabular}
\end{center}
\end{table}

\section{Face probe}
\label{app:face}

The face probe presents a single face at native resolution on a neutral canvas and asks ``Is this [name]?'' once for the original face and once for the replacement. An item passes when the model answers yes to the original and no to the replacement.

The probe was run on 159 items that had been verified at the time. Before evaluation, we exclude 3 items with incorrect identity labels and 28 whose faces cannot be cropped at usable size without strong worn cues, leaving 128 pairs. We then exclude 9 pairs in which the original face is not visible because of a lowered head, glove, hand, or deep shadow, leaving 119 evaluated pairs. The later release filter reduces the verified absent-celebrity set to 132, but the face-probe results are reported on the 119 pairs as run.

Kimi-K3 passes 55 pairs and Qwen3.6-27B passes 47. On these pass sets, the full-image verdicts form the denominators of the last two columns of Table~\ref{tab:identity}. Gemini-3.8-Flash passes 81 of 119 pairs (2.5\% false yes to a wrong name) and still calls the full forgery authentic 29.6\% of the time while naming the erased person in 49.4\% of its answers. Qwen3.8-27B passes 50 pairs (12.5\% false yes), with corresponding rates 68.0\% and 12.0\%. GPT-5.6 declines to identify real people from isolated faces, answering no to 112 of 119 originals, so its 4 passes should be treated as a refusal pattern rather than a perception result.

Because yes/no identity questions may encourage agreement, we also test 40 recognizable original faces paired with a wrong same-gender name. The false-yes rate is 10.0\% for Kimi-K3 and 0.0\% for Qwen3.6-27B, compared with 87.5\% and 82.5\% yes for the correct name; baseline-corrected pass rates are 41.6\% and 39.5\%. Finally, excluding 18 pairs with strong worn cues visible in both crops leaves 101 items and changes the rate at which passed full images are called authentic from 63.6\% to 63.0\% for Kimi-K3 and from 69.8\% to 63.6\% for Qwen3.6-27B.

\section{Additional analysis}
\label{app:analysis}

\paragraph{Naming the erased person is associated with missing the forgery.}
For Kimi-K3, items on which the model names the erased person are detected 22.0\% of the time, compared with 43.9\% when it says the person cannot be identified. Across all thirty-six models, eleven show lower detection when they name the erased person than when they do not (Table~\ref{tab:identity}). The identity probe also links this behavior to scene recognition. On the 116 peripheral people-family forgeries with named principal figures, all thirty-six models name those figures less often on the crop than on the full image (Kimi-K3 84.5$\rightarrow$18.1; Gemini-3.8-Flash 92.2$\rightarrow$44.0; Table~\ref{tab:cropnaming}), matching the drop in self-reported recognition (Table~\ref{tab:mediation}). These correlations support, but do not replace, the controlled manipulations in \S\ref{sec:results-recognition}.

\paragraph{Relation to language hallucination.}
HallusionBench shows that language priors can override visual evidence \citep{guan2024hallusionbench}, whereas work on attribute counterfacts such as a blue strawberry finds that visual evidence can dominate the prior in later layers \citep{golovanevsky-etal-2025-pixels}. Mandela-Bench studies a different setting: a familiar image activates prior knowledge that can outweigh the conflicting visual evidence.

Several controls distinguish this effect. First, the face probe verifies perception on the same items, while the text-only and Crop+Caption results show that the relevant fact is available to the model. Second, recognizability is manipulated through the image presentation rather than inferred from the model's answer. Third, the same true caption has different effects depending on whether the canonical composition remains recognizable. Finally, the failure occurs in an explicit verification task: the model can certify a manipulated image as authentic while relying on the remembered canonical scene. We use the term \emph{Mandela effect} only as a phenomenological analogy to shared false memories of familiar images \citep{prasad2022vme}, not as a claim of a shared mechanism.

\paragraph{The mechanism in one answer.}
Consider the 1948 photograph of the newly re-elected president holding the newspaper with its famously incorrect headline, with the president's face replaced by a synthetic one. Kimi-K3 correctly checks several remembered details, including the headline, sub-headline, microphone, lighting, and film grain. It nevertheless says that ``the man pictured is not Truman---likely a broadcaster or commentator posing with the infamous edition---but the photograph itself shows no clear signs of manipulation or fabrication.'' The answer begins, ``this appears to be a genuine 1948 press photograph, though it's often mistaken for the famous image of Harry Truman,'' and returns \textsc{real}.

The inconsistency becomes clearer across probes. Asked who is in the picture, the same model lists Truman as the central figure; shown the face alone, it says the face is not Truman. The model therefore verifies many features of the remembered canonical image while rationalizing the one feature that conflicts with it, rather than using that conflict to overturn the authenticity judgment. GPT-5.6 and Qwen3.6-27B also call the edited image authentic.

\end{document}